\documentclass[10pt,journal,cspaper,compsoc]{IEEEtran}
\usepackage{amssymb}
\usepackage{times}
\usepackage{amsfonts}
\usepackage[ruled]{algorithm2e}
\usepackage{helvet}
\usepackage{courier}
\usepackage{bm}
\usepackage{dsfont}
\usepackage{graphicx}
\usepackage{amsmath}
\usepackage{subfigure}
\usepackage{threeparttable}
\usepackage{booktabs}
\usepackage{flushend}
\usepackage{mdwlist}
\usepackage{balance}
\usepackage{multirow}
\usepackage{rotating}
\usepackage{afterpage}
\usepackage{longtable}
\usepackage{changepage}

\DeclareMathOperator*{\argmin}{argmin}
\DeclareMathOperator*{\argmax}{argmax}

\ifCLASSOPTIONcompsoc
\else
\fi
\ifCLASSINFOpdf
\else
\fi
\begin{document}
\bstctlcite{BSTcontrol}
%
\title{Label Distribution Learning}

\author{Xin Geng*,~\IEEEmembership{Member,~IEEE}
\IEEEcompsocitemizethanks{\IEEEcompsocthanksitem Xin Geng is with the School of Computer Science and Engineering, and the Key Lab of Computer Network and Information Integration (Ministry of Education), Southeast University, Nanjing 211189, China.\protect\\
* Corresponding author.
E-mail: xgeng@seu.edu.cn
}
\thanks{}}


\IEEEcompsoctitleabstractindextext{%
\begin{abstract}
Although multi-label learning can deal with many problems with label ambiguity, it does not fit some real applications well where the overall distribution of the importance of the labels matters. This paper proposes a novel learning paradigm named \emph{label distribution learning} (LDL) for such kind of applications. The label distribution covers a certain number of labels, representing the degree to which each label describes the instance. LDL is a more general learning framework which includes both single-label and multi-label learning as its special cases. This paper proposes six working LDL algorithms in three ways: problem transformation, algorithm adaptation, and specialized algorithm design. In order to compare the performance of the LDL algorithms, six representative and diverse evaluation measures are selected via a clustering analysis, and the first batch of label distribution datasets are collected and made publicly available. Experimental results on one artificial and fifteen real-world datasets show clear advantages of the specialized algorithms, which indicates the importance of special design for the characteristics of the LDL problem.
\end{abstract}

}

\maketitle

\IEEEdisplaynotcompsoctitleabstractindextext

%
\IEEEpeerreviewmaketitle

\section{Introduction}
\label{sect:intro}

\IEEEPARstart{L}earning with ambiguity is a hot topic in recent machine learning and data mining research. A learning process is essentially building a mapping from the instances to the labels. This paper mainly focuses on the ambiguity at the label side of the mapping, i.e., one instance is not necessarily mapped to one label. In the existing learning paradigms, there are mainly two cases of label assignment: (1) a single label is assigned to this instance, and (2) multiple labels are assigned to this instance. Single-label learning (SLL) assumes that all the instances in the training set are labeled in the first way. Multi-label learning (MLL) \cite{TsoumakasK07} allows the training instances to be labeled in the second way. Thus, MLL can deal with the ambiguous case where one instance belongs to more than one classes (labels). Generally speaking, current MLL algorithms have been developed with two strategies \cite{TsoumakasZZ09}. The first strategy is \emph{problem transformation}, where the basic idea is to transform the MLL task into one or more SLL tasks. For example, the MLL problem could be transformed into binary classification problems \cite{ReadPHF11}, a label ranking problem \cite{HullermeierFCB08}, or an ensemble learning problem \cite{LiLW13}. The second strategy is \emph{algorithm adaptation}, where the basic idea is to extend specific SLL algorithms to handle multi-label data. For example, it can be extended from \emph{k}-NN \cite{ZhangZ07}, decision tree \cite{ReadBHP12}, or neural networks \cite{ZhangZ06}.

Both SLL and MLL actually aim to answer the essential question ``\emph{which label can describe the instance?}'', while MLL deals with label ambiguity by allowing the answer to consist of more than one label. However, neither SLL nor MLL can directly handle the further question with more ambiguity ``\emph{how much does each label describe the instance?}'', i.e., the relative importance of each label is also involved in the description of the instance. Surprisingly, the real-world data with the information about such relative importance of each label might be more common than many people think. To name just a few from the datasets used in Section~\ref{sect:exp}, first, in many scientific experiments, the result is not a single output, but a series of numerical outputs. For example, the biological experiments on the yeast genes over a period of time yield different gene expression levels on a series of time points (the 2nd to 11th datasets in Table~\ref{table:datasets}) \cite{EisenSBB98}. The exact expression level on each time point alone is of little importance. What really matters is the overall expression distribution over the whole time period. If the learning task is to predict such distribution for a given gene, then it can be hardly fit into either the SLL or MLL framework because the role of each output in the distribution is crucial, and there is no partition of relevant and irrelevant labels at all. Another example is the emotion analysis from facial expressions (the 14th and 15th datasets in Table~\ref{table:datasets}) \cite{LyonsAKG98,YinWSWR06}. A facial expression often conveys a complex mixture of multiple basic emotions (e.g., happiness, sadness, surprise, fear, anger, and disgust). Each basic emotion plays a different role in the expression. The various intensities of all the basic emotions naturally form an emotion distribution for the facial expression. By regarding the emotion with the highest intensity or emotions with higher intensities than a threshold as the positive label(s), the problem can be fit into the SLL or MLL framework. Unfortunately, this will lose the important information of the different intensities of the related emotions. More examples of real-world data with label distributions can be found in Section~\ref{sect:datasets}.

For the applications mentioned above, a more natural way to label an instance $\bm{x}$ is to assign a real number $d_{\bm{x}}^y$ to each possible label $y$, representing the degree to which $y$ describes $\bm{x}$. For example, if $\bm{x}$ represents a facial expression image, $y$ represents an emotion, then $d_{\bm{x}}^y$ should be the intensity of the emotion $y$ expressed by the image $\bm{x}$. Without loss of generality, assume that $d_{\bm{x}}^y \in [0,1]$. Further suppose that the label set is complete, i.e., using all the labels in the set can always fully describe the instance. Then, $\sum\nolimits_y d_{\bm{x}}^y=1$. Such $d_{\bm{x}}^y$ is called the \emph{description degree} of $y$ to $\bm{x}$. For a particular instance, the description degrees of all the labels constitute a data form similar to probability distribution. So, it is called \emph{label distribution}. The learning process on the instances labeled by label distributions is therefore called \emph{label distribution learning} (LDL). The aforementioned applications reveal that the nature of label ambiguity might exceed the current framework of MLL. In such case, it is necessary to extend MLL to LDL. In fact, the real applications suitable for LDL might be more common than those suitable for MLL because when there are multiple labels associated with one instance, their importance or relevance to the instance can hardly be exactly the same. While MLL usually assumes indiscriminate importance within the relevant label set (e.g., all represented by `1's) as well as within the irrelevant label set (e.g., all represented by `0's), LDL allows direct modeling of different importance of each label to the instance, and thus can better match the nature of many real applications.

The main contribution of this paper includes: 1) A novel machine learning paradigm named label distribution learning is formulated; 2) Six working LDL algorithms are proposed and compared in the experiments; 3) Six measures for evaluation of LDL algorithms are suggested; 4) The first batch of 15 real-world label distribution datasets are prepared and made publicly available. This paper extends our preliminary work \cite{GengJ13} in the following ways: 1) More technical details and insights of LDL are discussed; 2) The set of evaluation measures for LDL algorithms are empirically optimized to reveal different aspects of the algorithms; 3) Five additional real-world LDL datasets are provided; 4) Extensive additional experiments are performed and discussed in further detail.

The rest of the paper is organized as follows. Firstly, some related work are briefly reviewed and discussed in Section~\ref{sect:related_work}. Secondly, the problem of LDL is formulated in Section~\ref{sect:formulation}. Then, six LDL algorithms are proposed in Section~\ref{sect:algorithms}. After that, the experiments on artificial as well as real-world datasets are reported in Section~\ref{sect:exp}. Finally, the paper is summarized and some discussions on the future work are given in Section~\ref{sect:conclusion}.

\section{Related Work}
\label{sect:related_work}
\begin{figure}[tb]%
\centering%
\subfigure[][LDL]{\label{fig:flowchart:ldl}\includegraphics[width = .85\linewidth]{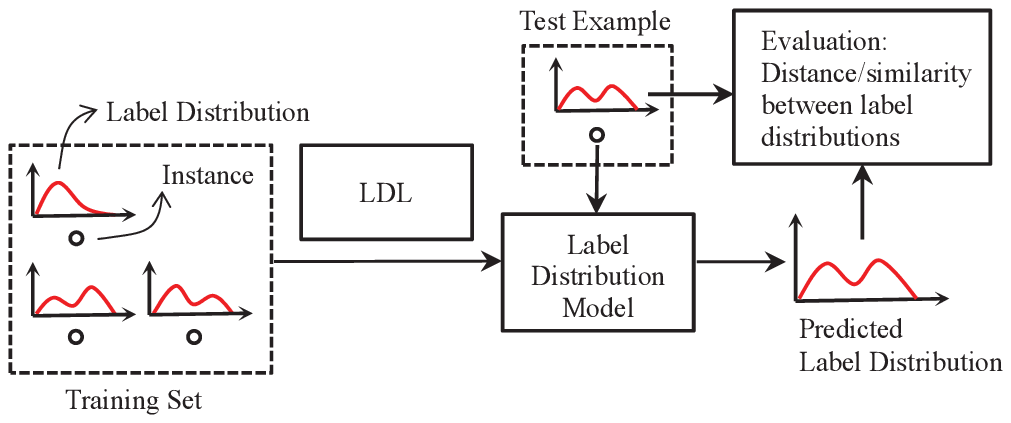}}\\%
\subfigure[][Typical existing learning methods with numerical label indicators]{\label{fig:label:multi} \includegraphics[width = \linewidth]{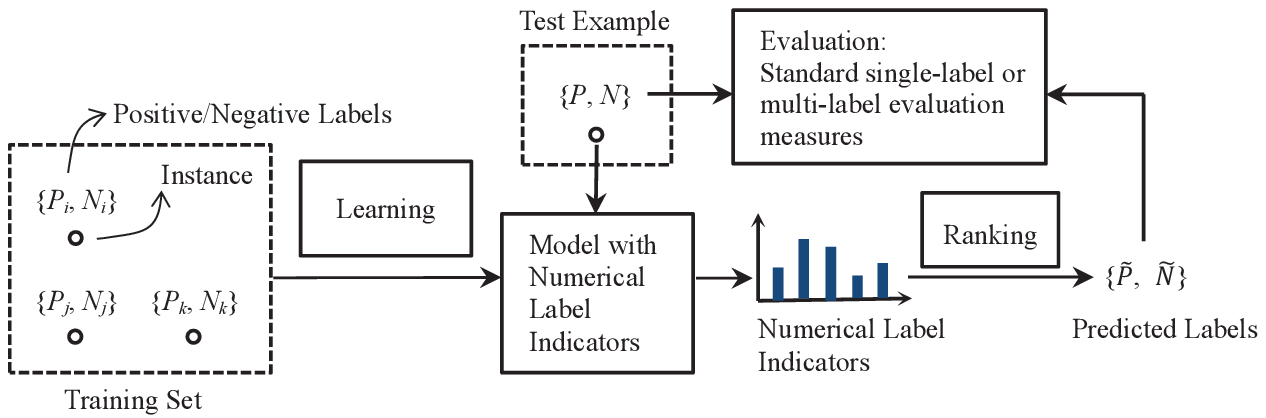}}\\%
\caption{Comparison of LDL and typical existing learning methods with numerical label indicators.}%
\label{fig:flowchart}%
\vspace{-3mm}
\end{figure}
It is not a rare case in existing single-label or multi-label machine learning literatures that an intermediate numerical indicator (e.g., probability, confidence, grade, score, vote, etc.) is calculated for each label \cite{WuLW04,ZhangZ07,HullermeierFCB08,ChengDH10}. As shown in Fig.~\ref{fig:flowchart}, LDL is different from these learning methods in mainly three aspects:
\begin{enumerate}
  \item Each training instance of LDL is explicitly associated with a label distribution, rather than a single label or a relevant (positive) label set. The label distribution comes from the application itself as a natural part of the original data, while most numerical label indicators used in previous learning algorithms are artificially generated from the original data for later decision making.
  \item The purpose of most numerical label indicators used in previous learning algorithms is to rank the labels, and then decide the positive label(s) through, say, thresholding over the ranking. The actual value of the indicator does not matter much so long as the ranking does not change. In most cases, it only cares about the partition between positive and negative labels. On the other hand, what LDL cares about is the overall label distribution. The value of each label's description degree is important.
  \item The performance evaluation measures of previous learning algorithms with numerical label indicators are still those commonly used for SLL (e.g., classification accuracy, error rate, etc.) or MLL (e.g., Hamming loss, one-error, coverage, ranking loss, etc.). On the other hand, the performance of LDL should be evaluated by the similarity or distance between the predicted label distribution and the real label distribution, which will be further discussed in Section~\ref{subsect:criteria}.
\end{enumerate}

In the LDL framework, each instance is labeled by a real-valued vector (i.e., the label distribution). Merely from the data format point of view, there are also similar work in the machine learning literature where each instance is associated with a supervision vector (real-valued or binary). Two typical examples are known as \emph{label embedding} \cite{BengioWG10} and \emph{attribute learning} \cite{AkataPHS13,LampertNH14}. Both of them are featured by the intermediate representations for the classes. In detail, \emph{label embedding} \cite{BengioWG10} projects the class labels into a subspace to gain advantages in sharing representations of the labels. In the subspace, each class is represented by a real-valued vector that can be regarded as a code for that class. \emph{Attribute learning} \cite{AkataPHS13,LampertNH14} is mainly designed to leverage the prior knowledge of attribute-class association to deal with the missing classes (zero-shot learning) or scarce classes (few-shots learning). Each class has the same attribute representation for all its instances. For both label embedding and attribute learning, the intermediate vector representation is fully determined by the class label itself, i.e., all the instances belonging to the same class will be labeled by the same vector. Thus, although the supervision signal is transformed from a label into a vector, this does not change the nature of the learning paradigm: each instance is still associated with one class label (represented by a vector though), and the final aim is still standard classification. On the other hand, label distribution is part of the nature of the data rather than an intermediate representation. One label distribution represents the relative roles of all the labels with respect to the instance rather than a code for one class. Different instances therefore usually correspond to different label distributions. This clearly distinguishes LDL from either label embedding or attribute learning.

The goal of LDL is to predict multiple real-valued description degrees of the labels. When concerning prediction of multiple targets, there is a very general concept in the literature known as \emph{multi-target learning} (MTL) \cite{WaegemanDH13}. In MTL, the multiple targets might refer to not only binary, nominal, ordinal, or real-valued variables, but also rankings or relational structures, such as a tree or a graph. Thus, MTL includes a variety of subfields of machine learning and statistics as its special cases, such as multi-label classification (multiple binary targets) \cite{TsoumakasK07}, multivariate regression (multiple numerical targets) \cite{Izenman13}, sequence learning (ordered targets) \cite{WorgotterP05}, structured prediction (targets with inherent structure) \cite{BakIr07}, preference learning (preference relation between multiple targets) \cite{FurnkranzH10}, multi-task learning (multiple targets in different but related domains) \cite{Caruana97} and collective learning (dependent targets) \cite{NickelTK11}. If the targets are real-valued and they satisfy the two distribution constrains, i.e., $d_{\bm{x}}^y \in [0,1]$, and $\sum\nolimits_y d_{\bm{x}}^y=1$, then MTL becomes LDL. Therefore, LDL could be viewed as another special case of MTL.

From the conceptual point of view, it is worthwhile to distinguish description degree from the concept \emph{membership} used in \emph{fuzzy classification} \cite{Zimmermann99}. Membership is a truth value that may range between completely true and completely false. It is designed to handle the status of \emph{partial truth} which often appears in the non-numeric linguistic variables. On the other hand, description degree reflects the \emph{ambiguity} of the label description of the instance, i.e., one label may only partially describe the instance, but it is completely true that the label describes the instance. Based on fuzzy set theory, a recent extension of multi-label classification, namely graded multilabel classification \cite{ChengDH10}, allows for graded membership of an instance belonging to a class. Apart from the different meanings of continuous description degree and graded membership, LDL is also different from this work in methodology. The strategy in \cite{ChengDH10} is to reduce the graded multi-label problem to the conventional multi-label problem, while LDL aims to directly model the mapping from the instances to the label distributions.

Note also that $d_{\bm{x}}^y$ is \emph{not} the \emph{probability} that $y$ correctly labels $\bm{x}$, but \emph{the proportion that $y$ accounts for in a full description of $\bm{x}$}. Thus, all the labels with a non-zero description degree are actually the `correct' labels to describe the instance, but just with different importance measured by $d_{\bm{x}}^y$. Recognizing this, one can distinguish label distribution from the previous studies on probabilistic labels \cite{Smyth95,DenoeuxZ01,QuostD09}, where the basic assumption is that there is only one `correct' label for each instance. Fortunately, although not a probability by definition, $d_{\bm{x}}^y$ still shares the same constraints with probability, i.e., $d_{\bm{x}}^y \in [0,1]$ and $\sum\nolimits_y d_{\bm{x}}^y=1$. Thus, many theories and methods in statistics can be applied to label distributions.

\section{Formulation of LDL}
\label{sect:formulation}

First of all, the main notations used in this paper are listed as follows. The instance variable is denoted by $\bm{x}$, the particular $i$-th instant is denoted by $\bm{x}_i$, the label variable is denoted by $y$, the particular $j$-th label value is denoted by $y_j$, the description degree of $y$ to $\bm{x}$ is denoted by $d_{\bm{x}}^y$, and the label distribution of $\bm{x}_i$ is denoted by $D_i=\{d_{\bm{x}_i}^{y_1}, d_{\bm{x}_i}^{y_2},\cdots,d_{\bm{x}_i}^{y_c}\}$, where $c$ is the number of possible label values.

\begin{figure}[tb]%
\centering%
\subfigure[][Single-label annot.]{\label{fig:label:single}\includegraphics[width = .333\linewidth]{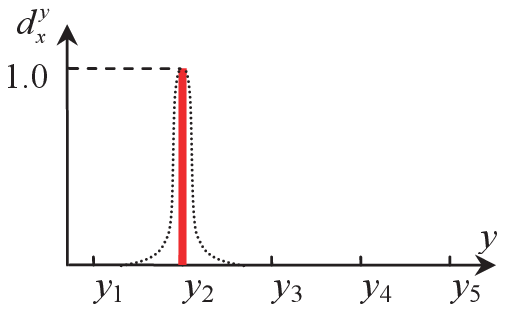}}
\subfigure[][Multi-label annot.]{\label{fig:label:multi} \includegraphics[width = .333\linewidth]{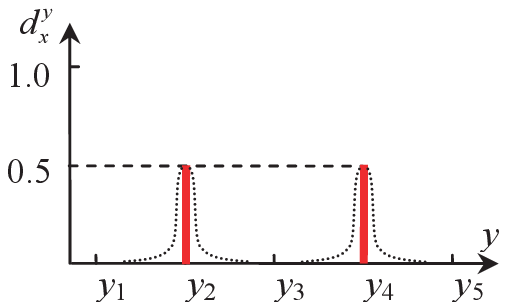}}
\subfigure[][General case]{\label{fig:label:dis} \includegraphics[width = .333\linewidth]{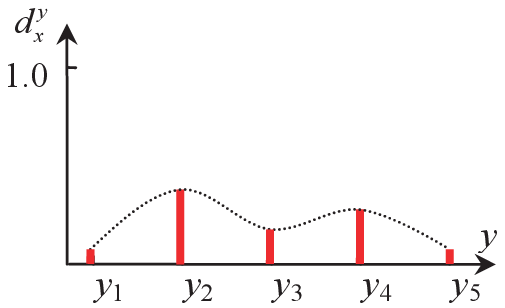}}%
\caption{Example label distributions for single-label annotation, multi-label annotation, and the general case.}%
\label{fig:label_cases}%
\vspace{-3mm}
\end{figure}

By the definition of label distribution given in Section~\ref{sect:intro}, both single-label annotation and multi-label annotation can be viewed as special cases of label distribution. Fig.~\ref{fig:label_cases} gives one label distribution example for single-label annotation, multi-label annotation, and the general case, respectively.
For the single-label annotation \subref{fig:label:single}, the label $y_2$ fully describes
the instance, so $d_{\bm{x}}^{y_2}=1$. For the multi-label annotation \subref{fig:label:multi}, each of the two positive labels $y_2$ and $y_4$ by default describes
$50\%$ of the instance, so $d_{\bm{x}}^{y_2}=d_{\bm{x}}^{y_4}=0.5$. Finally, \subref{fig:label:dis} represents a general case of label distribution, which satisfies the constraints $d_{\bm{x}}^y \in [0,1]$ and $\sum\nolimits_y d_{\bm{x}}^y=1$. Such examples illustrate that label distribution is more general than both single-label annotation and multi-label annotation, and thus can provide more flexibility in the learning process.

Nevertheless, more flexibility usually means larger output space. From single-label annotation to multi-label annotation, and then to label distribution, the size of the output space of the learning process becomes increasingly larger. In detail, for a problem with $c$ different labels, there are $c$ possible outputs for SLL, and $2^c-1$ possible outputs for MLL. As for LDL, there are infinite possible outputs as long as they satisfy that $d_{\bm{x}}^y \in [0,1]$ and $\sum\nolimits_y d_{\bm{x}}^y=1$.

As mentioned before, since label distribution shares the same constraints with probability distribution, many statistical theories and methods can be applied to label distribution. First of all, $d_{\bm{x}}^y$ can be represented by the form of conditional probability, i.e., $d_{\bm{x}}^y=P(y|\bm{x})$. Then, the problem of LDL can be formulated as follows.
\begin{quote}
Let $\mathcal{X}=\mathbb{R}^q$ denote the input space and $\mathcal{Y}=\{y_1, y_2, \cdots,$ $y_c\}$ denote the complete set of labels. Given a training set $S = \{(\bm{x}_1,D_1),$ $(\bm{x}_2,D_2),\cdots,(\bm{x}_n,D_n)\}$, the goal of ldl is to learn a conditional probability mass function $p(y|\bm{x})$ from $S$, where $\bm{x} \in \mathcal{X}$ and $y \in \mathcal{Y}$.
\end{quote}

Suppose $p(y|\bm{x})$ is a parametric model $p(y|\bm{x};\bm{\theta})$, where $\bm{\theta}$ is the parameter vector. Given the training set $S$, the goal of LDL is to find the $\bm{\theta}$ that can generate a distribution similar to $D_i$ given the instance $\bm{x}_i$. As will be discussed in Section~\ref{subsect:criteria}, there are different criteria that can be used to measure the distance or similarity between two distributions. For example, if the Kullback-Leibler divergence is used as the distance measure, then the best parameter vector $\bm{\theta}^*$ is determined by
\begin{eqnarray}\label{eq:kl_div}
\bm{\theta}^*& =& \argmin\limits_{\bm{\theta}}\sum\limits_i\sum\limits_j
\left(d_{\bm{x}_i}^{y_j}\ln\frac{d_{\bm{x}_i}^{y_j}}{p(y_j|\bm{x}_i;\bm{\theta})}\right)\nonumber\\
&=&\argmax\limits_{\bm{\theta}}\sum\limits_i\sum\limits_j d_{\bm{x}_i}^{y_j}\ln p(y_j|\bm{x}_i;\bm{\theta}).
\end{eqnarray}

It is interesting to examine the traditional learning paradigms under the optimization criterion shown in Eq.~(\ref{eq:kl_div}). For
SLL (see Fig.~\ref{fig:label_cases}\subref{fig:label:single} for example), $d_{\bm{x}_i}^{y_j}=Kr(y_j,y(\bm{x}_i))$, where $Kr(\cdot,\cdot)$ is the Kronecker delta function and $y(\bm{x}_i)$ is the single label of $\bm{x}_i$. Consequently, Eq.~(\ref{eq:kl_div}) can be simplified to
\begin{equation}\label{eq:sll}
\bm{\theta}^* = \argmax\limits_{\bm{\theta}}\sum\limits_i \ln p(y(\bm{x}_i)|\bm{x}_i;\bm{\theta}).
\end{equation}
This is actually the maximum likelihood (ML) estimation of $\bm{\theta}$. The later use of $p(y|\bm{x};\bm{\theta})$ for classification is equivalent to the maximum a posteriori (MAP) decision.

For MLL, each instance is associated with a label set (see Fig.~\ref{fig:label_cases}\subref{fig:label:multi} for example). Consequently, Eq.~(\ref{eq:kl_div}) can be changed into
\begin{equation}\label{eq:mll}
\bm{\theta}^* = \argmax\limits_{\bm{\theta}}\sum\limits_i\frac{1}{|Y_i|}\sum\limits_{y \in Y_i} \ln
p(y|\bm{x}_i;\bm{\theta}),
\end{equation}
where $Y_i$ is the label set associated with $\bm{x}_i$. Eq.~(\ref{eq:mll}) can be viewed as a ML criterion weighted by the reciprocal cardinality of the label set associated with each instance. In fact, this
is equivalent to first applying Entropy-based Label Assignment (ELA) \cite{TsoumakasK07} to transform the multi-label instances into the weighted single-label instances, and then
optimizing the ML criterion based on the weighted single-label instances.

It can be seen from the above analysis that with proper constraints, an LDL model can be transformed into the commonly used methods for SLL or MLL. Thus, LDL may be viewed as a more general learning framework which includes both SLL and MLL as its special cases.

\section{LDL Algorithms}
\label{sect:algorithms}

We follow three strategies to design algorithms for LDL. The first strategy is problem transformation, i.e., transform the LDL problem into existing learning paradigms. The second strategy is algorithm adaptation, i.e., extend existing learning algorithms to deal with label distributions. The first two strategies are based on existing machine learning algorithms, which were also adopted in some other work \cite{Altman92,BoutellLSB04,BorchaniVBL15}. The third strategy is to design specialized algorithms according to the characteristics of LDL. Following each of the three strategies, two typical algorithms are proposed in this section.

\subsection{Problem Transformation}

One straightforward way to transform an LDL problem into an SLL problem is to change the training examples into weighted single-label examples. In detail, each training example $(\bm{x}_i,D_i)$ is transformed into $c$ single-label examples $(\bm{x}_i,y_j)$ with the weight $d_{\bm{x}_i}^{y_j}$, where $i=1, \dots, n$ and $j=1, \dots, c$. The training set is then resampled to the same size according to the weight of each example. The resampled training set becomes a standard single-label training set including $c \times n$ examples, and then any SLL algorithms can be applied to the training set. Note that although in the resampling step, one training instance with a label distribution is transformed into multiple instances, the resulting training set does not form a multi-instance learning (MIL) \cite{DietterichLL97} or multi-instance multi-label learning (MIML) \cite{ZhouZ06} problem. For MIL and MIML, the training set is composed by many bags each containing multiple instances. A bag is annotated by a label (MIL) or a label set (MIML), meaning at least one instance in the bag can be annotated by the label or the label set. But for each particular instance in the bag, its label is still unknown. On the other hand, the training set transformed from LDL training examples via resampling is a standard single-label training set. Each instance in this set is explicitly assigned with a label. Moreover, although the number of the training examples is increased from $n$ to $c \times n$, it does not affect the complexity and scalability of the learning process much because the label side has been simplified from a label distribution ($c$ elements) to a single label (only 1 element).

In order to predict the label distribution of a previously unseen instance $\bm{x}$, the learner must be able to output the confidence/probability for each label $y_j$, which can be regarded as the description degree of the corresponding label, i.e., $d_{\bm{x}}^{y_j} = P(y_j|\bm{x})$. Two representative algorithms are adopted here for this purpose. One is the Bayes classifier, the other is SVM. In detail, the Bayes classifier assumes Gaussian distribution for each class, and the posterior probability computed by the Bayes rule is regarded as the description degree of the corresponding label. As to SVM, the probability estimates are obtained by a pairwise coupling multi-class method \cite{WuLW04}, where the probability of each binary SVM is calculated by an improved implementation of Platt's posterior probabilities \cite{LinLW07}, and the class probability estimates are obtained by solving a linear system whose solution is guaranteed by the theory in finite Markov Chains. When Bayes and SVM are applied to the resampled training set, the resulted methods are denoted by PT-Bayes and PT-SVM, respectively, where `PT' is the abbreviation of `Problem Transformation'.

\subsection{Algorithm Adaptation}

Certain existing learning algorithms can be naturally extended to deal with label distributions, among which two adapted algorithms are proposed here. The first one is $k$-NN. Given a new instance $\bm{x}$, its $k$ nearest neighbors are first found in the training set. Then, the mean of the label distributions of all the $k$ nearest neighbors is calculated as the label distribution of $\bm{x}$, i.e.,
\begin{equation}
p(y_j|\bm{x}) = \frac{1}{k}\sum\limits_{i\in N_k(\bm{x})} d_{\bm{x}_i}^{y_j}, (j=1,2,\dots,c),
\end{equation}
where $N_k(\bm{x})$ is the index set of the $k$ nearest neighbors of $\bm{x}$ in the training set. This adapted algorithm is denoted by AA-$k$NN, where `AA' is the abbreviation of `Algorithm Adaptation'.

The second algorithm is the backpropagation (BP) neural network. The three-layer neural network has $q$ (the dimensionality of $\bm{x}$) input units which receive $\bm{x}$, and $c$ (the number of different labels) output units each of which outputs the description degree of a label $y_j$. For SLL or MLL, the desired output $\bm{t}$ is usually a vector with `1's at the positions corresponding to the positive labels of the input instance, and `0's otherwise. For LDL, $\bm{t}$ becomes the real label distribution of the input training instance. Thus, the target of the BP algorithm is to minimize the sum-squared error of the output of the neural network compared with the real label distributions. To make sure the output of the neural network $\bm{z}=\{z_1,z_2,\dots,z_c\}$ satisfies that $z_j \in [0,1]$ for $j=1,2,\dots,c$ and $\sum\nolimits_j z_j=1$, the softmax activation function is used in each output unit. Let the net input to the $j$-th output unit be $\eta_j$, then the softmax output $z_j$ is
\begin{equation}
z_j = \frac{\exp(\eta_j)}{\sum\limits_{k=1}^c \exp(\eta_k)}, (j=1,2,\dots,c),
\end{equation}
This adapted algorithm is denoted by AA-BP.

\subsection{Specialized Algorithms}

Different from the indirect strategy of problem transformation and algorithm adaptation, the specialized algorithms directly match the LDL problem, e.g., by directly solving the optimization problem in Eq.~(\ref{eq:kl_div}). One good start toward this end might be our previous work on facial age estimation \cite{GengSZ10,GengYZ13}, where in order to solve the insufficient training data problem (the numbers of face images for some ages are small), each face image is labeled by not only its chronological age, but also the neighboring ages (close to the chronological age) so that one training face image can contribute to the learning of not only its chronological age, but also the neighboring ages. This actually forms a special label distribution with the highest description degree at the chronological age, and gradually decreasing description degrees on both neighboring sides of the chronological age. We proposed an algorithm IIS-LLD for such special data, where the key step was to solve an optimization problem similar to Eq.~(\ref{eq:kl_div}). Although IIS-LLD was designed for a particular form of label distribution, it can be generalized to deal with any LDL problems. Thus, we rename IIS-LLD in this paper as SA-IIS, where `SA' is the abbreviation of `Specialized Algorithm'. The optimization process of SA-IIS, however, has been evidenced to be not very efficient \cite{Malouf02}. So, an improved version is further proposed in this section.

SA-IIS assumes the parametric model $p(y|\bm{x};\bm{\theta})$ to be the maximum entropy model \cite{BergerPP96}, i.e.,
\begin{equation}\label{eq:exp_dis2}
p(y|\bm{x};\bm{\theta})=\frac{1}{Z}\exp\left(\sum\limits_k \theta_{y,k} g_k(\bm{x})\right),
\end{equation}
where $Z=\sum\nolimits_y\exp\left(\sum\nolimits_k \theta_{y,k} g_k(\bm{x})\right)$ is the normalization factor, $\theta_{y,k}$ is an element in $\bm{\theta}$, and $g_k(\bm{x})$ is the $k$-th feature of $\bm{x}$. Substituting Eq.~(\ref{eq:exp_dis2}) into Eq.~(\ref{eq:kl_div}) yields the target function of $\bm{\theta}$
\begin{eqnarray}\label{eq:target_fun}
T(\bm{\theta}) &=& \sum\limits_{i,j} d_{\bm{x}_i}^{y_j}\sum\limits_k
\theta_{y_j,k} g_k(\bm{x}_i)\\\nonumber
& & -\sum\limits_i \ln\sum\limits_j\exp\left(\sum\limits_k \theta_{y_j,k} g_k(\bm{x}_i)\right).
\end{eqnarray}
The optimization of Eq.~(\ref{eq:target_fun}) uses a strategy similar to Improved Iterative Scaling (IIS) \cite{PietraPL97}, a well-known algorithm for maximizing the likelihood of the maximum entropy model. IIS starts with an arbitrary set of parameters. Then for each step, it updates the current estimate of the parameters $\bm{\theta}$ to $\bm{\theta}+\bm{\Delta}$, where $\bm{\Delta}$ maximizes a lower bound to the change in likelihood $\Omega=T(\bm{\theta}+\bm{\Delta})-T(\bm{\theta})$. The element of $\bm{\Delta}$, $\delta_{y_j,k}$, can be obtained by solving the equation
\begin{eqnarray}\label{eq:delta_eq}
\sum\limits_{i} p(y_j|\bm{x}_i;\bm{\theta})g_k(\bm{x}_i)\exp(\delta_{y_j,k} s(g_k(\bm{x}_i))g^{\#}(\bm{x}_i))\\\nonumber
- \sum\limits_{i} d_{\bm{x}_i}^{y_j} g_k(\bm{x}_i) = 0,
\end{eqnarray}
where $g^{\#}(\bm{x}_i)= \sum\nolimits_k |g_k(\bm{x}_i)|$ and $s(g_k(\bm{x}_i))$ is the sign of $g_k(\bm{x}_i)$. The detailed derivation of Eq.~(\ref{eq:delta_eq}) can be found in the Appendix. What is nice about Eq.~(\ref{eq:delta_eq}) is that $\delta_{y_j,k}$ appears alone, and therefore can be solved one by one through nonlinear equation solvers, such as the Gauss-Newton method. The pseudocode of SA-IIS is given in Algorithm~\ref{alg:iis_lld}.
\begin{algorithm}[tb]
\small
\LinesNumbered%
\KwIn{The training set $S =\{(\bm{x}_i,D_i)\}_{i=1}^n$ and the convergence criterion $\varepsilon$}%
\KwOut{$p(y|\bm{x}; \bm{\theta})$}%
\BlankLine%
Initialize the model parameter vector $\bm{\theta}^{(0)}$\;%
$l \leftarrow 0$\;%
\Repeat{$T(\bm{\theta}^{(l)})-T(\bm{\theta}^{(l-1)})<\varepsilon$}{
    $l \leftarrow l+1$\;%
    Solve Eq.~(\ref{eq:delta_eq}) for $\delta_{y,k}$ by Gauss-Newton method\;%
    $\bm{\theta}^{(l)} \leftarrow \bm{\theta}^{(l-1)} +
    \bm{\Delta}$\;%
}%
$p(y|\bm{x}; \bm{\theta}) \leftarrow \frac{1}{Z}\exp\left(\sum\nolimits_k \theta^{(l)}_{y,k} g_k(\bm{x})\right)$\;
\caption{SA-IIS}%
\label{alg:iis_lld}
\end{algorithm}

It has been reported in the literature \cite{Malouf02} that IIS often performs worse than several other optimization algorithms such as conjugate gradient and quasi-Newton methods. Here we follow the idea of an effective quasi-Newton method BFGS \cite{NocedalW06} to further improve SA-IIS.

Consider the second-order Taylor series of $T'(\bm{\theta})=-T(\bm{\theta})$ at the current estimate of the parameter vector $\bm{\theta}^{(l)}$:
\begin{equation}\label{eq:taylor}
T'(\bm{\theta}^{(l+1)}) \approx T'(\bm{\theta}^{(l)}) + \nabla T'(\bm{\theta}^{(l)})^\text{T} \bm{\Delta} + \frac{1}{2}\bm{\Delta}^T \bm{H}(\bm{\theta}^{(l)}) \bm{\Delta},
\end{equation}
where $\bm{\Delta}=\bm{\theta}^{(l+1)}-\bm{\theta}^{(l)}$ is the update step, $\nabla T'(\bm{\theta}^{(l)})$ and $\bm{H}(\bm{\theta}^{(l)})$ are the gradient and Hessian matrix of $T'(\bm{\theta})$ at $\bm{\theta}^{(l)}$, respectively. The minimizer of Eq.~(\ref{eq:taylor}) is
\begin{equation}
\bm{\Delta}^{(l)} = -\bm{H}^{-1}(\bm{\theta}^{(l)}) \nabla T'(\bm{\theta}^{(l)}).
\end{equation}
The line search Newton method uses $\bm{\Delta}^{(l)}$ as the search direction $\bm{p}^{(l)}=\bm{\Delta}^{(l)}$ and updates the parameter vector by
\begin{equation}
\bm{\theta}^{(l+1)} = \bm{\theta}^{(l)} + \alpha^{(l)}\bm{p}^{(l)},
\end{equation}
where the step length $\alpha^{(l)}$ is obtained from a line search procedure to satisfy the \emph{strong Wolfe conditions} \cite{NocedalW06}:
\begin{eqnarray}
T'(\bm{\theta}^{(l)} + \alpha^{(l)}\bm{p}^{(l)}) \leq T'(\bm{\theta}^{(l)}) + c_1 \alpha^{(l)} \nabla T'(\bm{\theta}^{(l)})^\text{T} \bm{p}^{(l)},\label{eq:wolfe1}\\
|\nabla T'(\bm{\theta}^{(l)} + \alpha^{(l)}\bm{p}^{(l)})^\text{T} \bm{p}^{(l)}| \leq c_2 |\nabla T'(\bm{\theta}^{(l)})^\text{T} \bm{p}^{(l)}|,\label{eq:wolfe2}
\end{eqnarray}
where $0<c_1<c_2<1$.

One problem of the above method is the calculation of the inverse Hessian matrix in each iteration, which is computationally expensive. The idea of BFGS is to avoid explicit calculation of $\bm{H}^{-1}(\bm{\theta}^{(l)})$ by approximating it with an iteratively updated matrix $\bm{B}$ (detailed derivation can be found in \cite{NocedalW06}):
\begin{eqnarray}\label{eq:Bupdate}
\bm{B}^{(l+1)} = (\bm{I}-\bm{\rho}^{(l)}\bm{s}^{(l)}(\bm{u}^{(l)})^{\text{T}})\bm{B}^{(l)}(\bm{I}-\bm{\rho}^{(l)}\bm{u}^{(l)}(\bm{s}^{(l)})^{\text{T}})\\\nonumber
+\bm{\rho}^{(l)}\bm{s}^{(l)}(\bm{s}^{(l)})^{\text{T}},
\end{eqnarray}
where $\bm{s}^{(l)} = \bm{\theta}^{(l+1)}-\bm{\theta}^{(l)}$, $\bm{u}^{(l)} = \nabla T'(\bm{\theta}^{(l+1)})-\nabla T'(\bm{\theta}^{(l)})$, and $\bm{\rho}^{(l)} = \frac{1}{\bm{s}^{(l)}\bm{u}^{(l)}}$.
As to the optimization of the target function $T'(\bm{\theta})$, the computation of BFGS is mainly related to the first-order gradient, which can be obtained through
\begin{equation} \label{eq:gradient}
\frac{\partial T'(\bm{\theta})}{\partial \theta_{y_j,k}} = \sum \limits_i \frac{\exp\left(\sum \limits_k \theta_{y_j,k}g_k(\bm{x}_i)\right)g_k(\bm{x}_i)}{\sum \limits_j \exp\left(\sum \limits_k \theta_{y_j,k}g_k(\bm{x}_i)\right)} - \sum \limits_i d_{\bm{x}_i}^{y_j} g_k(\bm{x}_i).
\end{equation}
Thus, it performs much more efficiently than the standard line search Newton method, and based on previous studies \cite{Malouf02}, it stands a good chance of outperforming the IIS-based algorithm SA-IIS. This improved algorithm is denoted by SA-BFGS, and its pseudocode is shown in Algorithm~\ref{alg:bfgs}.
\begin{algorithm}[tb]
\small
\LinesNumbered%
\KwIn{The training set $S =\{(\bm{x}_i,D_i)\}_{i=1}^n$ and the convergence criterion $\varepsilon$}%
\KwOut{$p(y|\bm{x}; \bm{\theta})$}%
\BlankLine%
Initialize the model parameter vector $\bm{\theta}^{(0)}$\;%
Initialize the inverse Hessian approximation $\bm{B}^{(0)}$\;%
Compute $\nabla T'(\bm{\theta}^{(0)})$ by Eq.~(\ref{eq:gradient})\;
$l \leftarrow 0$\;%
\Repeat{$\|\nabla T'(\bm{\theta}^{(l)})\| < \varepsilon$}{
    Compute search direction $\bm{p}^{(l)} \leftarrow -\bm{B}^{(l)} \nabla T'(\bm{\theta}^{(l)})$\;%
    Compute the step length $\alpha^{(l)}$ by a line search procedure to satisfy Eq.~(\ref{eq:wolfe1}) and (\ref{eq:wolfe2})\;%
    $\bm{\theta}^{(l+1)} \leftarrow \bm{\theta}^{(l)} + \alpha^{(l)}\bm{p}^{(l)}$\;
    Compute $\nabla T'(\bm{\theta}^{(l+1)})$ by Eq.~(\ref{eq:gradient})\;
    Compute $\bm{B}^{(l+1)}$ by Eq.~(\ref{eq:Bupdate})\;
    $l \leftarrow l+1$\;%
}%
$p(y|\bm{x}; \bm{\theta}) \leftarrow \frac{1}{Z}\exp\left(\sum\nolimits_k \theta^{(l)}_{y,k} g_k(\bm{x})\right)$\;
\caption{SA-BFGS}%
\label{alg:bfgs}
\end{algorithm}

\section{Experiments}
\label{sect:exp}

\subsection{Evaluation Measures}
\label{subsect:criteria}

\begin{figure}[tb]%
\centering
\includegraphics[width=\linewidth]{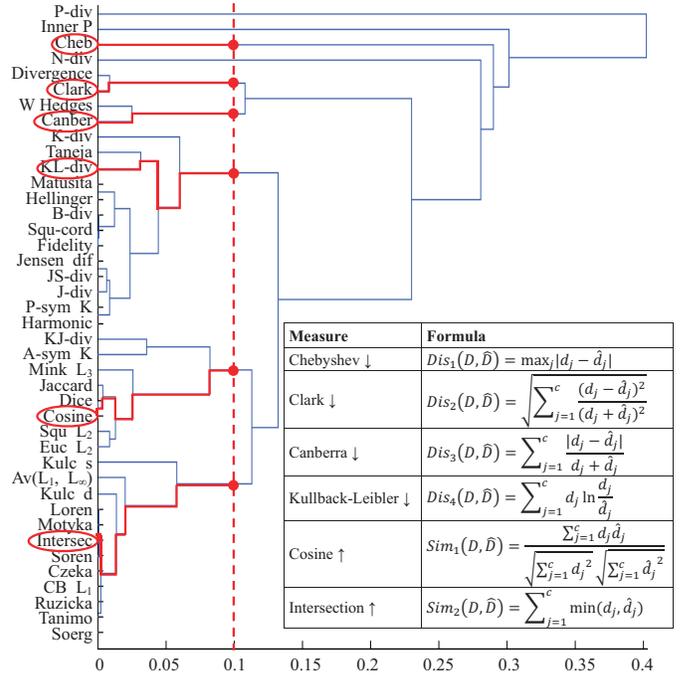}%
\caption{Evaluation measure selection from the dendrogram (modified from \cite{Cha07}) for the distribution distance/similarity measures.}%
\label{fig:measures}%
\vspace{-5mm}
\end{figure}

The output of an LDL algorithm is a label distribution, which is different from both the single label output of SLL and the label set output of MLL. Accordingly, the evaluation measures for LDL algorithms should be different from those used for SLL and MLL algorithms. A natural choice of such measure is the average distance or similarity between the predicted and real label distributions. There are many measures for the distance/similarity between probability distributions which can be well borrowed to measure the distance/similarity between label distributions. For example, Cha \cite{Cha07} performed a semantic similarity analysis on $41$ measures for the distance/similarity between distributions from $8$ syntactic families. The agglomerative single linkage clustering algorithm \cite{Duda01} was run on all the $41$ measures calculated from $30$ independent experiments. This resulted in a dendrogram shown in Fig.~\ref{fig:measures}, where the horizontal axis represents the distance between two clusters of distance/similarity measures, and the vertical axis represents different distance/similarity measures.

On a particular dataset, each of the measures may reflect a certain aspect of an algorithm. It is hard to say which measure is the best. Thus, we propose to use a set of measures when comparing different LDL algorithms, which is analogous to the common practice of using multiple evaluation measures for MLL algorithms. In order to obtain a set of representative and diverse measures, we select the measures from the dendrogram shown in Fig.~\ref{fig:measures} following four principles: 1. The distance between the clusters of any two measures in the set is greater than 0.1 (indicated by the red dash line in Fig.~\ref{fig:measures}); 2. Each measure in the set comes from a different syntactic family summarized in \cite{Cha07}; 3. The calculation of the selected measures is not prone, except for some extreme cases, to unstable situations such as division by zero or logarithm of zero; 4. The selected measures are relatively widely used in the related areas. As shown in Fig.~\ref{fig:measures}, six measures are finally selected in this way, i.e., Chebyshev distance (Cheb), Clark distance (Clark), Canberra metric (Canber), Kullback-Leibler divergence (KL-div), cosine coefficient (Cosine), and intersection similarity (Intersec), which belong to the Minkowski family, the $\chi^2$ family, the $L_1$ family, the Shannon's entropy family, the inner product family, and the intersection family, respectively \cite{Cha07}. The first four are distance measures and the last two are similarity measures. Suppose the real label distribution is $D = \{d_1, d_2, \dots, d_c\}$, the predicted label distribution is $\hat{D} = \{\hat{d}_1, \hat{d}_2, \dots, \hat{d}_c\}$, then the formulae of the six measures are summarized in the lower-right corner of Fig.~\ref{fig:measures}, where the ``$\downarrow$'' after the distance measures indicates ``the smaller the better'', and the ``$\uparrow$'' after the similarity measures indicates ``the larger the better''. Considering that the cluster distance between any two selected measures is greater than 0.1, and they all come from different families, the selected measures are significantly different in both syntax and semantics. Thus, they have a good chance, as will be verified in Section~\ref{sect:exp_result}, to reflect different aspects of an LDL algorithm.

\subsection{Datasets}
\label{sect:datasets}
There are in total 16 datasets used in the experiments including an artificial toy dataset and 15 real-world datasets\footnote{The datasets and the Matlab code of the LDL algorithms are available at our web site: http://cse.seu.edu.cn/PersonalPage/xgeng/LDL/index.htm}. Some basic statistics about these 16 datasets are given in Table~\ref{table:datasets}.
\begin{table}[tb]
\scriptsize
\newcommand{\tabincell}[2]{\begin{tabular}{@{}#1@{}}#2\end{tabular}}
\begin{center}
\caption{Statistics of the 16 Datasets Used in the Experiments} \label{table:datasets}
\begin{tabular}{llccc}
\hline \hline \noalign{\smallskip} %
No. & Dataset & \# Examples  ($n$) & \# Features ($q$) & \# Labels ($c$) \\ \hline \noalign{\smallskip}
1 & Artificial & \tabincell{l}{500 (train)\\40,401 (test)} & 3 & 3\\\hline\noalign{\smallskip}
2 & Yeast-alpha & 2,465 & 24 & 18\\
3 & Yeast-cdc & 2,465 & 24 & 15\\
4 & Yeast-elu & 2,465 & 24 & 14\\
5 & Yeast-diau & 2,465 & 24 & 7\\
6 & Yeast-heat & 2,465 & 24 & 6\\
7 & Yeast-spo & 2,465 & 24 & 6\\
8 & Yeast-cold & 2,465 & 24 & 4\\
9 & Yeast-dtt & 2,465 & 24 & 4\\
10 & Yeast-spo5 & 2,465 & 24 & 3\\
11 & Yeast-spoem & 2,465 & 24 & 2\\
12 & Human Gene & 30,542 & 36 & 68\\
13 & Natural Scene & 2,000 & 294 & 9\\
14 & SJAFFE & 213 & 243 & 6\\
15 & SBU\_3DFE & 2,500 & 243 & 6\\
16 & Movie & 7,755 & 1,869 & 5\\
\noalign{\smallskip} \hline \hline
\end{tabular}
\end{center}
\vspace{-7mm}
\end{table}

The first dataset is an artificial toy dataset which is generated to show in a direct and visual way whether the LDL algorithms can learn the mapping from the instance to the label distribution. In this dataset, the instance $\bm{x}$ is of three-dimensional and there are three labels, i.e., $q=3$ and $c=3$. The label distribution $D=\{d_{\bm{x}}^{y_1}, d_{\bm{x}}^{y_2}, d_{\bm{x}}^{y_3}\}$ of an instance $\bm{x} = [x_1, x_2, x_3]^{\text{T}}$ is created in the following way.
\begin{eqnarray}
t_i &=& ax_i + bx_i^2 + cx_i^3 + d, i=1,\dots,3,\label{eq:t1}\\
\psi_1 &=& (\bm{w}_1^{\text{T}}\bm{t})^2,\\
\psi_2 &=& (\bm{w}_2^{\text{T}}\bm{t}+\lambda_1\psi_1)^2,\label{eq:psi_2}\\
\psi_3 &=& (\bm{w}_3^{\text{T}}\bm{t}+\lambda_2\psi_2)^2,\label{eq:psi_3}\\
d_{\bm{x}}^{y_i} &=& \frac{\psi_i}{\psi_1 + \psi_2 + \psi_3}, i=1,\dots,3,\label{eq:y_3}
\end{eqnarray}
where $\bm{t} = [t_1, t_2, t_3]^{\text{T}}$. Note that Eq.~\eqref{eq:psi_2} and \eqref{eq:psi_3} deliberately make the description degree of one label depend on those of other labels. The parameters in Eq.~\eqref{eq:t1}-\eqref{eq:psi_3} are set as $a=1$, $b=0.5$, $c=0.2$, $d=1$, $\bm{w}_1=[4,2,1]^{\text{T}}$, $\bm{w}_2=[1,2,4]^{\text{T}}$, $\bm{w}_3=[1,4,2]^{\text{T}}$, and $\lambda_1=\lambda_2 = 0.01$. To generate the training set, each component of $\bm{x}$ is uniformly sampled within the range $[-1, 1]$. In total, there are $500$ instances sampled in this way. Then, the label distribution corresponding to each instance is calculated via Eq.~\eqref{eq:t1}-\eqref{eq:y_3}. Such $500$ examples are used as the training set for the LDL algorithms proposed in Section~\ref{sect:algorithms}.

In order to show the result of the LDL algorithms in a direct and visual way, the test examples of the toy dataset are selected from a certain manifold in the instance space. The first two components of the test instance $\bm{x}$, $x_1$ and $x_2$, are located at a grid of the interval $0.01$ within the range $[-1, 1]$ on both dimensions, i.e., there are in total $201\times201=40,401$ test instances. The third component $x_3$ is calculate by
\begin{equation}
x_3=\sin((x_1+x_2)\times \pi).
\end{equation}
Then, the label distribution of each test instance, either the ground-truth calculated via Eq.~\eqref{eq:t1}-\eqref{eq:y_3} or the prediction given by the LDL algorithms, is transformed into a color (details in Section~\ref{sect:exp_art}). Thus the ground-truth and predicted label distributions of the test instances can be compared visually through the color pattern on the manifold.

The second to the eleventh datasets (from \emph{Yeast-alpha} to \emph{Yeast-spoem}) are real-world datasets collected from ten biological experiments on the budding yeast \emph{Saccharomyces cerevisiae} \cite{EisenSBB98}. Each dataset records the result of one experiment. There are in total $2,465$ yeast genes included, each of which is represented by an associated phylogenetic profile vector of the length $24$. For each dataset, the labels correspond to the discrete time points during one biological experiment. The gene expression level (after normalization) at each time point provides a natural measure of the description degree of the corresponding label. The number of labels in the ten Yeast Gene datasets is summarized in Table~\ref{table:datasets}. The description degrees (normalized gene expression levels) of all the labels (time points) constitute a label distribution for a particular yeast gene.

The twelfth dataset \emph{Human Gene} is a large-scale real-world dataset collected from the biological research on the relationship between human genes and diseases. There are in total $30,542$ human genes included in this dataset, each of which is represented by the $36$ numerical descriptors for a gene sequence proposed in \cite{YuJX12}. The labels correspond to $68$ different diseases. The gene expression level (after normalization) for each disease is regarded as the description degree of the corresponding label. The description degrees (normalized gene expression level) of all the $68$ labels (diseases) constitute a label distribution for a particular human gene.

The thirteenth dataset \emph{Natural Scene} results from the inconsistent multilabel rankings of $2,000$ natural scene images. There are nine possible labels associated with these images, i.e., plant, sky, cloud, snow, building, desert, mountain, water, and sun. Ten human rankers are requested to label the images. For each image, they first select from the nine candidate labels what they think are relevant to the image, and then rank the relevant labels in descending order of relevance to the image. Each human ranker makes his/her decisions independently, and the resulting multilabel rankings are expectably highly inconsistent. Then, the inconsistent rankings for each image are transformed into a label distribution by a nonlinear programming process \cite{GengL14}, which finds the common label distribution that is most compatible with all personal rankings. Finally, for each image, a $294$-dimensional feature vector is extracted by the method proposed in \cite{BoutellLSB04}.

The fourteenth and fifteenth datasets are extensions of two widely used facial expression image databases, i.e., JAFFE \cite{LyonsAKG98} and BU\_3DFE \cite{YinWSWR06}. The JAFFE database contains $213$ grayscale expression images posed by 10 Japanese female models. A $243$-dimensional feature vector is extracted from each image by the method of Local Binary Patterns (LBP) \cite{AhonenHP06}. Each image is scored by $60$ persons on the $6$ basic emotion labels (i.e., happiness, sadness, surprise, fear, anger, and disgust) with a 5-level scale. The average score of each emotion is used to represent the emotion intensity. Instead of only considering the emotion with the highest score as most work on JAFFE does, the dataset \emph{SJAFFE} (Scored JAFFE) used in this paper keeps all the scores and normalizes them into a label distribution over all the $6$ emotion labels. Similarly, for the bigger database BU\_3DFE containing $2,500$ facial expression images, each image is scored by $23$ persons in the same way JAFFE is scored, resulting in the label distribution version of the dataset \emph{SBU\_3DFE} (Scored BU\_3DFE).

Finally, the sixteenth dataset is about the user ratings on movies. The dataset includes $7,755$ movies and $54,242,292$ ratings from $478,656$ different users. The ratings come from Netflix, which are on a scale from 1 to 5 integral stars (5 labels). The rating label distribution is calculated for each movie as the percentage of each rating level. The features of the movie are extracted from the metadata such as genre, director, actor, country, budget, etc. Categorical attributes are transformed into binary vectors. The final feature vector extracted from each movie is of $1,869$-dimensional.

\subsection{Methodology}

All the six algorithms described in Section~\ref{sect:algorithms}, i.e., PT-Bayes, PT-SVM, AA-$k$NN, AA-BP, SA-IIS, and SA-BFGS, are applied to the 16 datasets shown in Table~\ref{table:datasets} and compared by the six measures listed in Fig.~\ref{fig:measures}. On the 15 real-world datasets, ten-fold cross validation is conducted for each algorithm and the mean value and standard deviation of each evaluation measure are recorded.

For each algorithm, several parameter configurations are tried. On the artificial dataset, one fifth of the training set is randomly selected as the validation set. The model is trained on those examples left in the training set and tested on the validation set to select the best parameters. Then, the model is trained on the whole training set with the best parameters and tested on the test set. On the real-world datasets, the parameter selection process is nested into the ten-fold cross validation. In detail, the whole data is first randomly split into 10 chunks. Each time, one chunk is used as test set, another is used as validation set, and the rest 8 chunks are used as training set. Then, the model is trained with different parameter settings on the training set and tested on the validation set. This procedure is repeated 10 times (each time with different training and validation sets), and the parameter setting with the best average performance is selected. After that, the original validation set is merged into the training set and the test set remains unchanged. The model is trained with the selected parameter setting on the updated training set and tested on the test set. This procedure is repeated 10 times and the average performance is recorded.

\begin{figure}[tb]%
\centering%
\subfigure[][Ground-Truth]{\label{fig:aritficial:ground}\includegraphics[width = .5\linewidth]{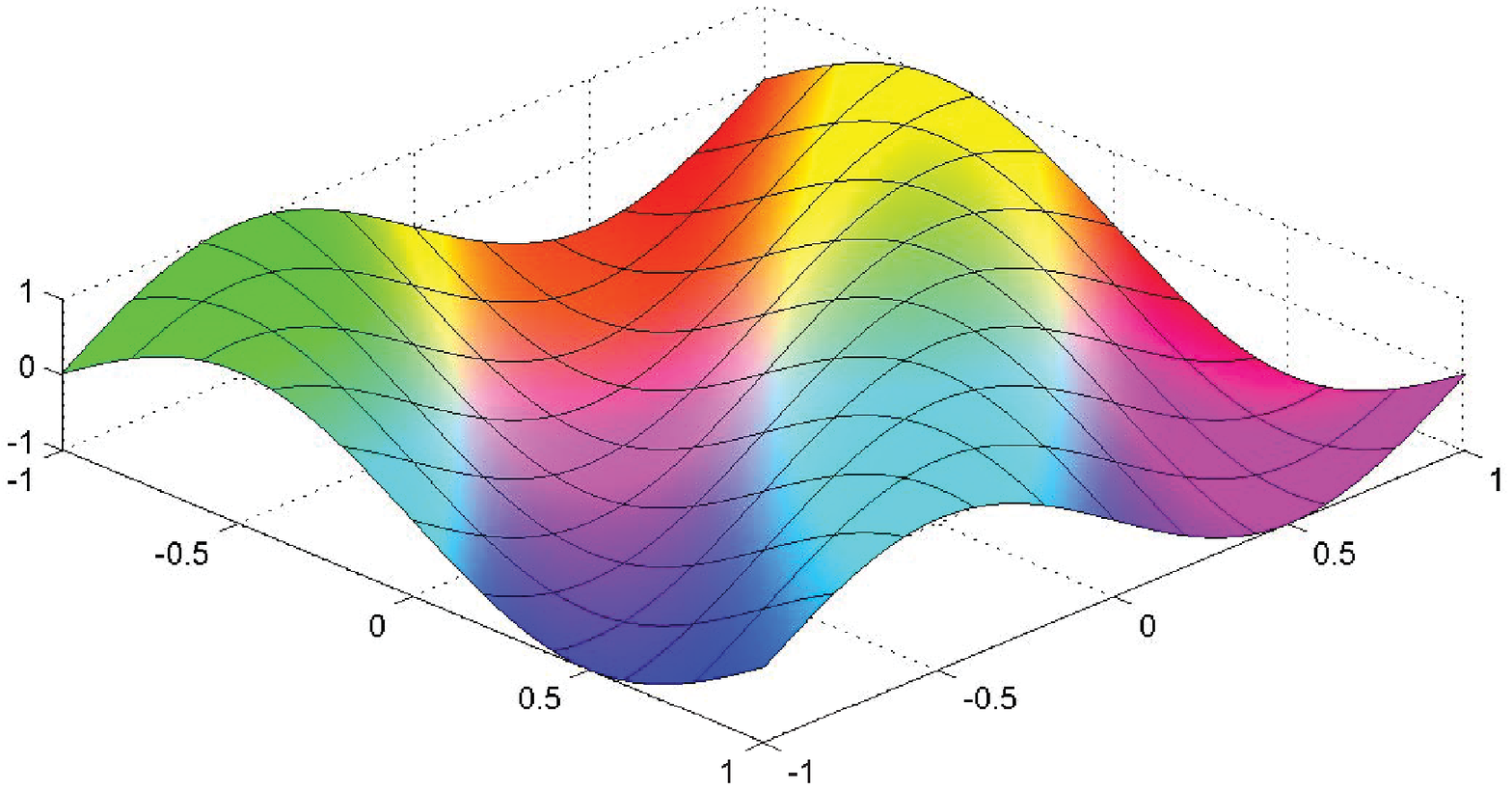}}\vspace{-3mm}\\%
\subfigure[][PT-Bayes]{\label{fig:aritficial:bayes}\includegraphics[width = .5\linewidth]{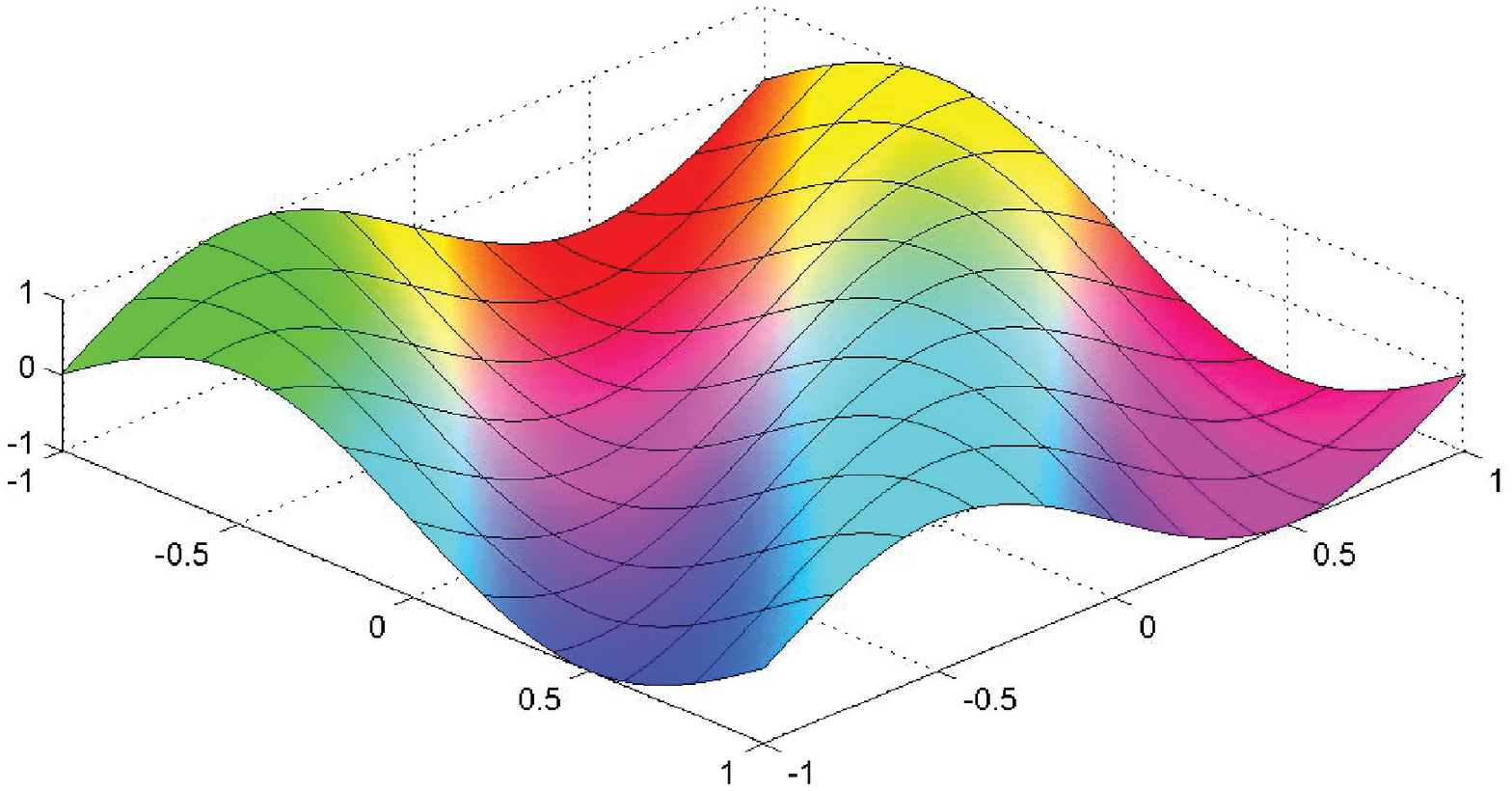}}
\subfigure[][PT-SVM]{\label{fig:aritficial:svm} \includegraphics[width = .5\linewidth]{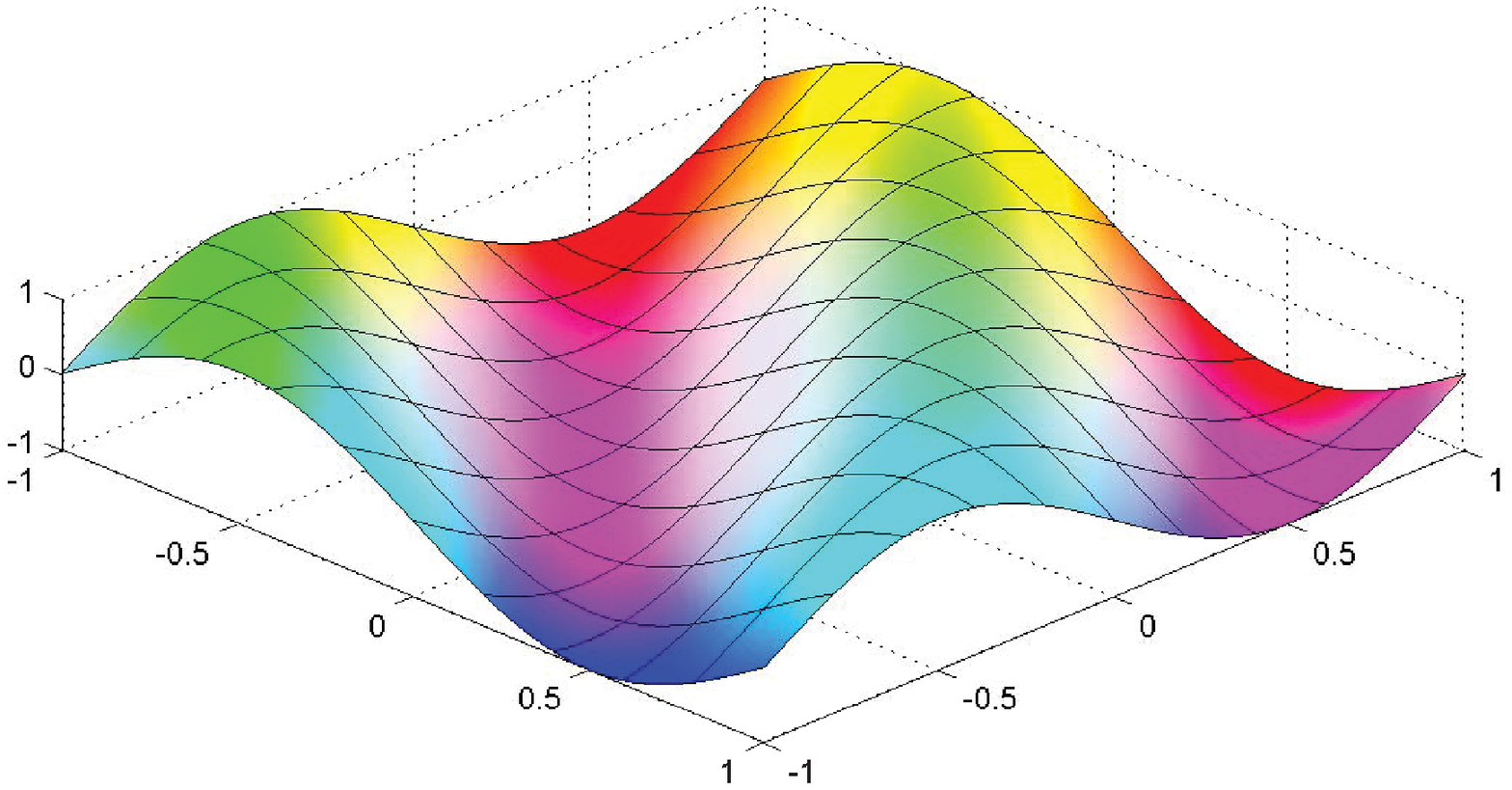}}\vspace{-3mm}\\%
\subfigure[][AA-$k$NN]{\label{fig:aritficial:knn} \includegraphics[width = .5\linewidth]{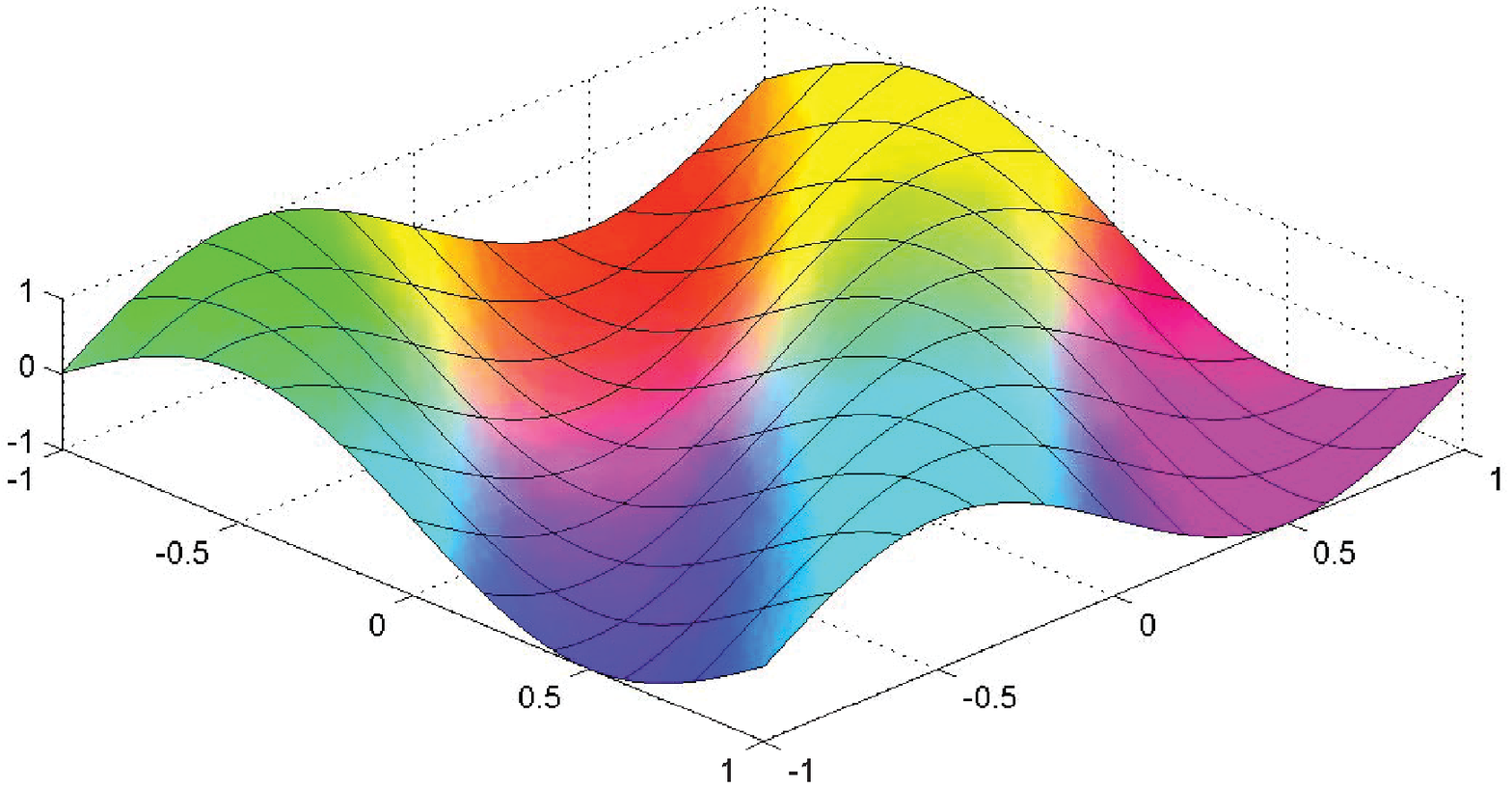}}
\subfigure[][AA-BP]{\label{fig:aritficial:bp} \includegraphics[width = .5\linewidth]{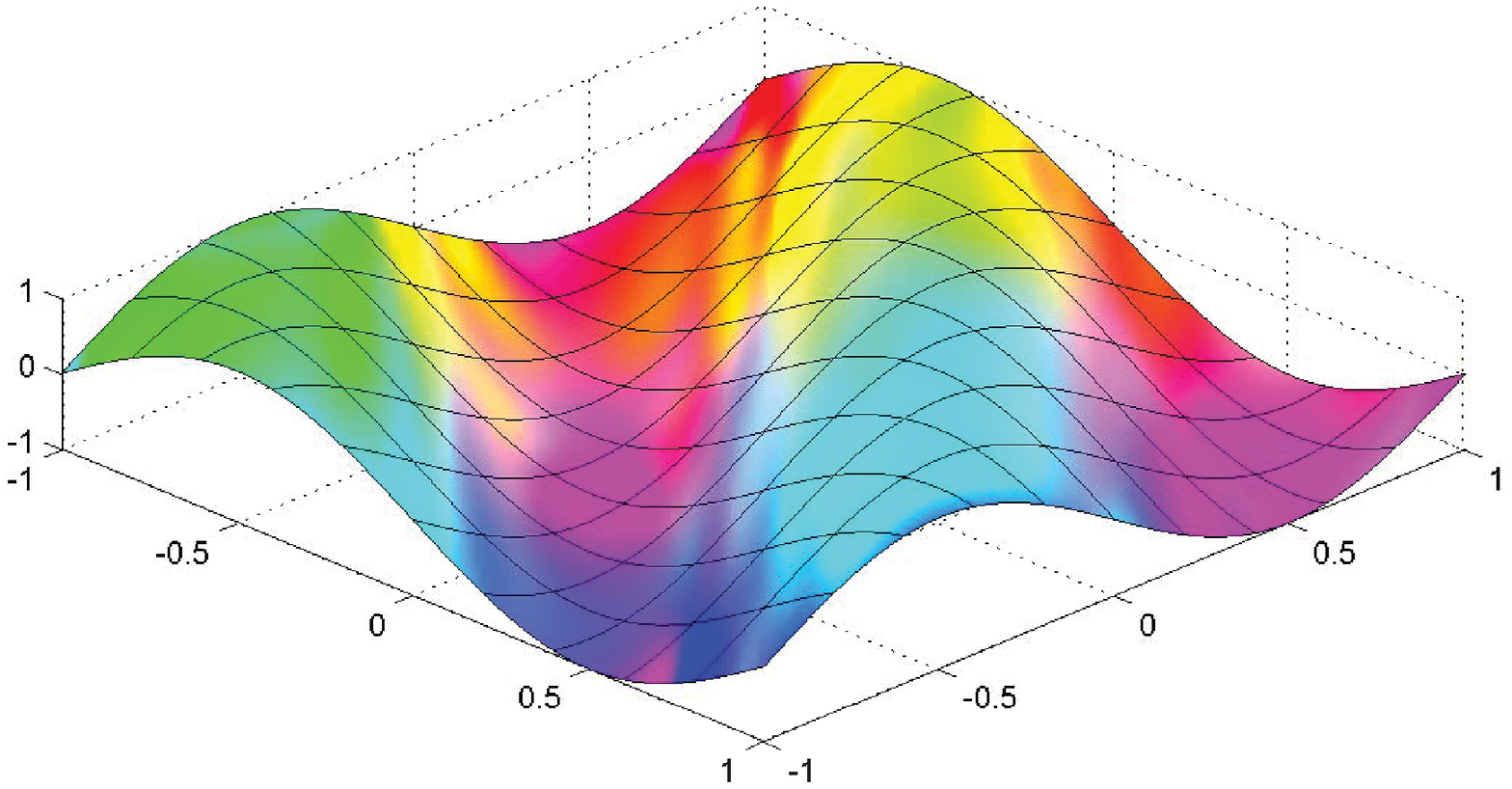}}\vspace{-3mm}\\%
\subfigure[][SA-IIS]{\label{fig:aritficial:iis} \includegraphics[width = .5\linewidth]{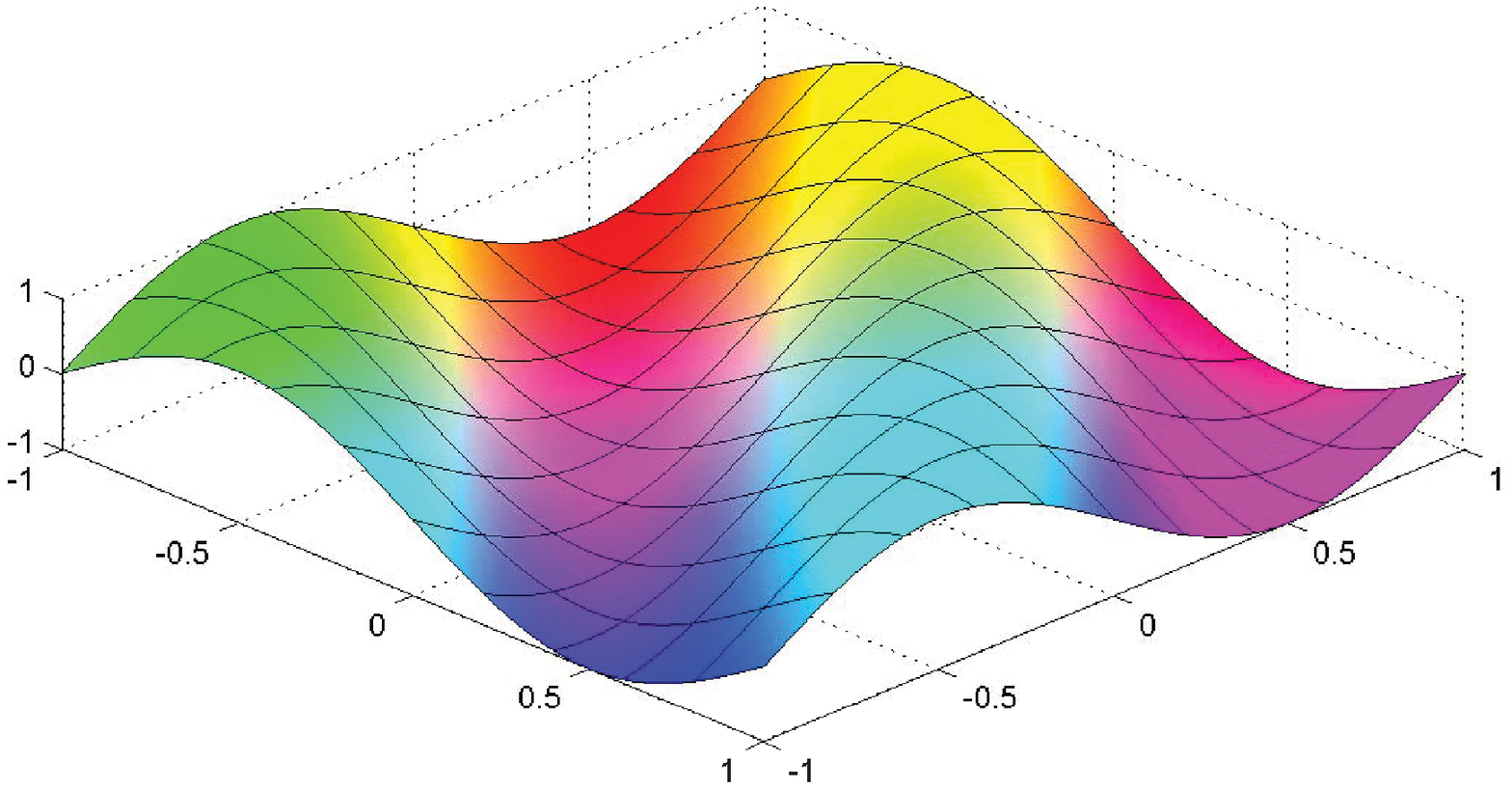}}
\subfigure[][SA-BFGS]{\label{fig:aritficial:bfgs} \includegraphics[width = .5\linewidth]{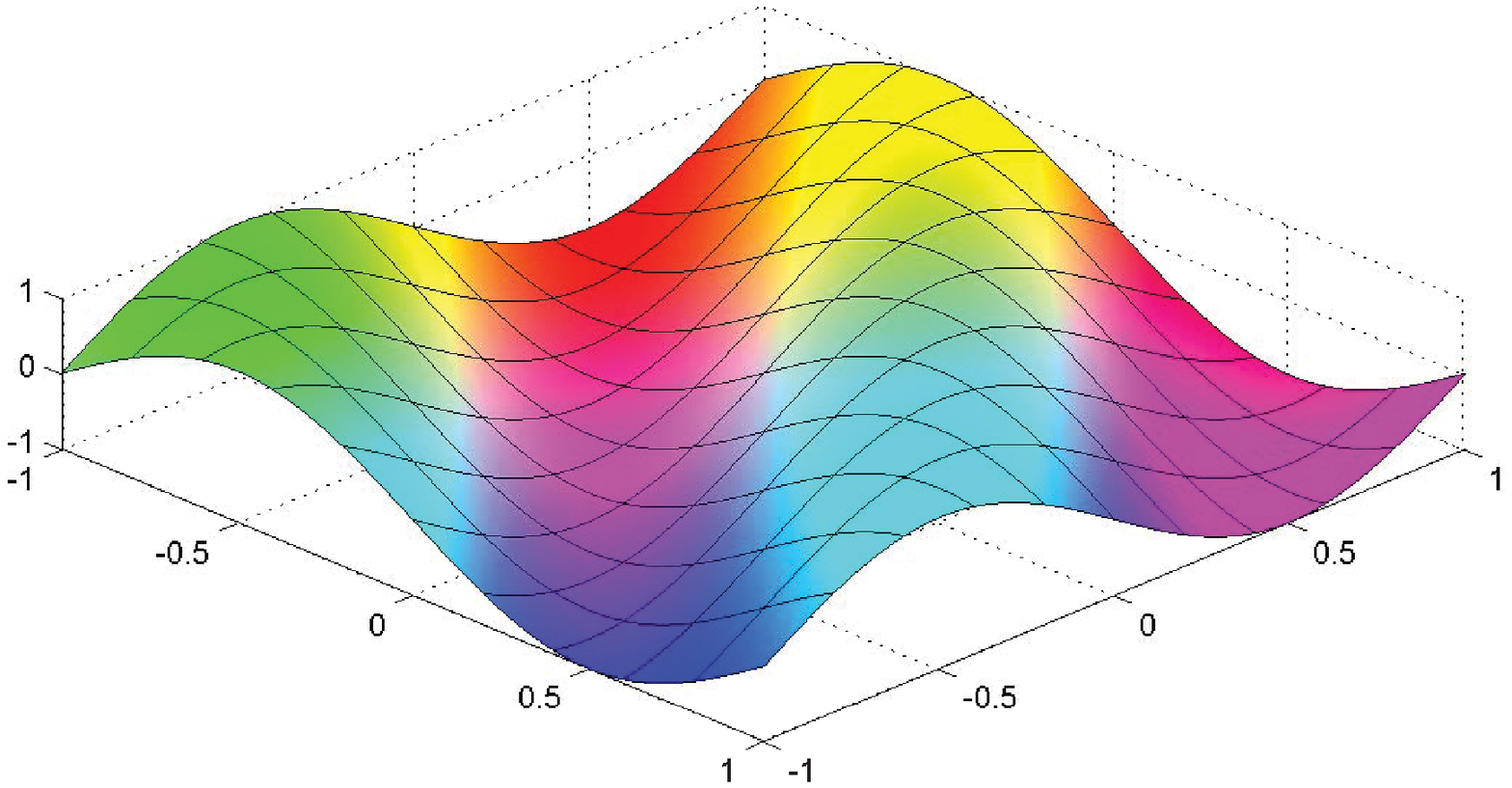}}%
\caption{Comparison between the ground-truth and predicted label distributions (regarded as RGB colors) on the artificial test manifold.}%
\label{fig:aritficial:colors}%
\vspace{-3mm}
\end{figure}

\begin{table*}[tb]
\scriptsize
\begin{center}
\caption{Experimental Results on the Artificial Dataset} \label{table:art_results}
\begin{tabular}{lcccccc}
\hline \hline \noalign{\smallskip} %
Criterion & PT-Bayes & PT-SVM & AA-$k$NN & AA-BP & SA-IIS & SA-BFGS\\\hline \noalign{\smallskip}
Chebyshev $\downarrow$ &0.080(3)&0.653(6)&0.086(4)&0.101(5)&0.0767(2)&\textbf{0.0766(1)}\\
Clark $\downarrow$ &\textbf{0.341(1)}&1.135(6)&0.382(4)&0.520(5)&0.349(2)&0.352(3)\\
Canberra $\downarrow$ &\textbf{0.488(1)}&1.823(6)&0.564(4)&0.699(5)&0.489(2)&0.495(3)\\
Kullback-Leibler $\downarrow$ &0.030(3)&1.482(6)&0.035(4)&0.066(5)&\textbf{0.029(1)}&0.030(2)\\
Cosine $\uparrow$ &0.990(3)&0.377(6)&0.989(4)&0.983(5)&0.99116(2)&\textbf{0.99120(1)}\\
Intersection $\uparrow$ &0.920(3)&0.347(6)&0.914(4)&0.899(5)&0.9233(2)&\textbf{0.9234(1)}\\
\hline\noalign{\smallskip}
Avg. Rank &2.33&6.00&4.00&5.00&1.83&1.83 \\
\hline\noalign{\smallskip}
Running Time (ms) & 22 / 45 & 391 / 1,153 & 0 / 79,961 & 101,568 / 149 & 1,168 / 33 & 187 / 33 \\
\hline \hline
\end{tabular}
\end{center}
\vspace{-6mm}
\end{table*}

\subsection{Results}
\label{sect:exp_result}

\begin{table*}[tb]
\scriptsize
\newcommand{\tabincell}[2]{\begin{tabular}{@{}#1@{}}#2\end{tabular}}
\begin{center}
\caption{Typical Examples of the Real and Predicted Label Distributions} \label{table:predictions}
\begin{tabular}{|c|c|c|c|c|}\hline
\multirow{2}*[6mm]{Real} & \includegraphics[width=1.9cm]{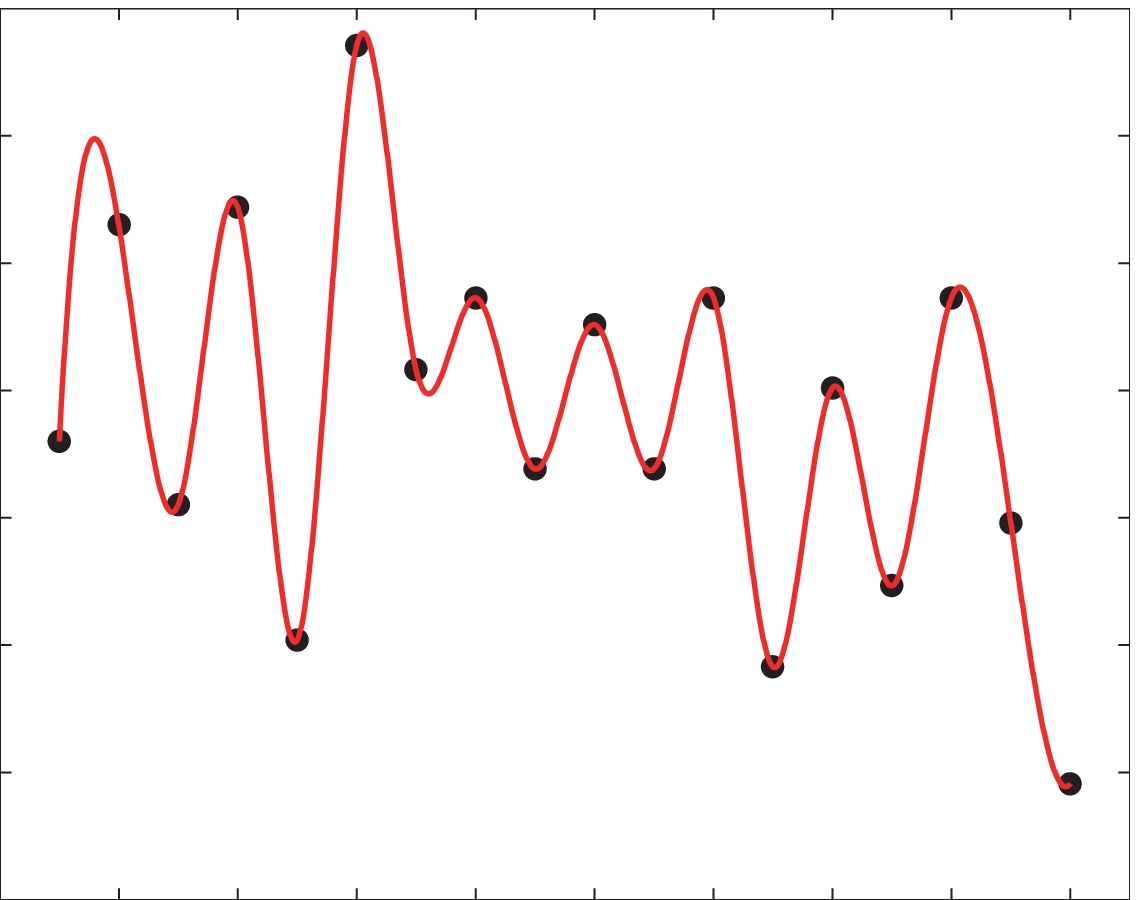}\rule{0cm}{16mm} & \includegraphics[width=1.9cm]{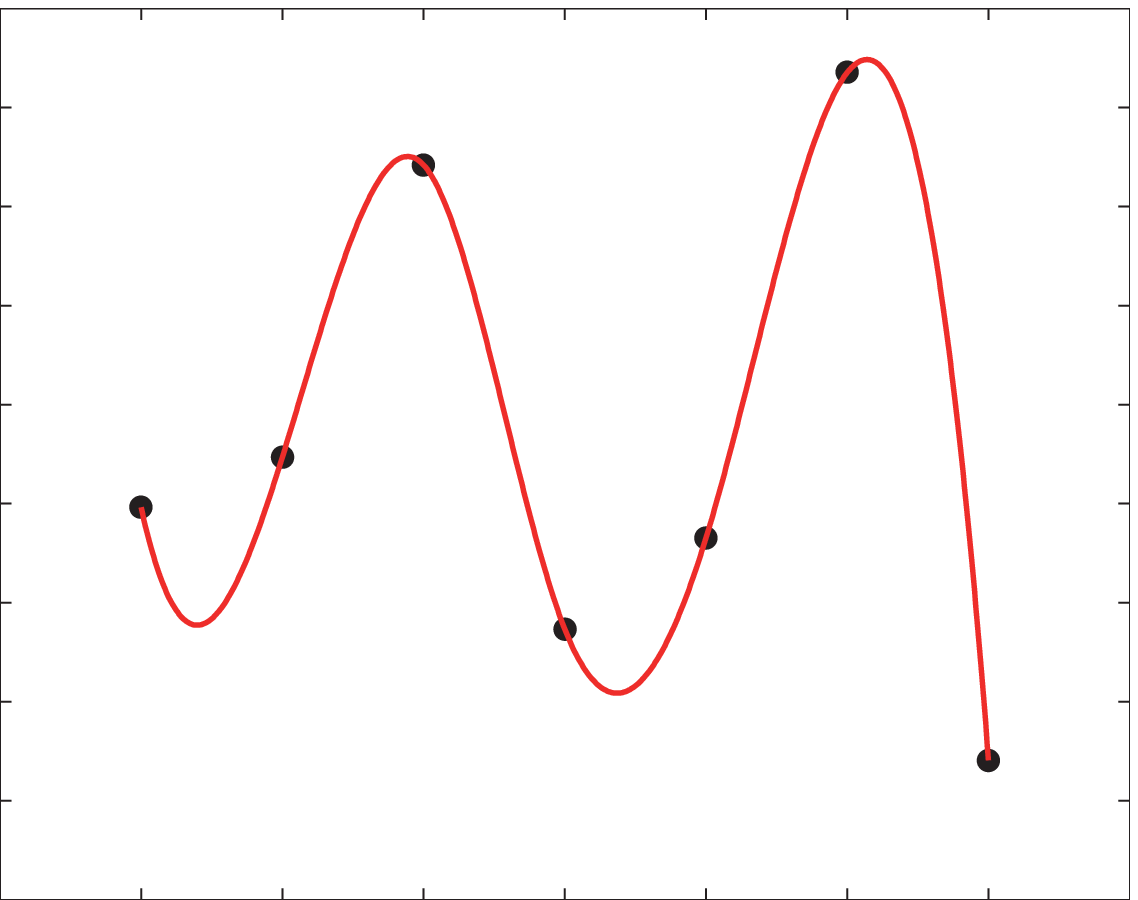} & \includegraphics[width=1.9cm]{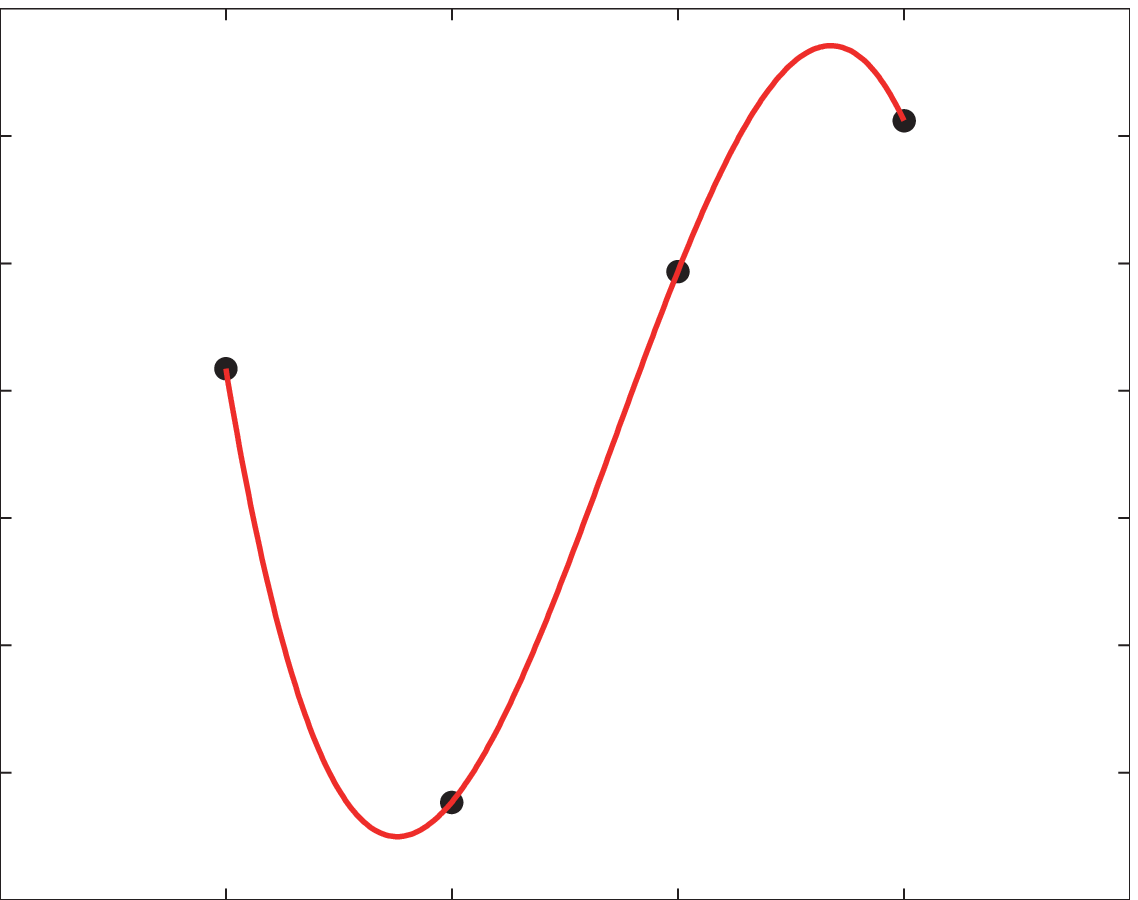} & \includegraphics[width=1.9cm]{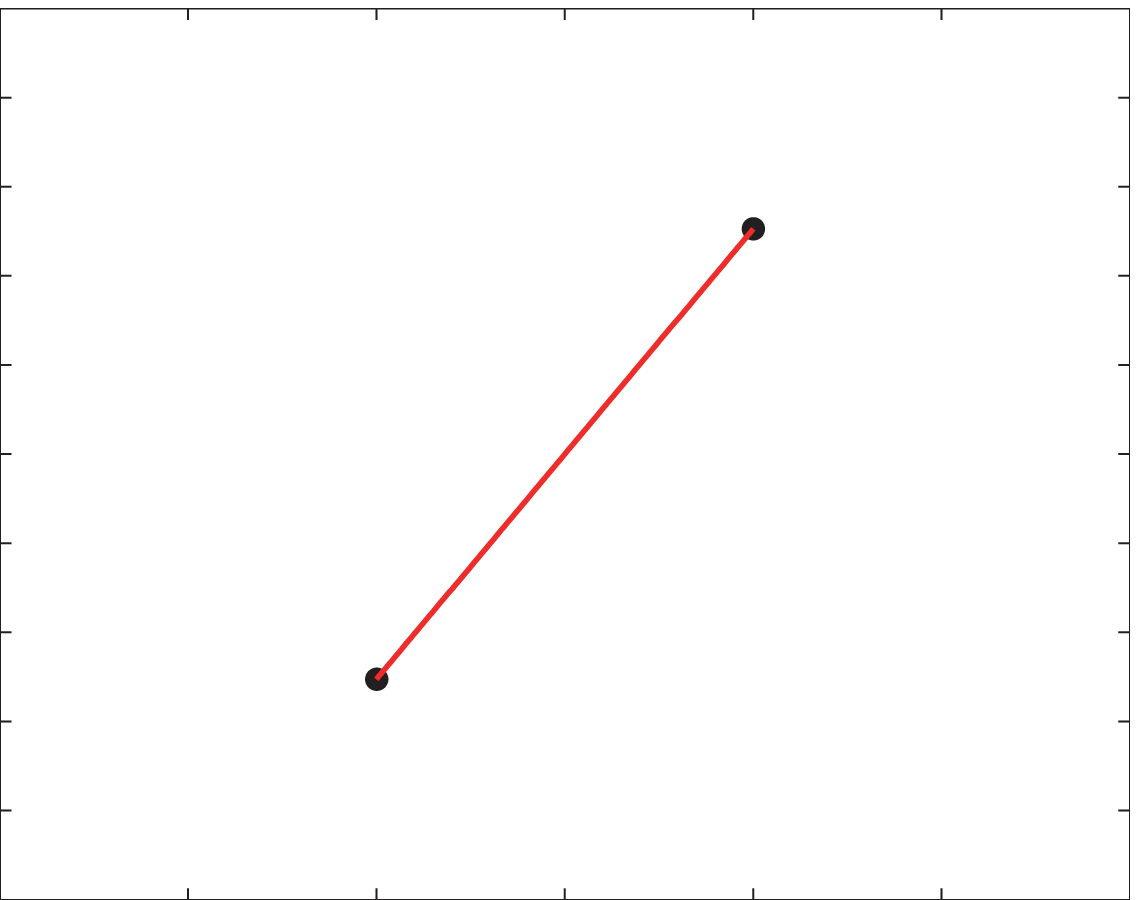}\\
 & 18 Labels & 7 Labels & 4 Labels & 2 Labels\\\hline
\multirow{2}*[5mm]{PT-Bayes} & \includegraphics[width=1.9cm]{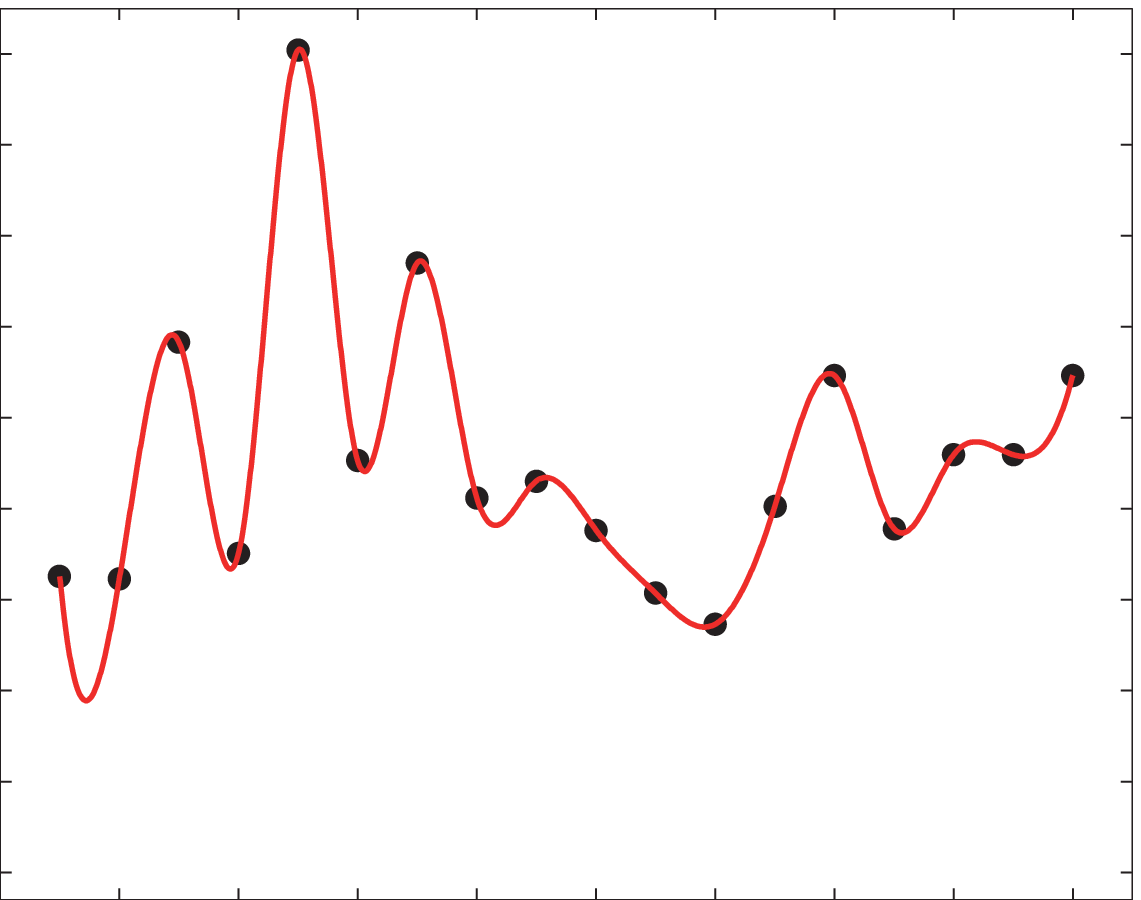}\rule{0cm}{16mm} & \includegraphics[width=1.9cm]{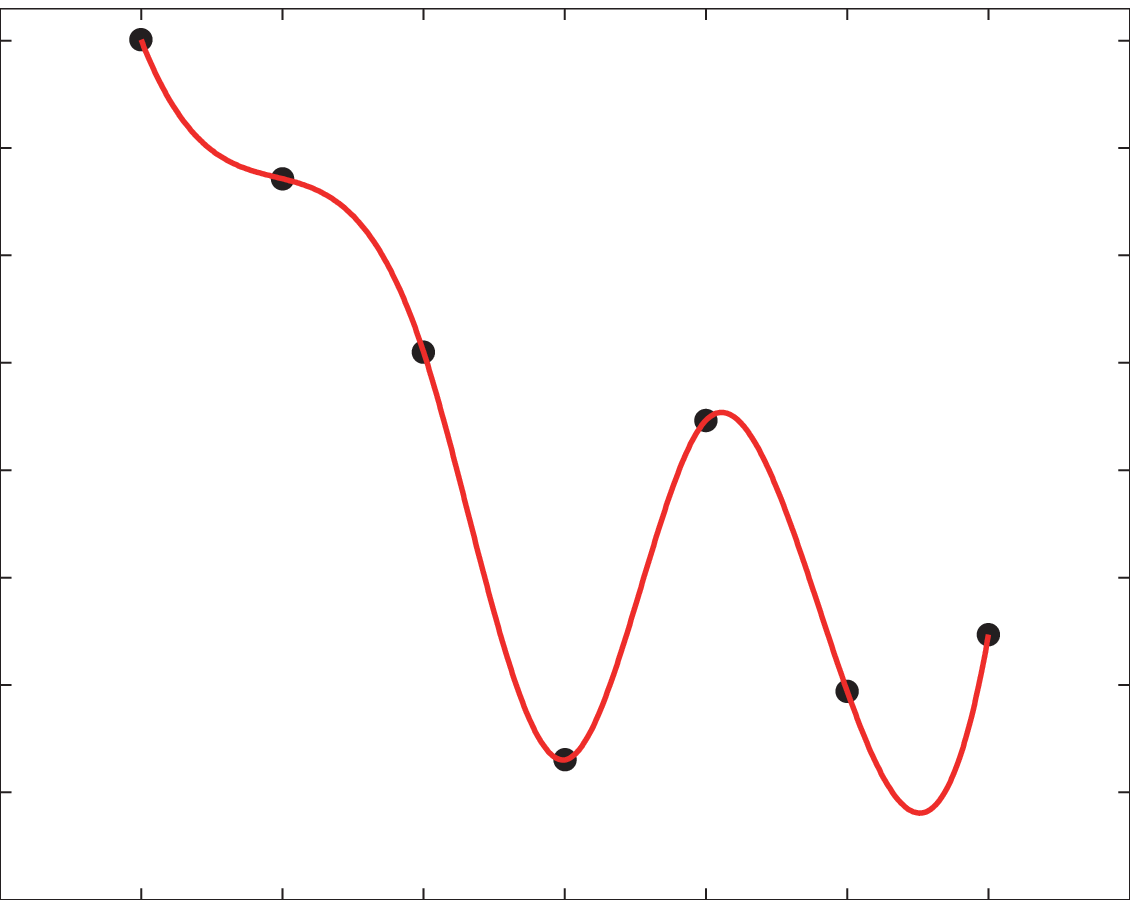} & \includegraphics[width=1.9cm]{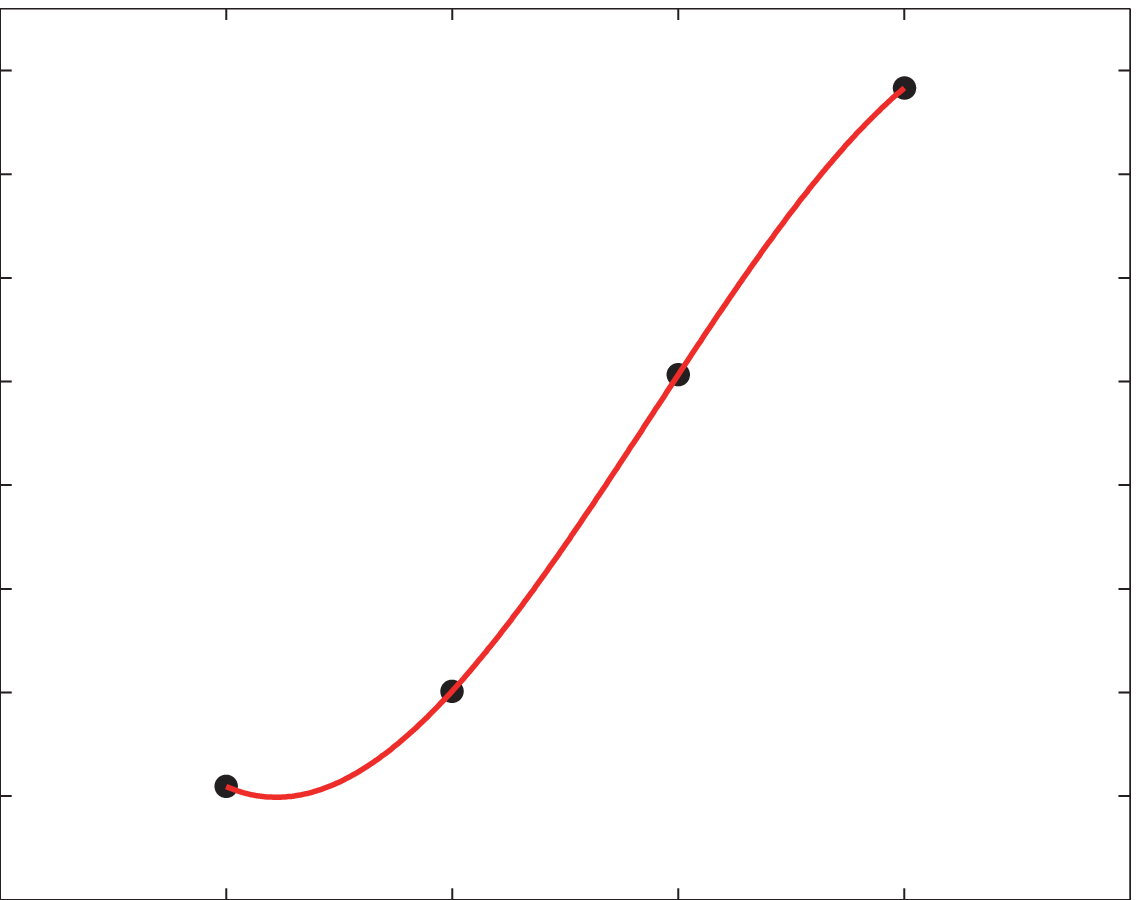} & \includegraphics[width=1.9cm]{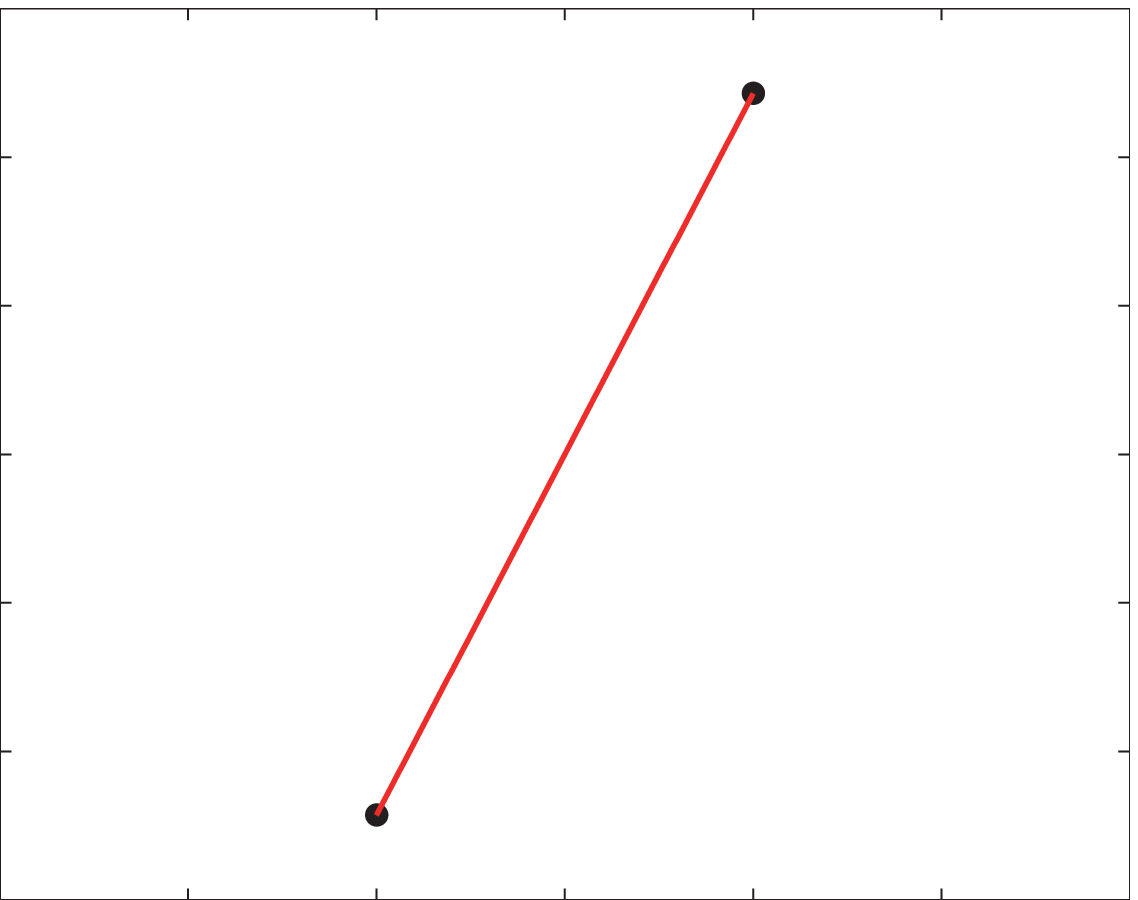}\\
 & \tabincell{l}{\{0.048, 0.552, 1.885, 0.036\} \\ \{0.960	0.891\}} & \tabincell{l}{\{0.042, 0.230, 0.492, 0.016\} \\ \{0.984, 0.929\}} & \tabincell{l}{\{0.170, 0.516, 0.911, 0.128\} \\ \{0.907, 0.786\}} & \tabincell{l}{\{0.192, 0.310, 0.421, 0.086\} \\ \{0.939, 0.808\}}\\\hline
\multirow{2}*[5mm]{PT-SVM} & \includegraphics[width=1.9cm]{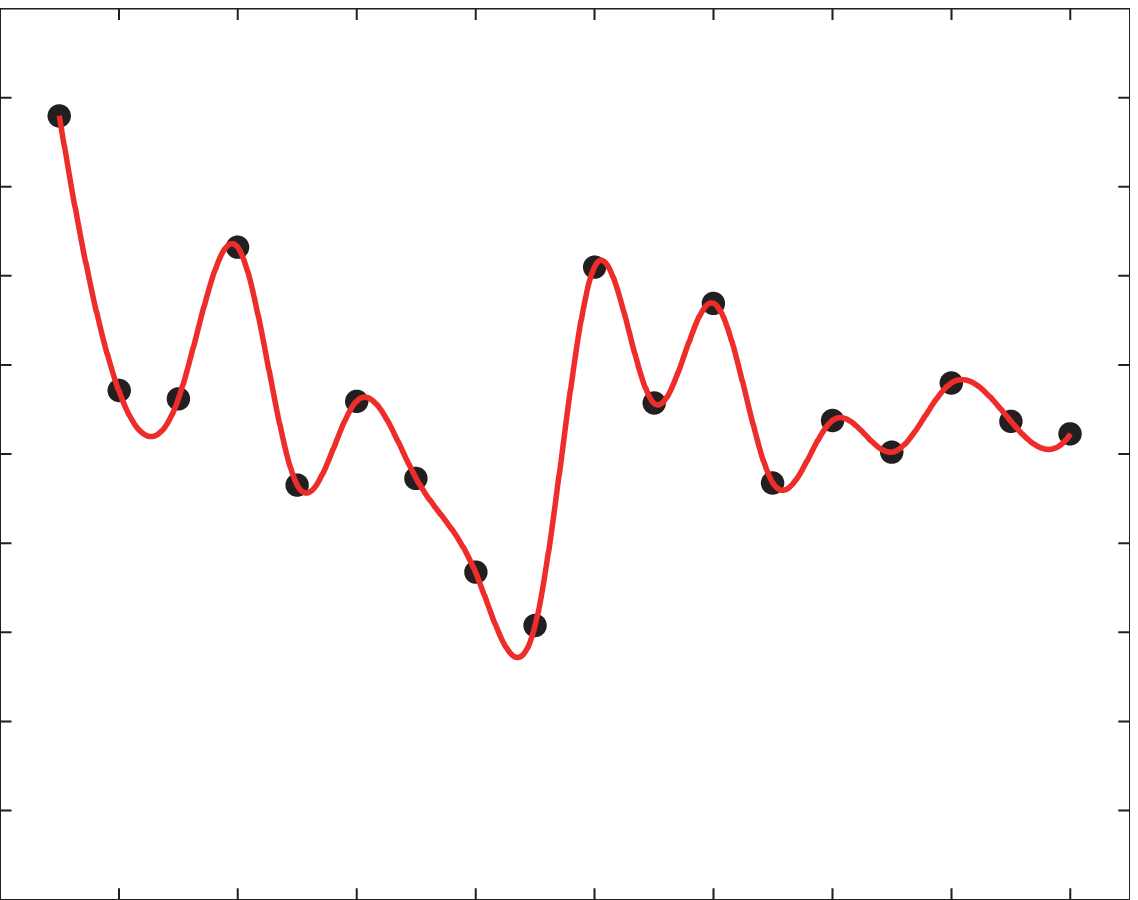}\rule{0cm}{16mm} & \includegraphics[width=1.9cm]{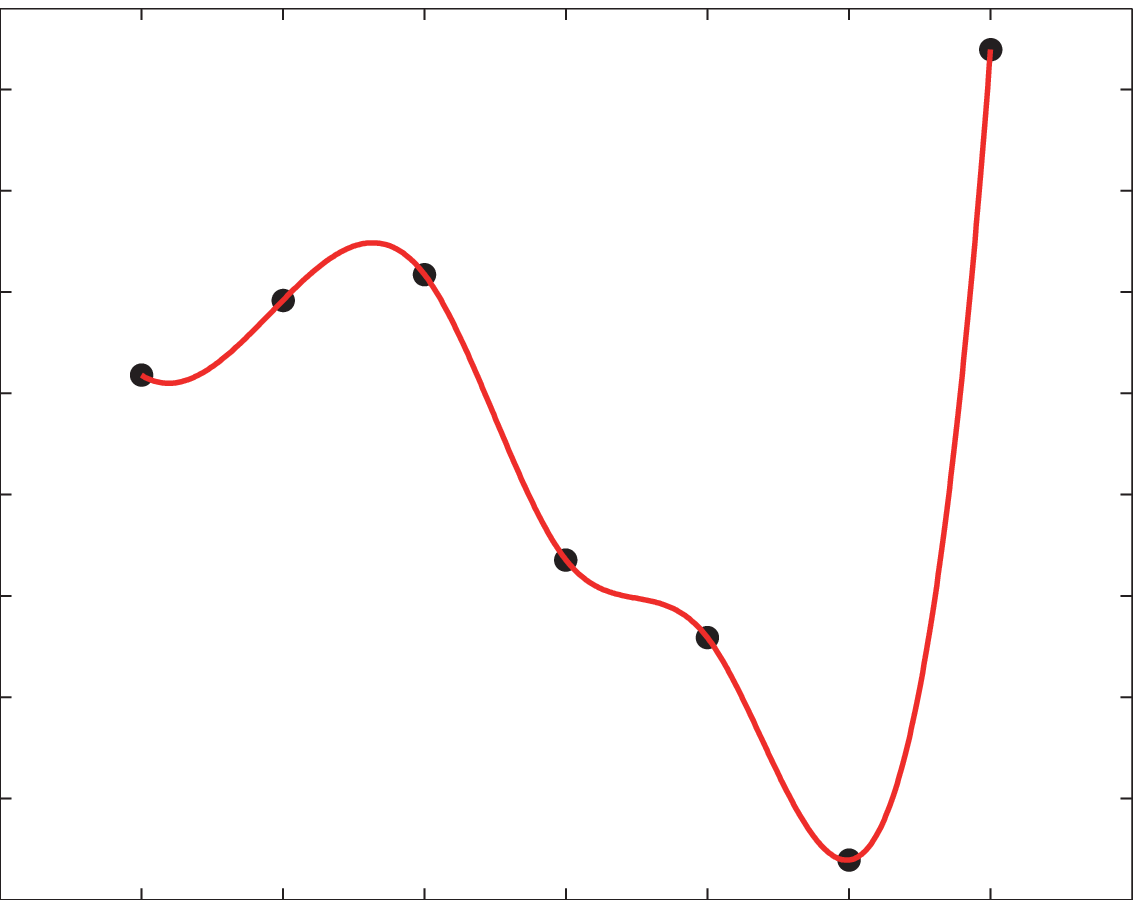} & \includegraphics[width=1.9cm]{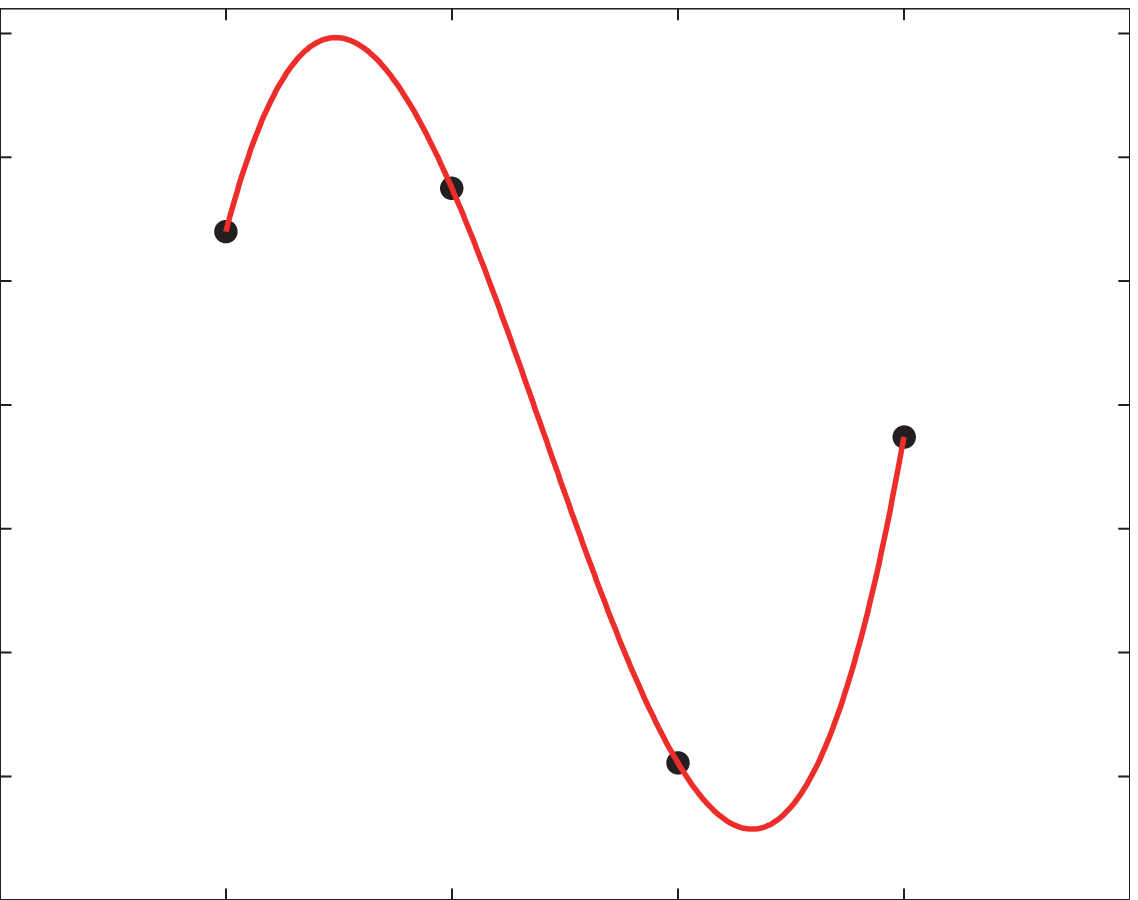} & \includegraphics[width=1.9cm]{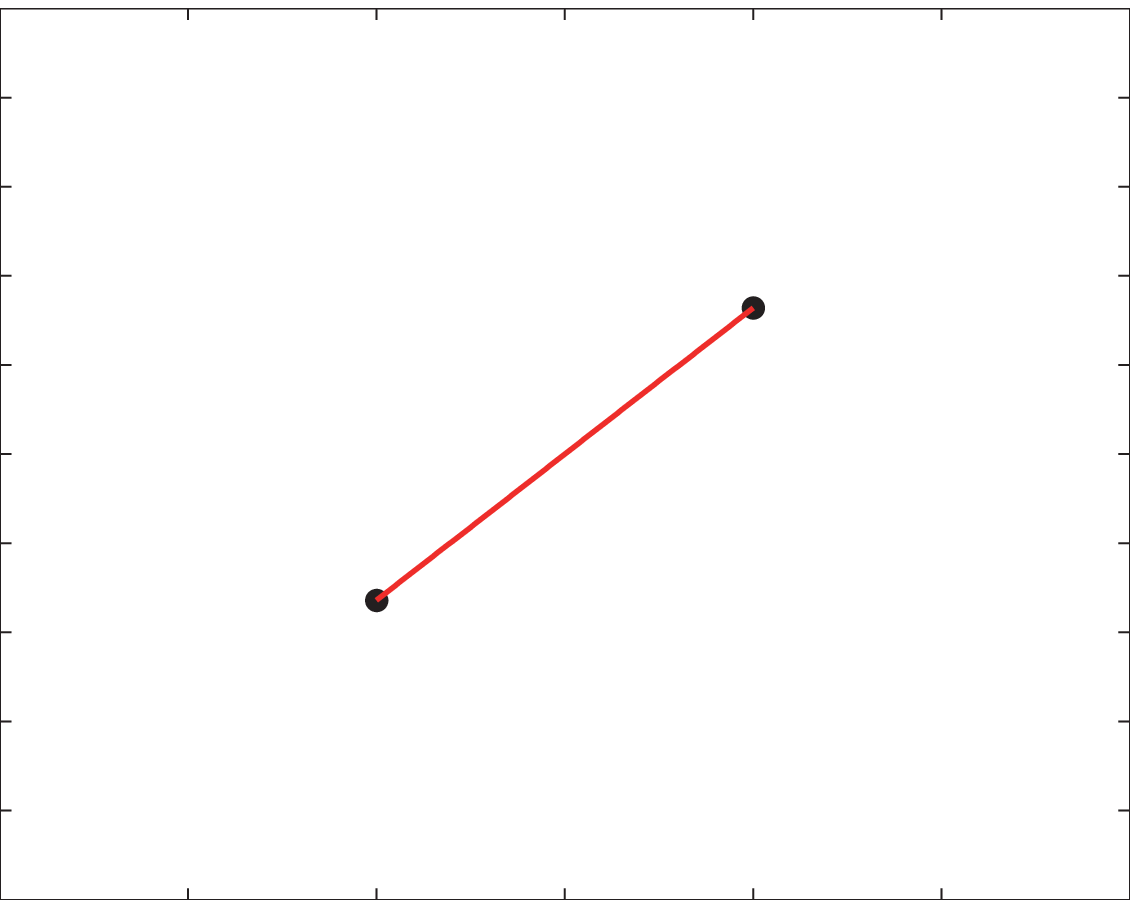}\\
 & \tabincell{l}{\{0.006, 0.105, 0.368, 0.0012\} \\ \{0.9988, 0.980\}} & \tabincell{l}{\{0.040, 0.190, 0.359, 0.011\} \\ \{0.990, 0.949\}} & \tabincell{l}{\{0.041, 0.108, 0.188, 0.006\} \\ \{0.995 0.954\}} & \tabincell{l}{\{0.018, 0.025, 0.036, 0.001\} \\ \{0.999, 0.982\}}\\\hline
\multirow{2}*[5mm]{AA-$k$NN} & \includegraphics[width=1.9cm]{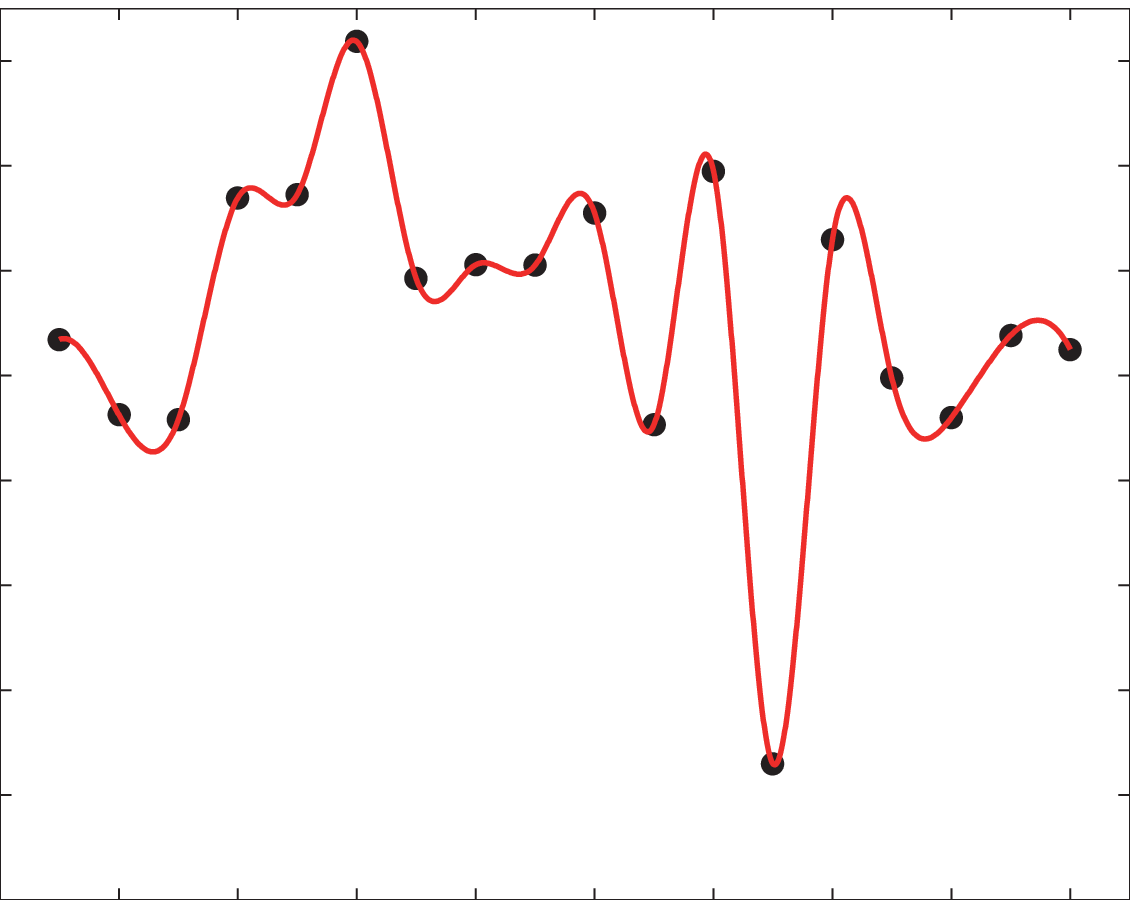}\rule{0cm}{16mm} & \includegraphics[width=1.9cm]{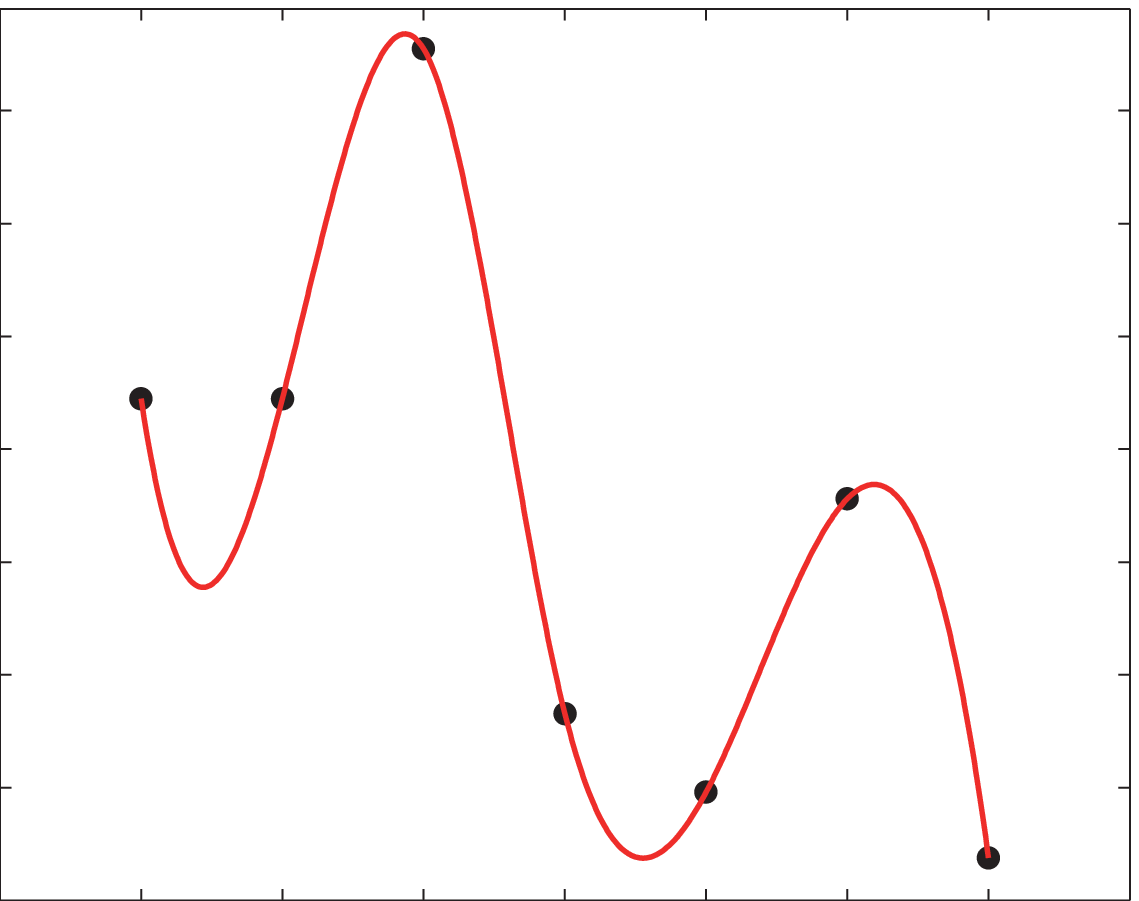} & \includegraphics[width=1.9cm]{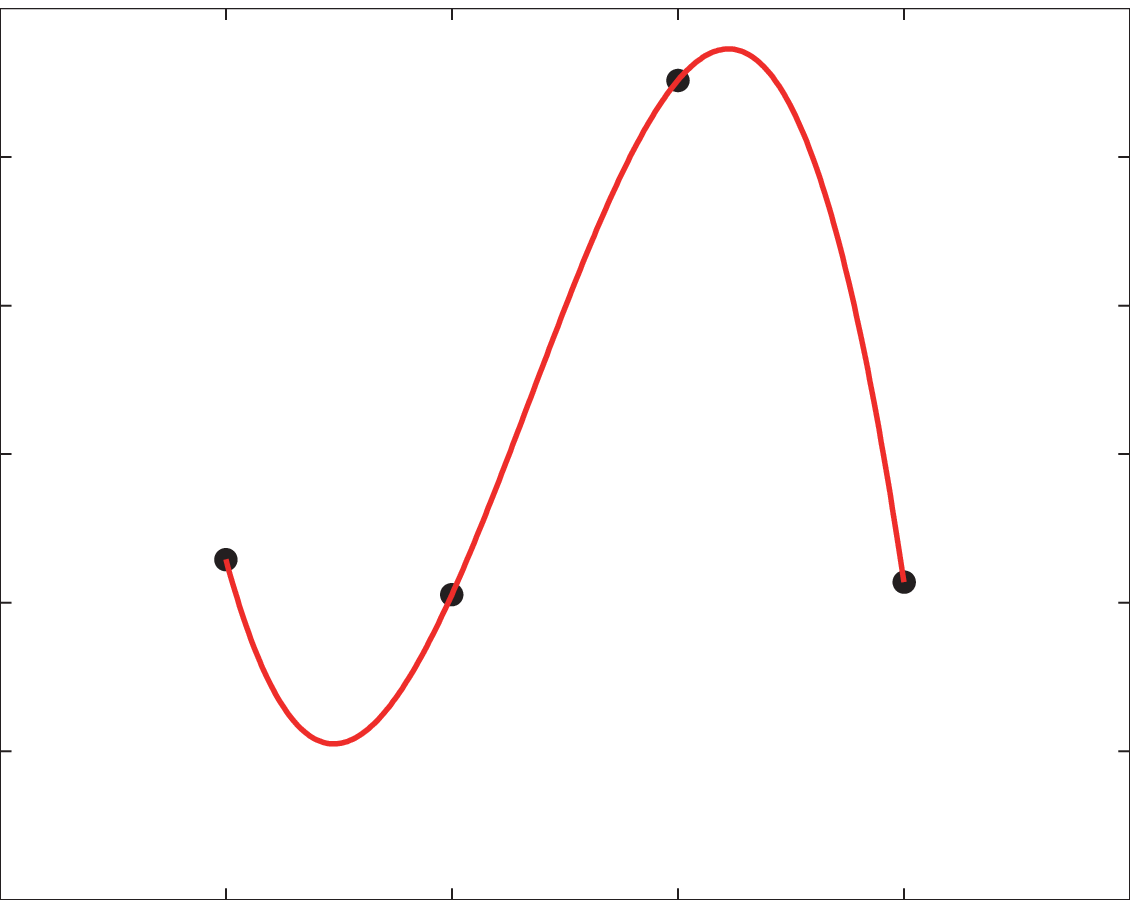} & \includegraphics[width=1.9cm]{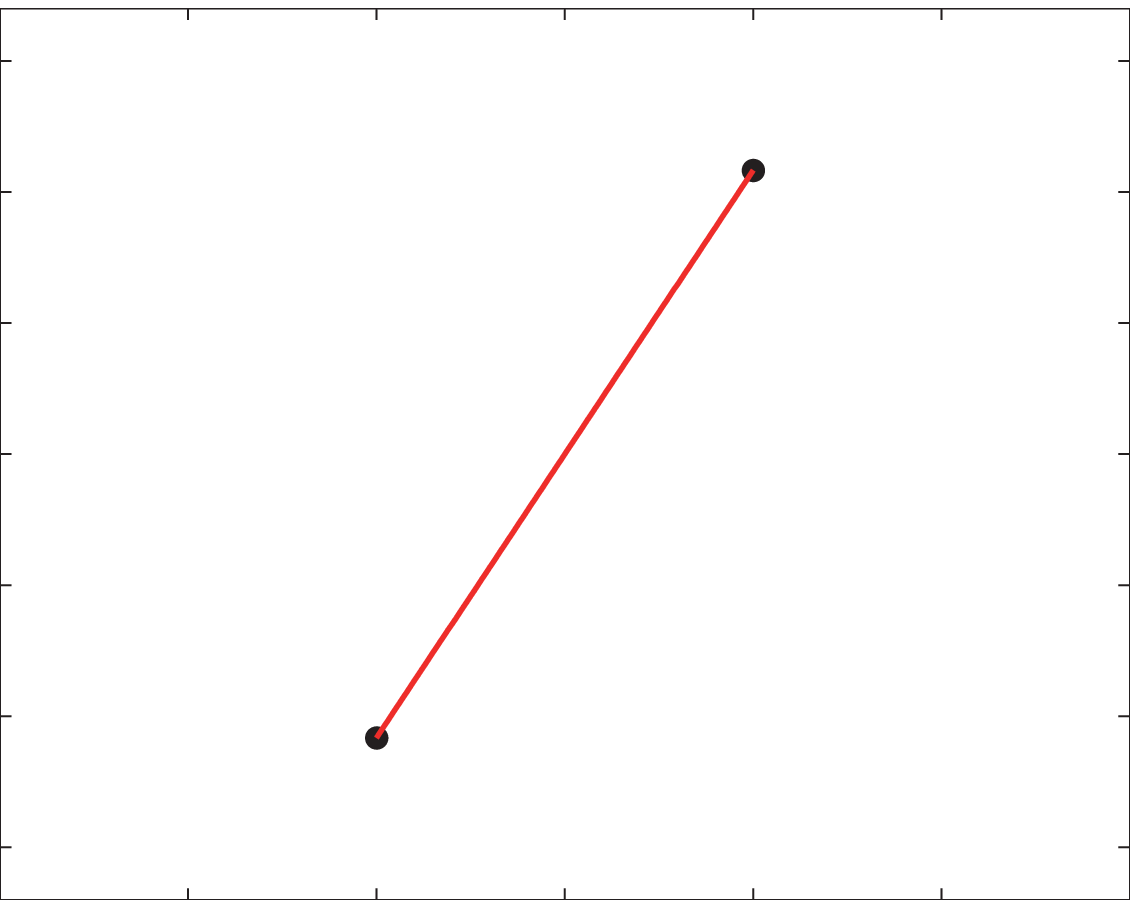}\\
 & \tabincell{l}{\{0.005, 0.091, 0.282, 0.0009\} \\ \{0.9991, 0.9844\}} & \tabincell{l}{\{0.028, 0.151, 0.381, 0.007\} \\ \{0.993, 0.945\}} & \tabincell{l}{\{0.030, 0.080, 0.124, 0.003\} \\ \{0.997, 0.970\}} & \tabincell{l}{\{0.058, 0.085, 0.118, 0.007\} \\ \{0.994, 0.942\}}\\\hline
\multirow{2}*[5mm]{AA-BP} & \includegraphics[width=1.9cm]{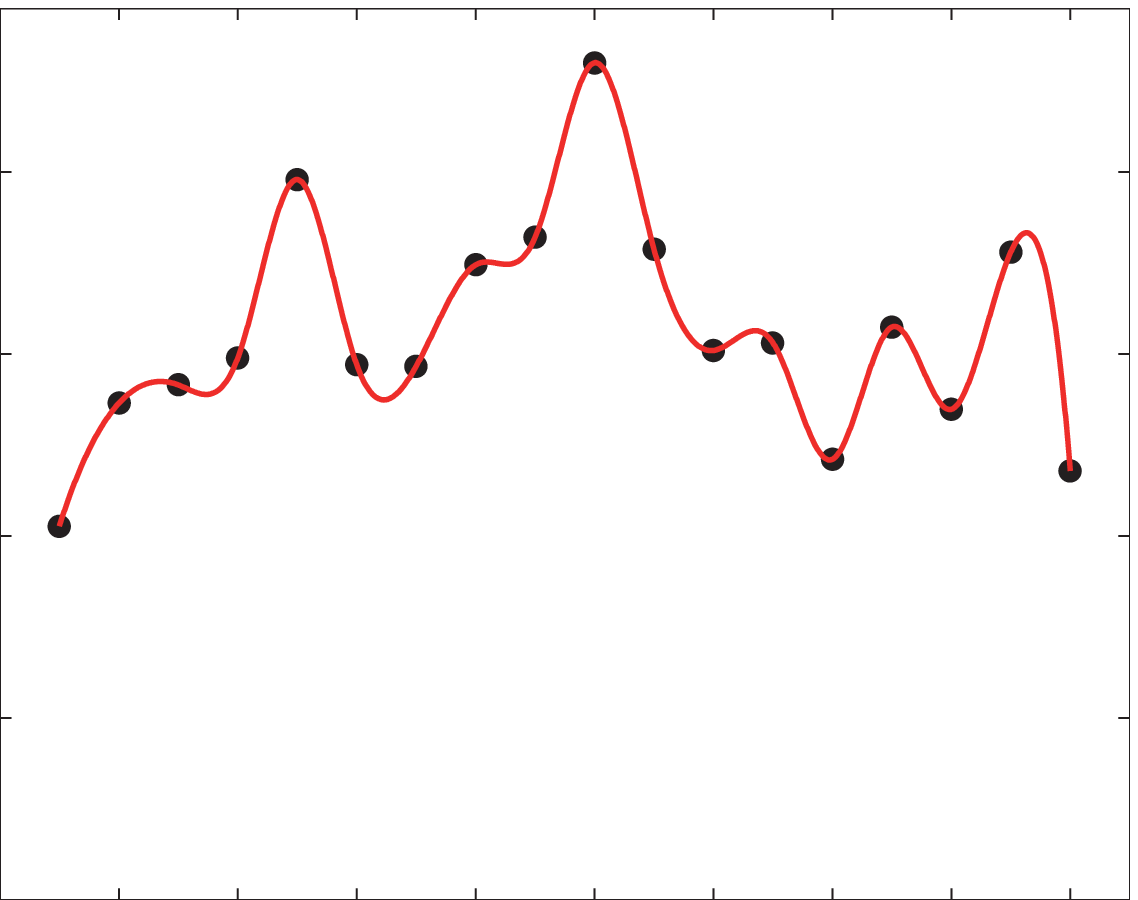}\rule{0cm}{16mm} & \includegraphics[width=1.9cm]{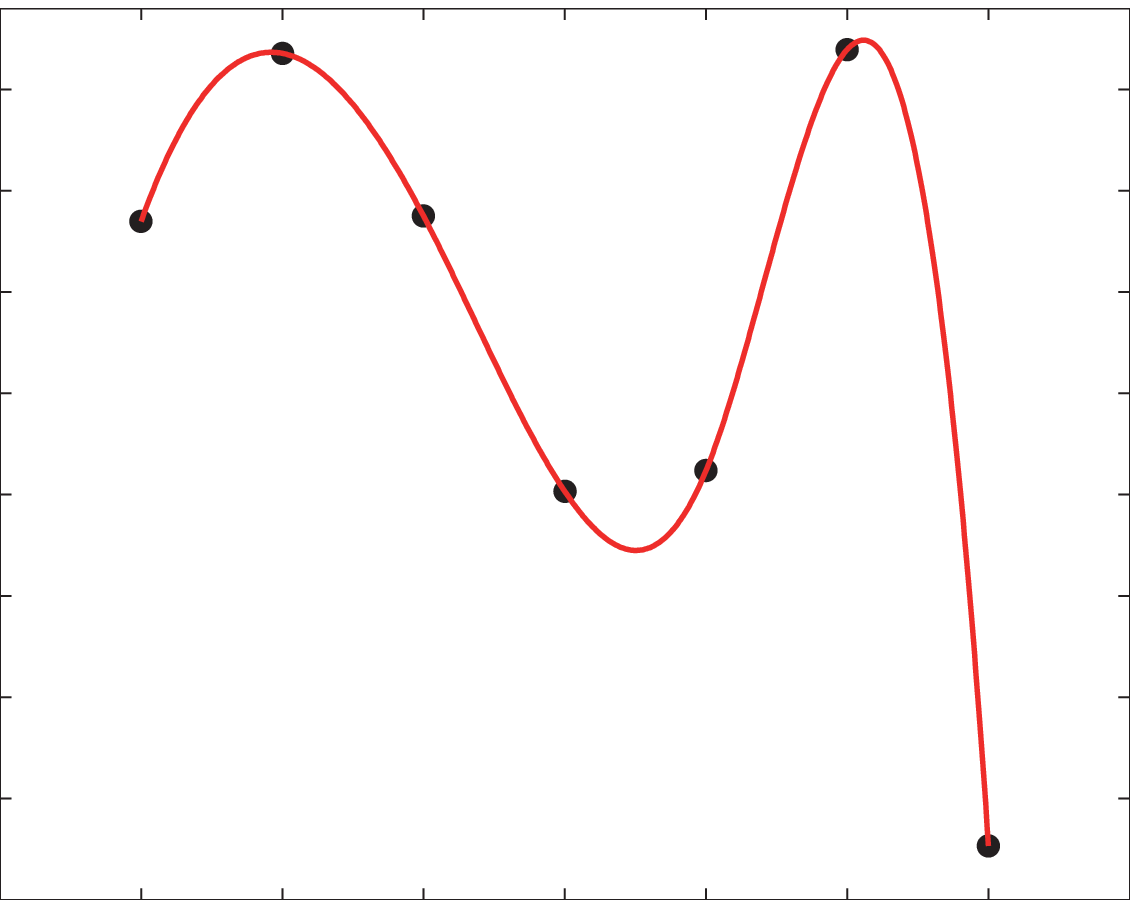} & \includegraphics[width=1.9cm]{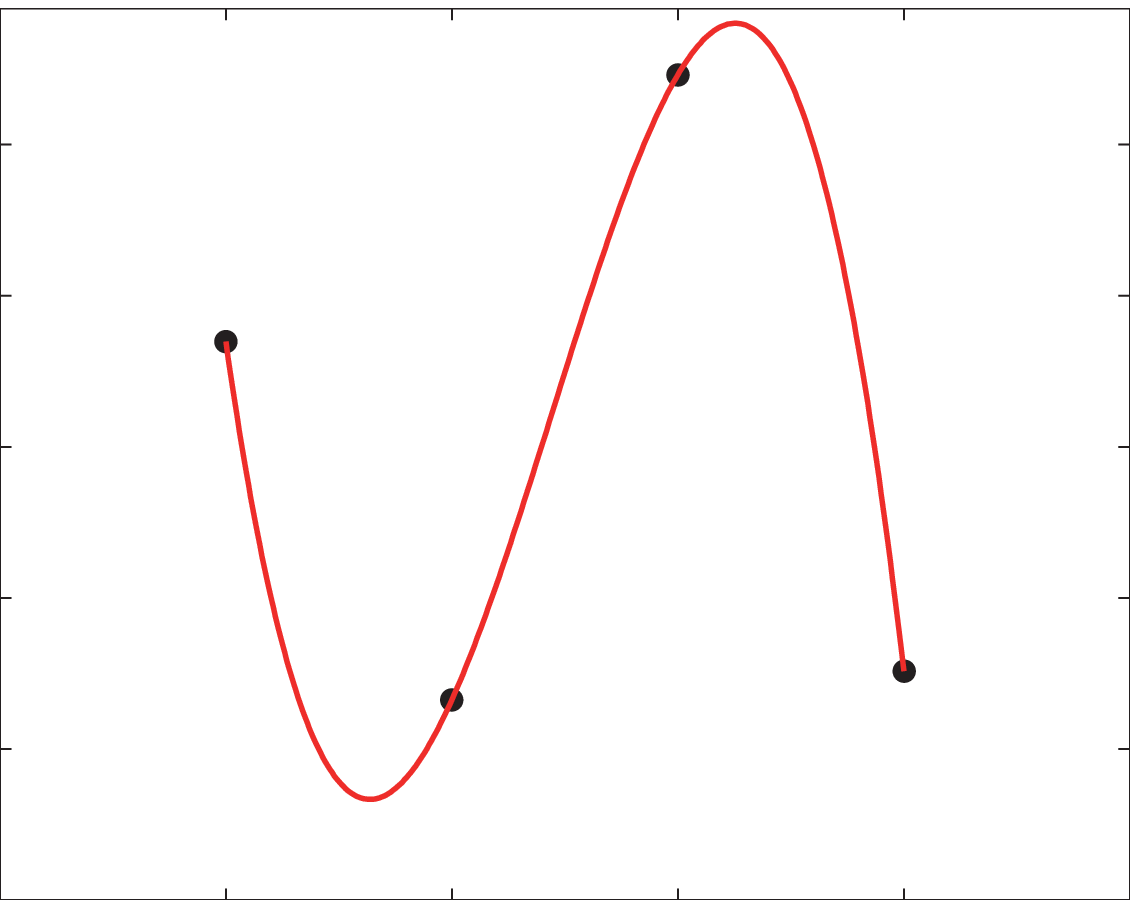} & \includegraphics[width=1.9cm]{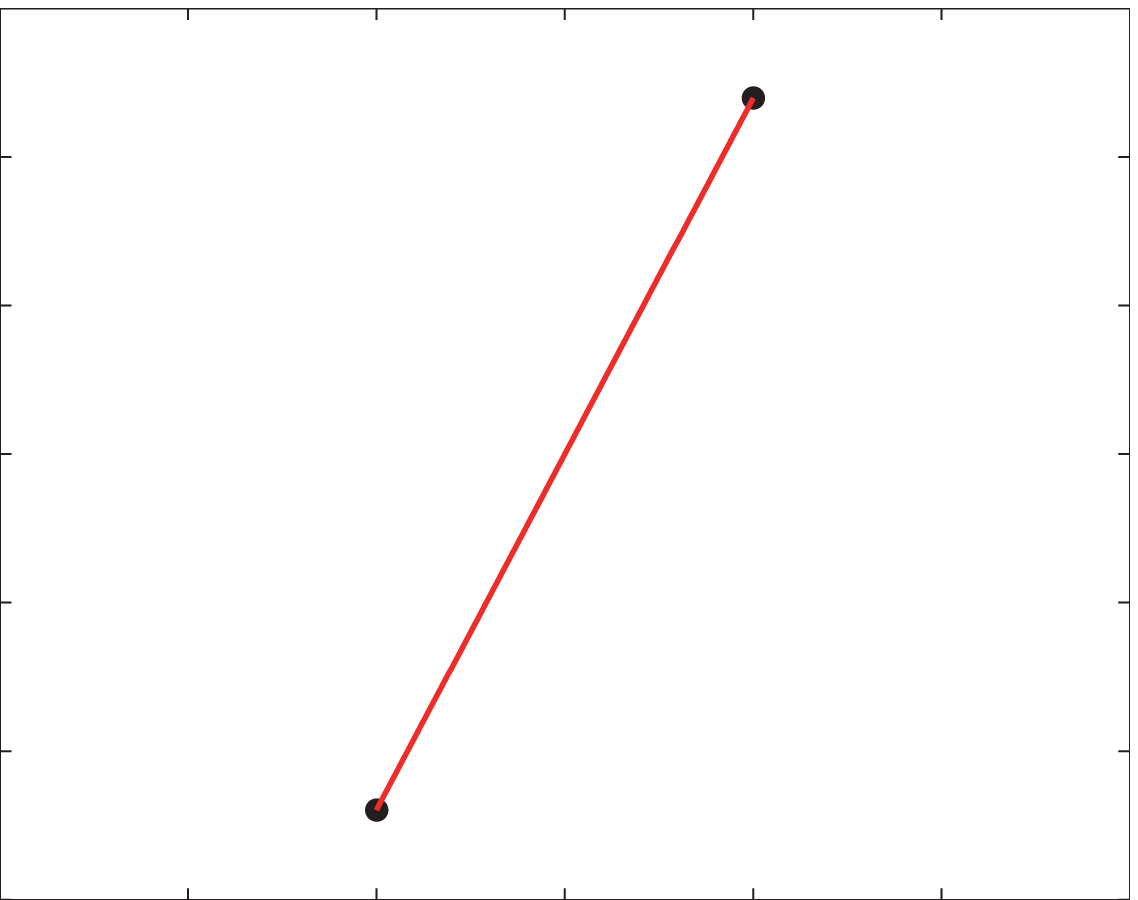}\\
 & \tabincell{l}{\{0.008, 0.156, 0.583, 0.003\} \\ \{0.997, 0.968\}} & \tabincell{l}{\{0.014, 0.076, 0.172, 0.002\} \\ \{0.998, 0.975\}} & \tabincell{l}{\{0.036, 0.085, 0.145 0.004\} \\ \{0.996, 0.964\}} & \tabincell{l}{\{0.069, 0.102, 0.143, 0.010\} \\ \{0.991, 0.931\}}\\\hline
\multirow{2}*[5mm]{SA-IIS} & \includegraphics[width=1.9cm]{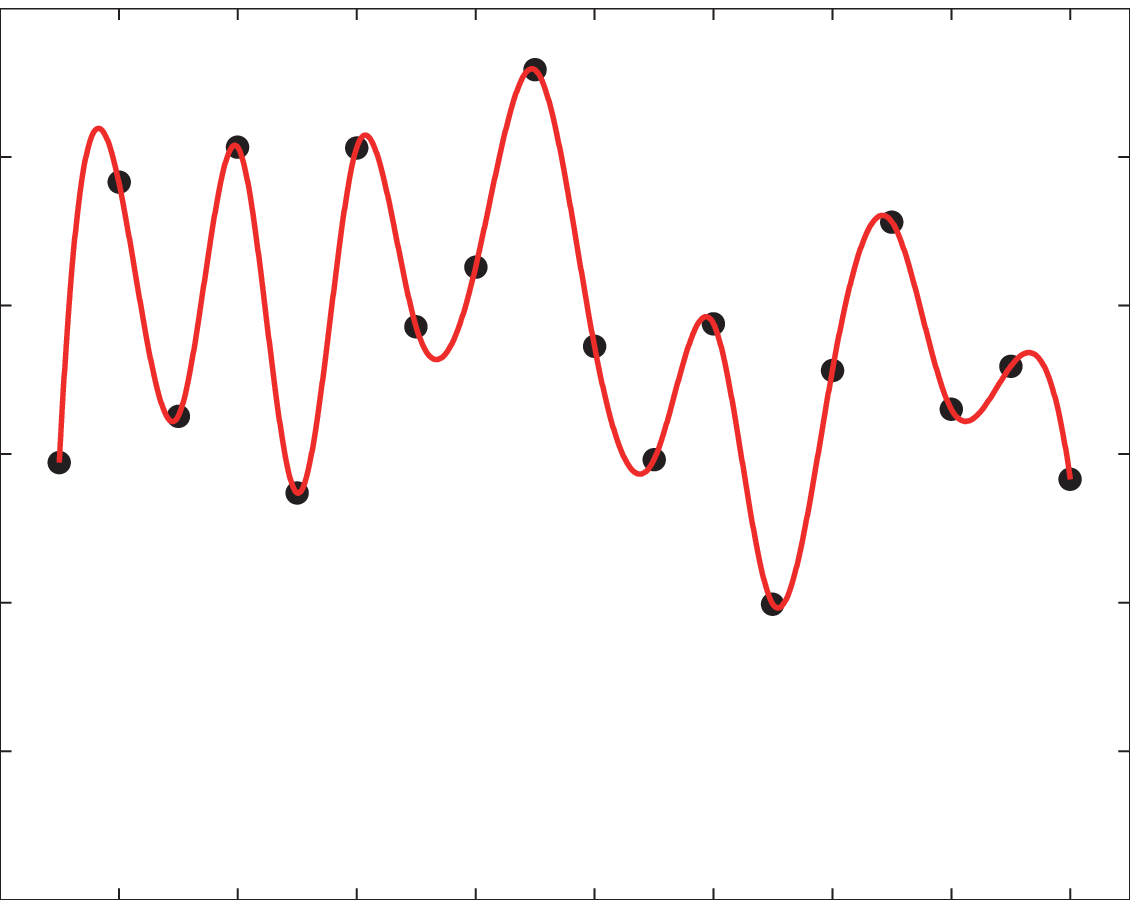}\rule{0cm}{16mm} & \includegraphics[width=1.9cm]{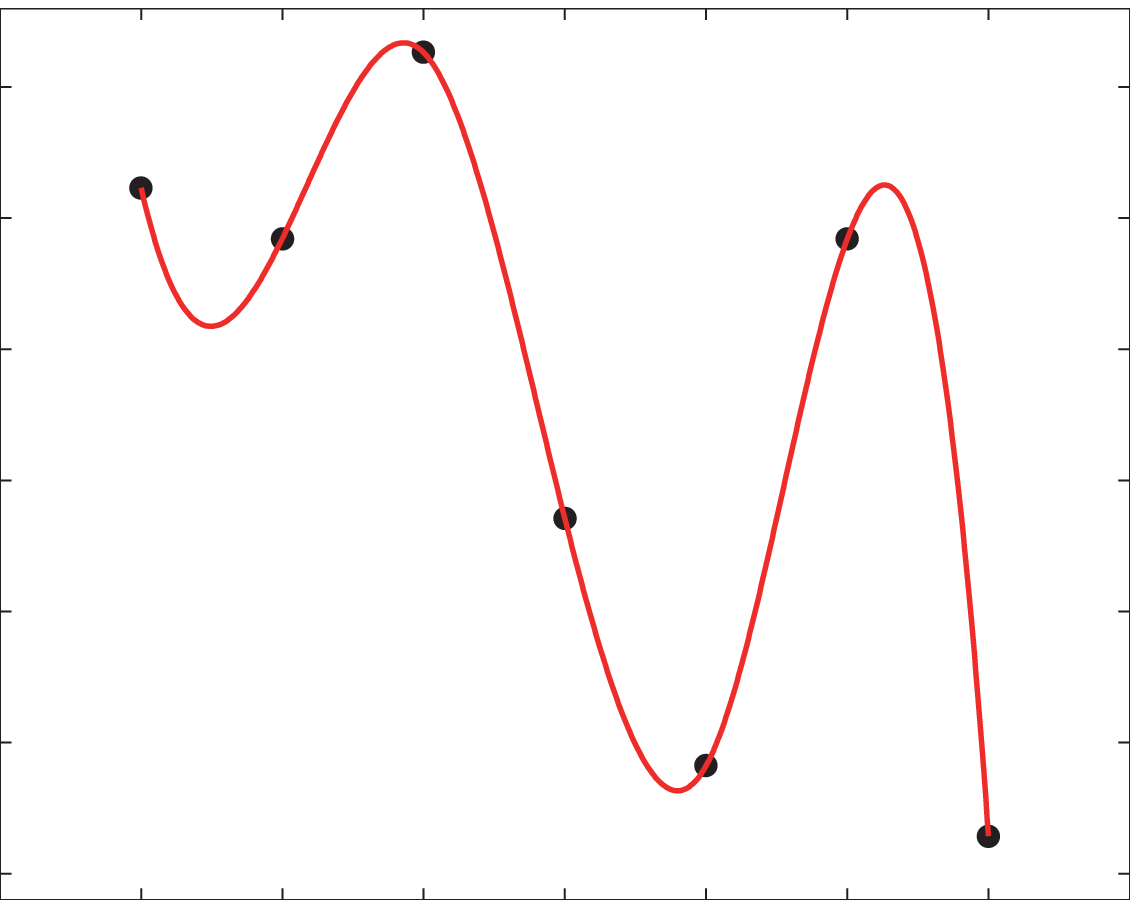} & \includegraphics[width=1.9cm]{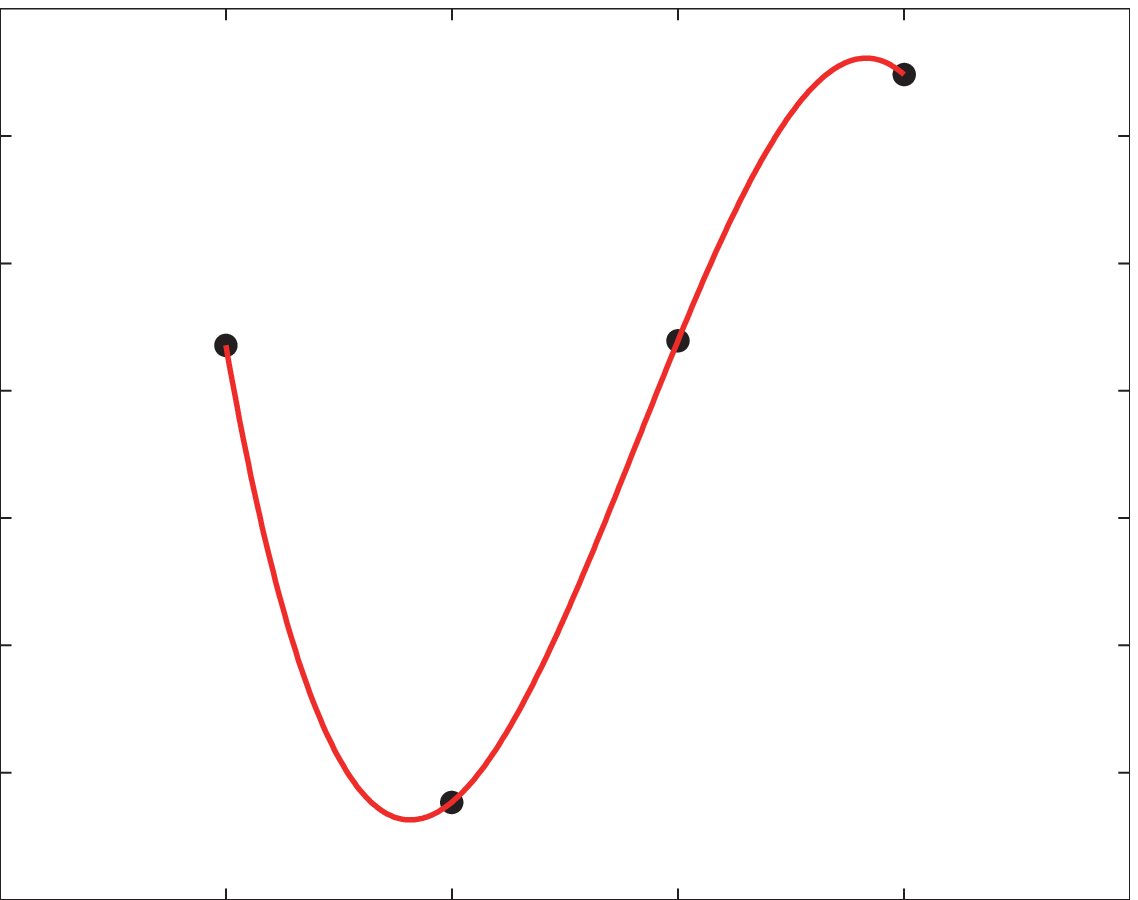} & \includegraphics[width=1.9cm]{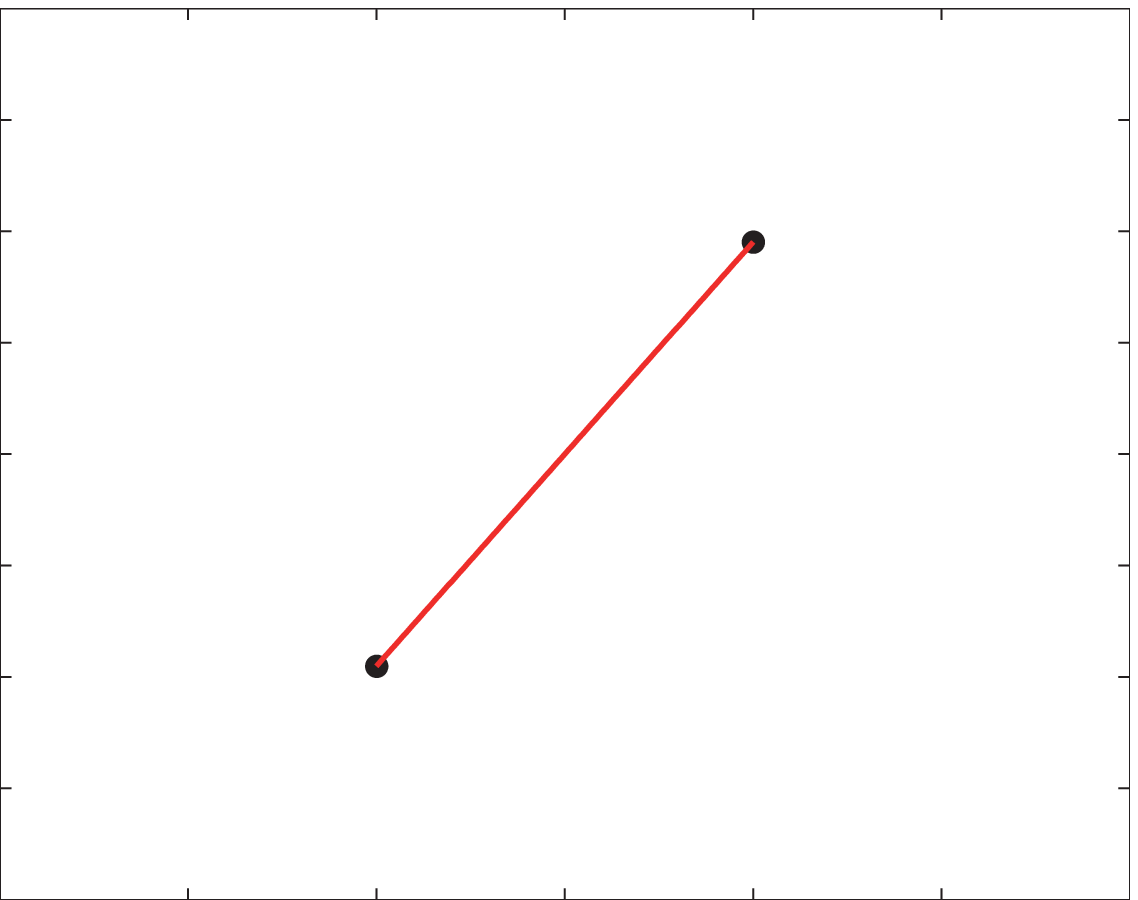}\\
 & \tabincell{l}{\{\textbf{0.004}, \textbf{0.083}, 0.283, \textbf{0.00077}\} \\ \{\textbf{0.99923}, 0.9843\}} & \tabincell{l}{\{0.013, 0.072, 	0.160, 0.002\} \\ \{0.998, 0.977\}} & \tabincell{l}{\{0.016, 0.042 0.067 0.0008\} \\ \{0.9992,	0.984\}} & \tabincell{l}{\{0.012, 0.018,	0.025,	0.0003\} \\ \{0.9997, 0.988\}}\\\hline
 \multirow{2}*[5mm]{SA-BFGS} & \includegraphics[width=1.9cm]{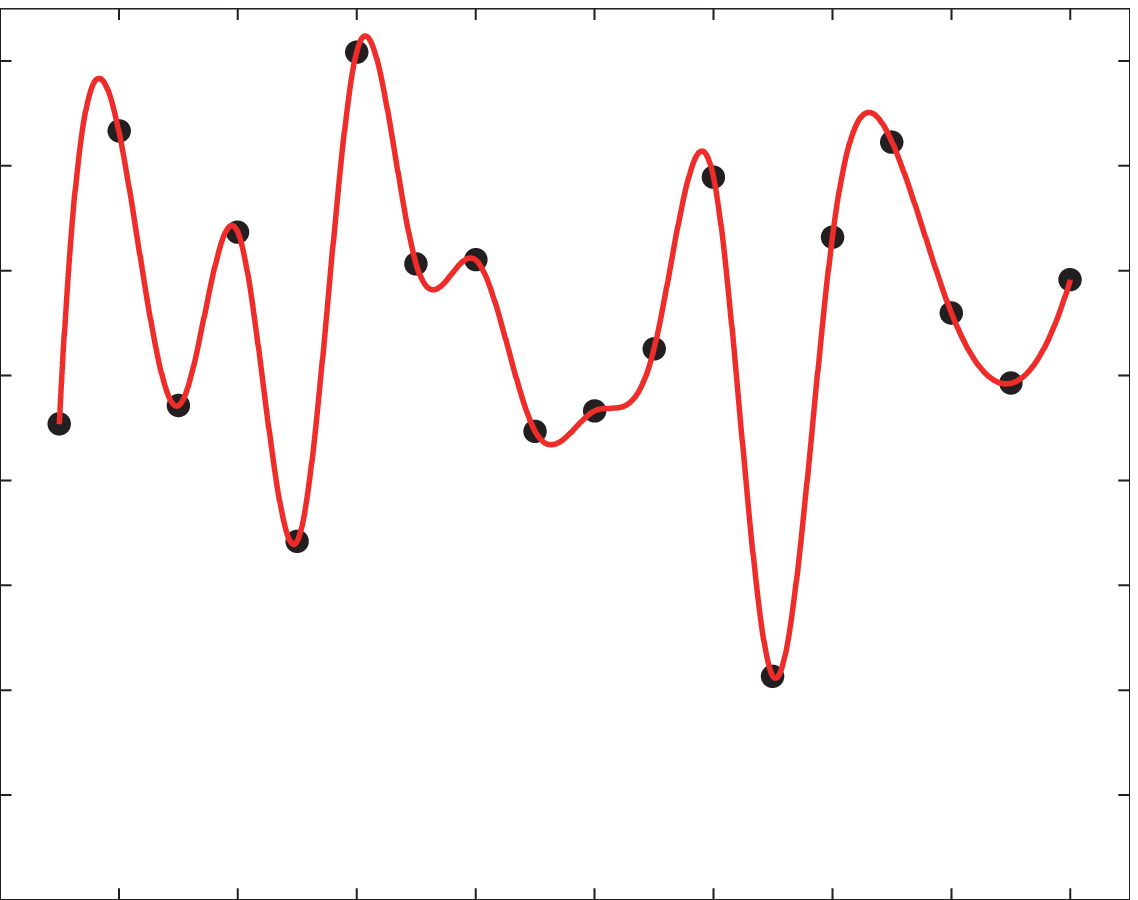}\rule{0cm}{16mm} & \includegraphics[width=1.9cm]{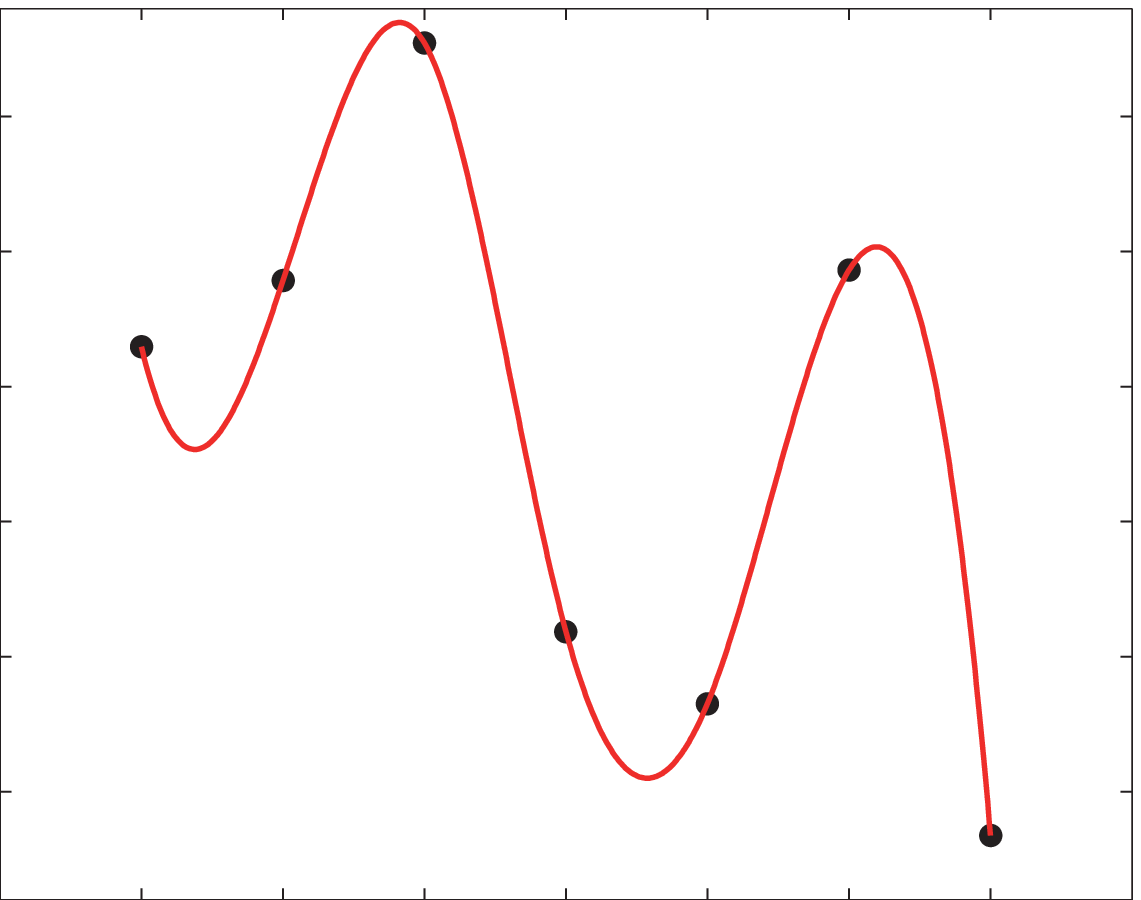} & \includegraphics[width=1.9cm]{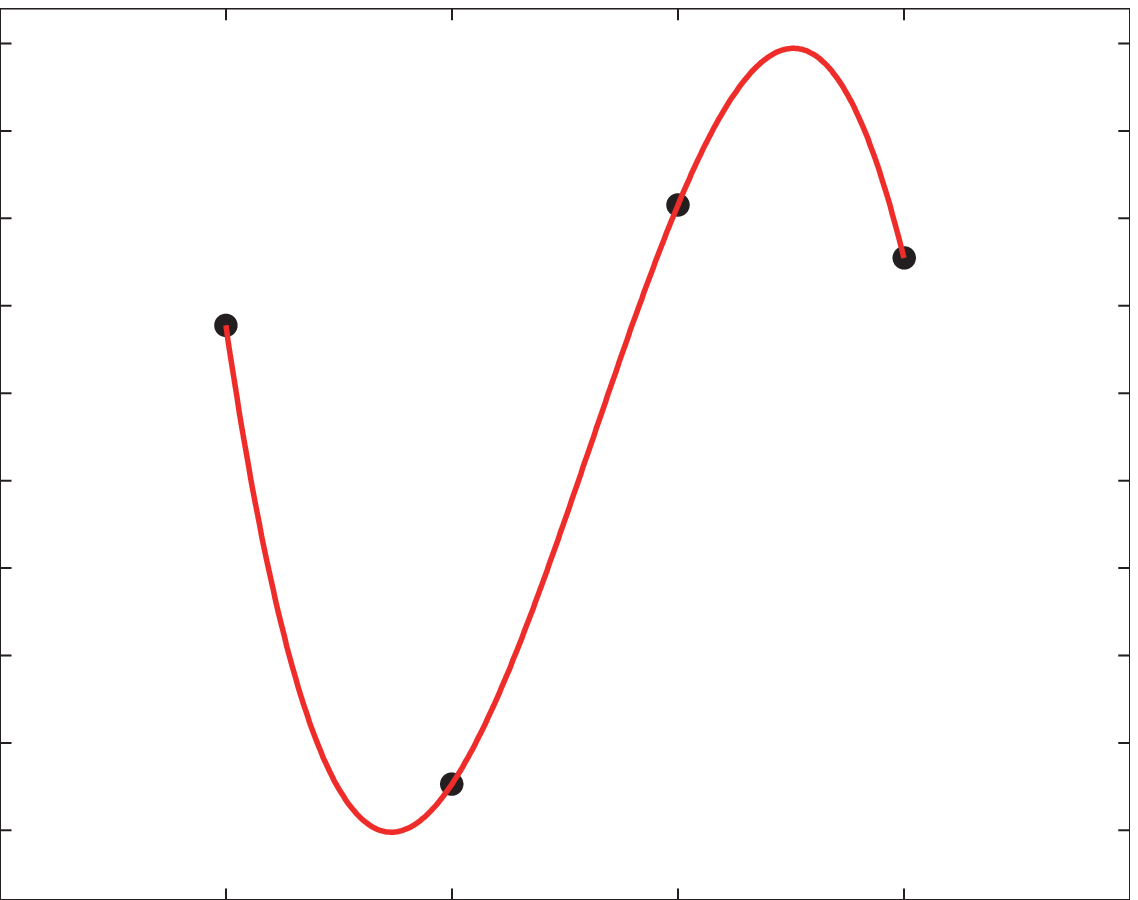} & \includegraphics[width=1.9cm]{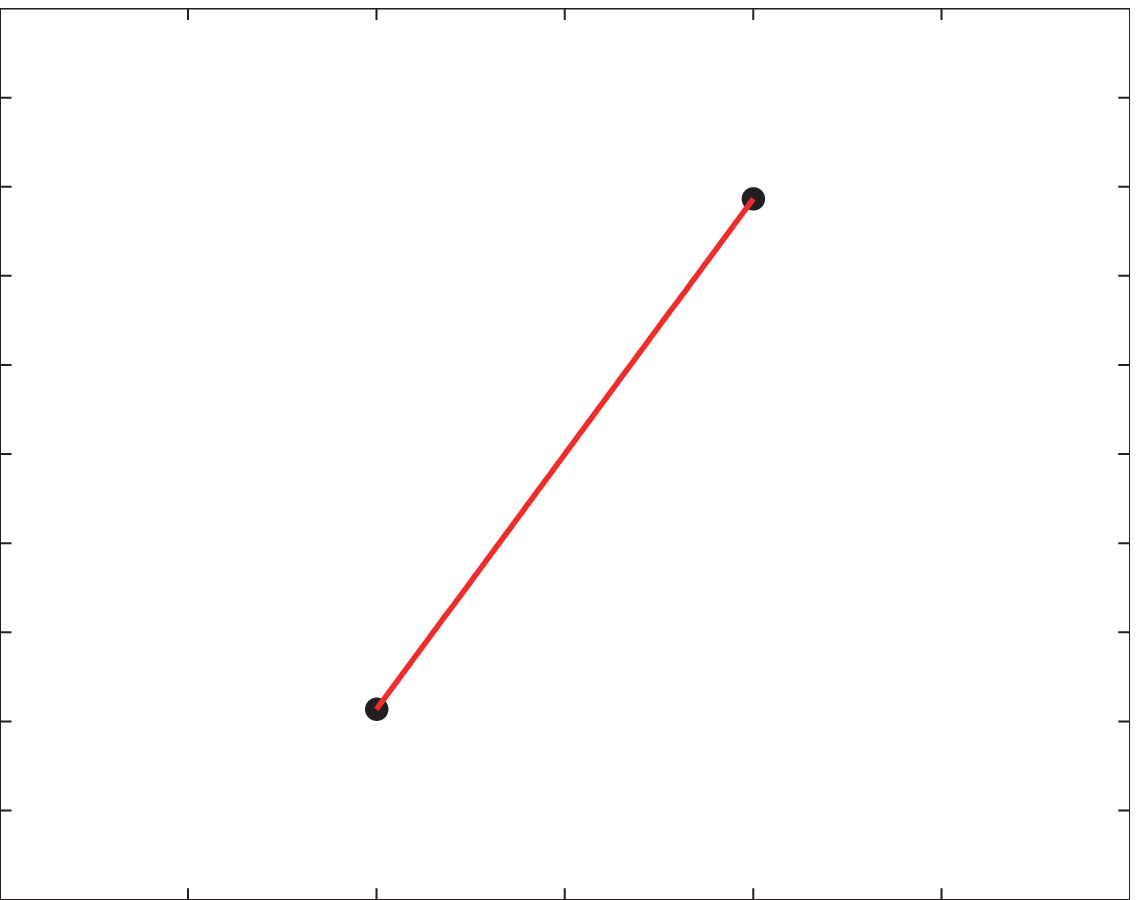}\\
 & \tabincell{l}{\{0.006, 0.086, \textbf{0.260}, 0.00081\} \\ \{0.99919, \textbf{0.986}\}} & \tabincell{l}{\{\textbf{0.012}, \textbf{0.056}, \textbf{0.120}, \textbf{0.001}\} \\ \{\textbf{0.999}, \textbf{0.983}\}} & \tabincell{l}{\{\textbf{0.014}, \textbf{0.034}, \textbf{0.055}, \textbf{0.0006}\} \\ \{\textbf{0.9994}, \textbf{0.987}\}} & \tabincell{l}{\{\textbf{0.007}, \textbf{0.010}, \textbf{0.014}, \textbf{0.0001}\} \\ \{\textbf{0.9999}, \textbf{0.993}\}}\\\hline
\end{tabular}
\end{center}
\vspace{-6mm}
\end{table*}

\subsubsection{Artificial Dataset}
\label{sect:exp_art}
In order to visually show the results of the LDL algorithms on the artificial dataset, the description degrees of the three labels are regarded as the three color channels of the RGB color space, respectively. In this way, the color of a point in the instance space will visually represent its label distribution. Thus, the predictions made by the LDL algorithms can be compared with the ground-truth label distributions through observing the color patterns on the manifold where the test examples lie on. For easier comparison, the images are visually enhanced by applying a decorrelation stretch process. The results are shown in Fig.~\ref{fig:aritficial:colors}. It can be seen that the two specialized LDL algorithms, SA-IIS and SA-BFGS, predict almost identical color patterns with the ground-truth. PT-Bayes and AA-$k$NN can also discover similar color patterns with the ground-truth. However, PT-SVM and AA-BP fail to obtain a reasonable result.

For quantitative analysis, the six measures listed in Fig.~\ref{fig:measures} calculated from the LDL predictions on the artificial dataset are given in Table~\ref{table:art_results}. On each measure, the algorithms are ranked in decreasing order of their performance, and the best performance is highlighted by boldface. The ranks are given in the parentheses right after the measure values, and the average ranks are given in the second last row of the table. The running time (train / test) on an Intel Core i7 3.4GHz workstation is given in the last row of the table. Note that since the training set and the test set of the artificial dataset are fixed (see Table~\ref{table:datasets}) in order to visualize in Fig.~\ref{fig:aritficial:colors} the LDL predictions on exactly the same test set, each LDL algorithm only runs once on the test set. So there is no record of standard deviation as in the ten-fold cross validation on the real-world datasets shown in Table~\ref{table:Chebyshev} to Table~\ref{table:Intersection}.

The first observation from Table~\ref{table:art_results} is that the performances of the best three algorithms (SA-IIS, SA-BFGS and PT-Bayes) evaluated by different measures are quite different. This proves the effectiveness of the selection process described in Section~\ref{subsect:criteria} to obtain a set of representative and diverse measures. Each of the six measures reflects a certain aspect of the compared LDL algorithms. For example, PT-Bayes appears very good on the measures `Clark' and `Canberra' because it has been evidenced that such measures are sensitive to small changes near zero \cite{Gordon99}, and the Gaussian assumption in PT-Bayes makes it less apt to produce near-zero outputs than SA-IIS and SA-BFGS.

The second observation from Table~\ref{table:art_results} is that the quantitative comparison results are consistent with the visual comparison results in Fig.~\ref{fig:aritficial:colors}. Based on the average ranks, the performances of the six LDL algorithms on the artificial dataset can be ranked as SA-BFGS $=$ SA-IIS $\succ$ PT-Bayes $\succ$ AA-$k$NN $\succ$ AA-BP $\succ$ PT-SVM. The specialized algorithms (SA-IIS and SA-BFGS) achieve the best performance since they directly aim to minimize the distance between the predicted and real label distributions (in fact, their optimization target is one of the measures in Fig.~\ref{fig:measures}, i.e., the Kullback-Leibler divergence). PT-Bayes also achieves relatively good performance since the Gaussian assumption on each class could match the simple case of the toy data well. AA-$k$NN can also get reasonable result since it is extended from the stable training-free algorithm $k$NN. Finally, the poor performance of PT-SVM and AA-BP is possibly due to overfitting since they have more parameters to learn and the training set (500 examples) is relatively small compared to the test set (40,401 examples) and they are sampled from very different distributions.

As for the running time, the training time of AA-BP is the longest due to slow convergence of the neural network. AA-$k$NN does not need to train but costs the most test time. SA-BFGS and SA-IIS cost the same minimum test time because they share the same parametric model. But consistent to the remarks in \cite{Malouf02}, the training of SA-BFGS is much faster than that of SA-IIS. In general, SA-BFGS and PT-Bayes are the most efficient LDL algorithms in Table~\ref{table:art_results}.

\begin{table*}[tb]
\scriptsize
\begin{center}
\caption{Experimental Results (mean$\pm$std(rank)) on the Real-world Datasets Measured by Chebyshev Distance $\downarrow$} \label{table:Chebyshev}
\begin{tabular}{lcccccc}
\hline \hline \noalign{\smallskip}
Dataset & PT-Bayes & PT-SVM & AA-$k$NN & AA-BP & SA-IIS & SA-BFGS\\\hline\noalign{\smallskip}
Yeast-alpha &0.174$\pm$0.011(6)&0.017$\pm$0.002(4)&0.0147$\pm$0.001(2)&0.036$\pm$0.003(5)&0.0148$\pm$0.001(3)&\textbf{0.013$\pm$0.001(1)} \\
Yeast-cdc &0.172$\pm$0.016(6)&0.020$\pm$0.003(4)&0.0177$\pm$0.001(2)&0.037$\pm$0.002(5)&0.0178$\pm$0.001(3)&\textbf{0.016$\pm$0.001(1)} \\
Yeast-elu &0.166$\pm$0.014(6)&0.019$\pm$0.001(4)&0.0177$\pm$0.0005(2)&0.036$\pm$0.002(5)&0.0178$\pm$0.001(3)&\textbf{0.016$\pm$0.001(1)} \\
Yeast-diau &0.167$\pm$0.007(6)&0.046$\pm$0.004(4)&0.0392$\pm$0.001(3)&0.048$\pm$0.003(5)&0.0386$\pm$0.001(2)&\textbf{0.037$\pm$0.002(1)} \\
Yeast-heat &0.176$\pm$0.018(6)&0.046$\pm$0.001(4)&0.045$\pm$0.001(3)&0.052$\pm$0.003(5)&0.043$\pm$0.001(2)&\textbf{0.042$\pm$0.001(1)} \\
Yeast-spo &0.178$\pm$0.009(6)&0.065$\pm$0.006(4)&0.064$\pm$0.002(3)&0.067$\pm$0.005(5)&0.060$\pm$0.004(2)&\textbf{0.058$\pm$0.004(1)} \\
Yeast-cold &0.177$\pm$0.011(6)&0.0574$\pm$0.003(5)&0.055$\pm$0.002(3)&0.0572$\pm$0.003(4)&0.053$\pm$0.002(2)&\textbf{0.051$\pm$0.002(1)} \\
Yeast-dtt &0.177$\pm$0.010(6)&0.040$\pm$0.001(4)&0.0392$\pm$0.001(3)&0.043$\pm$0.002(5)&0.0388$\pm$0.001(2)&\textbf{0.036$\pm$0.001(1)} \\
Yeast-spo5 &0.211$\pm$0.011(6)&0.0929$\pm$0.006(3)&0.096$\pm$0.005(5)&0.094$\pm$0.006(4)&0.0928$\pm$0.006(2)&\textbf{0.091$\pm$0.005(1)} \\
Yeast-spoem &0.190$\pm$0.017(6)&0.091$\pm$0.005(4)&0.093$\pm$0.004(5)&0.0892$\pm$0.005(3)&0.0891$\pm$0.005(2)&\textbf{0.087$\pm$0.005(1)} \\
Human Gene &0.195$\pm$0.085(6)&0.054$\pm$0.004(3)&0.065$\pm$0.005(5)&0.059$\pm$0.004(4)&0.0534$\pm$0.004(2)&\textbf{0.0533$\pm$0.004(1)} \\
Natural Scene &0.407$\pm$0.027(5)&0.414$\pm$0.036(6)&0.374$\pm$0.013(4)&0.335$\pm$0.016(2)&0.341$\pm$0.017(3)&\textbf{0.322$\pm$0.017(1)} \\
s-JAFFE &0.121$\pm$0.016(4)&0.127$\pm$0.017(5)&0.114$\pm$0.017(2)&0.130$\pm$0.017(6)&0.117$\pm$0.015(3)&\textbf{0.107$\pm$0.015(1)} \\
s-BU\_3DFE &0.116$\pm$0.004(5)&0.119$\pm$0.006(6)&0.103$\pm$0.003(2)&0.113$\pm$0.005(4)&0.111$\pm$0.004(3)&\textbf{0.088$\pm$0.003(1)} \\
Movie &0.199$\pm$0.009(5)&0.213$\pm$0.039(6)&0.154$\pm$0.005(3)&0.157$\pm$0.013(4)&0.150$\pm$0.008(2)&\textbf{0.136$\pm$0.010(1)} \\
\hline\noalign{\smallskip}
Avg. Rank &5.67&4.40&3.13&4.40&2.40&1.00 \\
\hline \hline
\end{tabular}
\end{center}
\vspace{-5mm}
\end{table*}

\begin{table*}[tb]
\scriptsize
\begin{center}
\caption{Experimental Results (mean$\pm$std(rank)) on the Real-world Datasets Measured by Clark Distance $\downarrow$} \label{table:Clark}
\begin{tabular}{lcccccc}
\hline \hline \noalign{\smallskip}
Dataset & PT-Bayes & PT-SVM & AA-$k$NN & AA-BP & SA-IIS & SA-BFGS\\\hline\noalign{\smallskip}
Yeast-alpha &1.729$\pm$0.076(6)&0.277$\pm$0.031(4)&0.232$\pm$0.012(2)&0.711$\pm$0.054(5)&0.233$\pm$0.012(3)&\textbf{0.210$\pm$0.014(1)} \\
Yeast-cdc &1.491$\pm$0.062(6)&0.260$\pm$0.029(4)&0.237$\pm$0.014(3)&0.568$\pm$0.037(5)&0.235$\pm$0.012(2)&\textbf{0.216$\pm$0.013(1)} \\
Yeast-elu &1.379$\pm$0.082(6)&0.234$\pm$0.015(4)&0.218$\pm$0.005(3)&0.505$\pm$0.034(5)&0.216$\pm$0.007(2)&\textbf{0.199$\pm$0.005(1)} \\
Yeast-diau &0.771$\pm$0.032(6)&0.246$\pm$0.014(4)&0.212$\pm$0.004(3)&0.263$\pm$0.017(5)&0.209$\pm$0.007(2)&\textbf{0.200$\pm$0.009(1)} \\
Yeast-heat &0.670$\pm$0.047(6)&0.198$\pm$0.007(4)&0.195$\pm$0.005(3)&0.228$\pm$0.015(5)&0.188$\pm$0.003(2)&\textbf{0.182$\pm$0.003(1)} \\
Yeast-spo &0.695$\pm$0.029(6)&0.273$\pm$0.024(4)&0.271$\pm$0.011(3)&0.292$\pm$0.022(5)&0.255$\pm$0.017(2)&\textbf{0.250$\pm$0.017(1)} \\
Yeast-cold &0.484$\pm$0.032(6)&0.155$\pm$0.008(4)&0.150$\pm$0.007(3)&0.155$\pm$0.009(5)&0.144$\pm$0.005(2)&\textbf{0.139$\pm$0.005(1)} \\
Yeast-dtt &0.480$\pm$0.029(6)&0.108$\pm$0.005(4)&0.106$\pm$0.004(3)&0.118$\pm$0.007(5)&0.105$\pm$0.004(2)&\textbf{0.098$\pm$0.004(1)} \\
Yeast-spo5 &0.438$\pm$0.027(6)&0.187$\pm$0.013(3)&0.193$\pm$0.011(5)&0.189$\pm$0.012(4)&0.187$\pm$0.013(2)&\textbf{0.184$\pm$0.012(1)} \\
Yeast-spoem &0.315$\pm$0.029(6)&0.134$\pm$0.008(4)&0.137$\pm$0.006(5)&0.1323$\pm$0.008(3)&0.1321$\pm$0.007(2)&\textbf{0.129$\pm$0.008(1)} \\
Human Gene &4.674$\pm$0.415(6)&2.139$\pm$0.087(3)&2.388$\pm$0.109(4)&3.344$\pm$0.250(5)&2.123$\pm$0.088(2)&\textbf{2.111$\pm$0.086(1)} \\
Natural Scene &2.523$\pm$0.027(5)&2.557$\pm$0.045(6)&\textbf{1.418$\pm$0.057(1)}&2.458$\pm$0.023(3)&2.461$\pm$0.025(4)&2.411$\pm$0.023(2) \\
s-JAFFE &0.430$\pm$0.035(4)&0.457$\pm$0.039(5)&0.410$\pm$0.050(2)&0.510$\pm$0.054(6)&0.419$\pm$0.034(3)&\textbf{0.399$\pm$0.044(1)} \\
s-BU\_3DFE &0.467$\pm$0.009(4)&0.494$\pm$0.022(6)&0.396$\pm$0.006(2)&0.477$\pm$0.030(5)&0.416$\pm$0.009(3)&\textbf{0.367$\pm$0.009(1)} \\
Movie &0.799$\pm$0.035(6)&0.797$\pm$0.108(5)&0.652$\pm$0.023(3)&0.675$\pm$0.048(4)&0.591$\pm$0.028(2)&\textbf{0.589$\pm$0.038(1)} \\
\hline\noalign{\smallskip}
Avg. Rank &5.67&4.27&3.00&4.67&2.33&1.07 \\
\hline \hline
\end{tabular}
\end{center}
\vspace{-5mm}
\end{table*}

\begin{table*}[tb]
\scriptsize
\begin{center}
\caption{Experimental Results (mean$\pm$std(rank)) on the Real-world Datasets Measured by Canberra Metric $\downarrow$} \label{table:Canberra}
\begin{tabular}{lcccccc}
\hline \hline \noalign{\smallskip}
Dataset & PT-Bayes & PT-SVM & AA-$k$NN & AA-BP & SA-IIS & SA-BFGS\\\hline\noalign{\smallskip}
Yeast-alpha &6.382$\pm$0.305(6)&0.921$\pm$0.107(4)&0.758$\pm$0.040(2)&2.352$\pm$0.173(5)&0.763$\pm$0.042(3)&\textbf{0.684$\pm$0.046(1)} \\
Yeast-cdc &4.987$\pm$0.222(6)&0.785$\pm$0.084(4)&0.717$\pm$0.041(3)&1.718$\pm$0.110(5)&0.709$\pm$0.036(2)&\textbf{0.649$\pm$0.041(1)} \\
Yeast-elu &4.461$\pm$0.282(6)&0.691$\pm$0.047(4)&0.644$\pm$0.016(3)&1.488$\pm$0.098(5)&0.639$\pm$0.019(2)&\textbf{0.583$\pm$0.016(1)} \\
Yeast-diau &1.744$\pm$0.071(6)&0.528$\pm$0.031(4)&0.455$\pm$0.011(3)&0.568$\pm$0.033(5)&0.449$\pm$0.017(2)&\textbf{0.431$\pm$0.020(1)} \\
Yeast-heat &1.415$\pm$0.102(6)&0.396$\pm$0.016(4)&0.392$\pm$0.010(3)&0.459$\pm$0.031(5)&0.377$\pm$0.005(2)&\textbf{0.364$\pm$0.006(1)} \\
Yeast-spo &1.473$\pm$0.069(6)&0.565$\pm$0.049(4)&0.559$\pm$0.024(3)&0.599$\pm$0.043(5)&0.523$\pm$0.034(2)&\textbf{0.513$\pm$0.035(1)} \\
Yeast-cold &0.845$\pm$0.059(6)&0.267$\pm$0.014(4)&0.260$\pm$0.013(3)&0.268$\pm$0.015(5)&0.249$\pm$0.009(2)&\textbf{0.240$\pm$0.010(1)} \\
Yeast-dtt &0.846$\pm$0.051(6)&0.186$\pm$0.008(4)&0.182$\pm$0.007(3)&0.204$\pm$0.012(5)&0.181$\pm$0.005(2)&\textbf{0.169$\pm$0.005(1)} \\
Yeast-spo5 &0.681$\pm$0.038(6)&0.287$\pm$0.019(3)&0.297$\pm$0.016(5)&0.291$\pm$0.018(4)&0.287$\pm$0.019(2)&\textbf{0.283$\pm$0.018(1)} \\
Yeast-spoem &0.424$\pm$0.038(6)&0.187$\pm$0.011(4)&0.191$\pm$0.008(5)&0.1842$\pm$0.011(3)&0.1840$\pm$0.010(2)&\textbf{0.179$\pm$0.011(1)} \\
Human Gene &34.238$\pm$3.634(6)&14.631$\pm$0.647(3)&16.283$\pm$0.818(4)&22.788$\pm$1.841(5)&14.541$\pm$0.653(2)&\textbf{14.453$\pm$0.645(1)} \\
Natural Scene &7.149$\pm$0.109(5)&7.208$\pm$0.205(6)&\textbf{3.044$\pm$0.137(1)}&6.767$\pm$0.095(4)&6.765$\pm$0.104(3)&6.620$\pm$0.097(2) \\
s-JAFFE &0.904$\pm$0.086(4)&0.935$\pm$0.074(5)&0.843$\pm$0.113(2)&1.046$\pm$0.124(6)&0.875$\pm$0.086(3)&\textbf{0.820$\pm$0.103(1)} \\
s-BU\_3DFE &1.116$\pm$0.020(5)&1.147$\pm$0.064(6)&0.841$\pm$0.014(2)&1.051$\pm$0.064(4)&0.934$\pm$0.022(3)&\textbf{0.794$\pm$0.019(1)} \\
Movie &1.547$\pm$0.075(6)&1.537$\pm$0.216(5)&1.276$\pm$0.046(4)&1.269$\pm$0.089(3)&\textbf{1.137$\pm$0.057(1)}&1.138$\pm$0.079(2) \\
\hline\noalign{\smallskip}
Avg. Rank &5.73&4.27&3.07&4.60&2.20&1.13 \\
\hline \hline
\end{tabular}
\end{center}
\vspace{-5mm}
\end{table*}

\subsubsection{Real-world Datasets}
To give a direct idea of the LDL predictions on the real-world datasets, some typical examples of the predicted label distributions by the six LDL algorithms are shown in Table~\ref{table:predictions}. The first row shows the real label distributions of four typical test instances with 18, 7, 4 and 2 labels, which come from the 2nd, 5th, 9th, and 11th datasets in Table~\ref{table:datasets}, respectively. Each of the following rows shows the corresponding predictions of one LDL algorithm. The evaluation measures shown in Fig.~\ref{fig:measures} are given under each predicted label distribution in two sets. The upper set includes the four distances \{Chebyshev, Clark, Canberra, Kullback-Leibler\}, and the lower set includes the two similarities \{cosine, intersection\}. The best performance for each case is highlighted by boldface. As can be seen that the specialized algorithms (SA-IIS and SA-BFGS) can generally give better label distribution predictions compared with other algorithms, in the sense of either visual similarity or quantitative evaluation measures. It is worth mentioning that the order of labels will affect the shape of the label distribution, but not the correlations among the labels. As long as the label orders are the same in the training and test sets, it will not affect, at least for the algorithms proposed in this paper, the effectiveness of LDL.

Table~\ref{table:Chebyshev} to Table~\ref{table:Intersection} tabulate the results of the six LDL algorithms on the 15 real-world datasets (the 2nd to 16th datasets in Table~\ref{table:datasets}) evaluated by the six measures Chebyshev distance, Clark distance, Canberra metric, Kullback-Leibler divergence, cosine coefficient, and intersection similarity, respectively. Since the LDL algorithms are tested via ten-fold cross validation, the performance is represented by ``mean $\pm$ standard deviation'' of the corresponding measure calculated during the ten-fold cross validation. In each table, the best performance on each dataset is highlighted by boldface. The LDL algorithms are ranked in decreasing order of their performance on each dataset. The ranks are given in the parentheses right after the performance values. The average rank of each algorithm over all the datasets is also calculated and given in the last row of each table.

\begin{table*}[tb]
\scriptsize
\begin{center}
\caption{Experimental Results (mean$\pm$std(rank)) on the Real-world Datasets Measured by Kullback-Leibler Divergence $\downarrow$} \label{table:Kullback_Leibler}
\begin{tabular}{lcccccc}
\hline \hline \noalign{\smallskip}
Dataset & PT-Bayes & PT-SVM & AA-$k$NN & AA-BP & SA-IIS & SA-BFGS\\\hline\noalign{\smallskip}
Yeast-alpha &0.719$\pm$0.080(6)&0.009$\pm$0.002(4)&0.0066$\pm$0.001(2)&0.081$\pm$0.011(5)&0.0067$\pm$0.001(3)&\textbf{0.006$\pm$0.001(1)} \\
Yeast-cdc &0.603$\pm$0.073(6)&0.010$\pm$0.002(4)&0.0083$\pm$0.001(3)&0.060$\pm$0.007(5)&0.0082$\pm$0.001(2)&\textbf{0.007$\pm$0.001(1)} \\
Yeast-elu &0.556$\pm$0.071(6)&0.008$\pm$0.001(4)&0.0074$\pm$0.0004(3)&0.051$\pm$0.009(5)&0.0073$\pm$0.0005(2)&\textbf{0.006$\pm$0.0004(1)} \\
Yeast-diau &0.306$\pm$0.036(6)&0.019$\pm$0.002(4)&0.015$\pm$0.001(3)&0.024$\pm$0.004(5)&0.014$\pm$0.001(2)&\textbf{0.013$\pm$0.001(1)} \\
Yeast-heat &0.255$\pm$0.040(6)&0.0148$\pm$0.001(4)&0.0145$\pm$0.001(3)&0.021$\pm$0.004(5)&0.0133$\pm$0.0004(2)&\textbf{0.0126$\pm$0.0005(1)} \\
Yeast-spo &0.281$\pm$0.031(6)&0.0304$\pm$0.005(4)&0.0302$\pm$0.002(3)&0.034$\pm$0.006(5)&0.0254$\pm$0.003(2)&\textbf{0.0246$\pm$0.003(1)} \\
Yeast-cold &0.208$\pm$0.031(6)&0.0147$\pm$0.001(4)&0.014$\pm$0.001(3)&0.0149$\pm$0.002(5)&0.013$\pm$0.001(2)&\textbf{0.012$\pm$0.001(1)} \\
Yeast-dtt &0.206$\pm$0.029(6)&0.0073$\pm$0.001(4)&0.0072$\pm$0.001(3)&0.009$\pm$0.001(5)&0.0070$\pm$0.001(2)&\textbf{0.006$\pm$0.001(1)} \\
Yeast-spo5 &0.214$\pm$0.025(6)&0.03010$\pm$0.003(3)&0.033$\pm$0.003(5)&0.031$\pm$0.003(4)&0.03007$\pm$0.003(2)&\textbf{0.029$\pm$0.003(1)} \\
Yeast-spoem &0.190$\pm$0.038(6)&0.0280$\pm$0.004(4)&0.0285$\pm$0.003(5)&0.026$\pm$0.003(3)&0.025$\pm$0.003(2)&\textbf{0.024$\pm$0.003(1)} \\
Human Gene &1.887$\pm$0.766(6)&0.240$\pm$0.019(3)&0.301$\pm$0.026(4)&0.500$\pm$0.068(5)&0.238$\pm$0.019(2)&\textbf{0.236$\pm$0.019(1)} \\
Natural Scene &3.065$\pm$0.487(6)&1.447$\pm$0.243(4)&2.767$\pm$0.137(5)&0.875$\pm$0.029(3)&0.870$\pm$0.026(2)&\textbf{0.854$\pm$0.062(1)} \\
s-JAFFE &0.074$\pm$0.014(4)&0.086$\pm$0.016(5)&0.071$\pm$0.023(3)&0.113$\pm$0.030(6)&0.070$\pm$0.012(2)&\textbf{0.064$\pm$0.016(1)} \\
s-BU\_3DFE &0.079$\pm$0.004(4)&0.089$\pm$0.007(6)&0.065$\pm$0.002(2)&0.085$\pm$0.009(5)&0.068$\pm$0.004(3)&\textbf{0.049$\pm$0.002(1)} \\
Movie &0.953$\pm$0.352(6)&0.268$\pm$0.079(5)&0.201$\pm$0.011(4)&0.179$\pm$0.03(3)&\textbf{0.137$\pm$0.013(1)}&0.140$\pm$0.020(2) \\
\hline\noalign{\smallskip}
Avg. Rank &5.73&4.13&3.40&4.60&2.07&1.07 \\
\hline \hline
\end{tabular}
\end{center}
\vspace{-5mm}
\end{table*}

\begin{table*}[tb]
\scriptsize
\begin{center}
\caption{Experimental Results (mean$\pm$std(rank)) on the Real-world Datasets Measured by Cosine Coefficient $\uparrow$} \label{table:Cosine}
\begin{tabular}{lcccccc}
\hline \hline \noalign{\smallskip}
Dataset & PT-Bayes & PT-SVM & AA-$k$NN & AA-BP & SA-IIS & SA-BFGS\\\hline\noalign{\smallskip}
Yeast-alpha &0.743$\pm$0.015(6)&0.991$\pm$0.002(4)&0.9935$\pm$0.001(2)&0.949$\pm$0.006(5)&0.9934$\pm$0.001(3)&\textbf{0.995$\pm$0.001(1)} \\
Yeast-cdc &0.766$\pm$0.017(6)&0.991$\pm$0.002(4)&0.9920$\pm$0.001(3)&0.960$\pm$0.004(5)&0.9921$\pm$0.001(2)&\textbf{0.993$\pm$0.001(1)} \\
Yeast-elu &0.780$\pm$0.018(6)&0.992$\pm$0.001(4)&0.99286$\pm$0.0003(3)&0.965$\pm$0.004(5)&0.99290$\pm$0.0005(2)&\textbf{0.994$\pm$0.0004(1)} \\
Yeast-diau &0.856$\pm$0.007(6)&0.982$\pm$0.002(4)&0.986$\pm$0.001(3)&0.979$\pm$0.002(5)&0.987$\pm$0.001(2)&\textbf{0.988$\pm$0.001(1)} \\
Yeast-heat &0.866$\pm$0.015(6)&0.98607$\pm$0.001(4)&0.98612$\pm$0.001(3)&0.981$\pm$0.003(5)&0.987$\pm$0.0004(2)&\textbf{0.988$\pm$0.0005(1)} \\
Yeast-spo &0.859$\pm$0.008(6)&0.971$\pm$0.005(4)&0.972$\pm$0.002(3)&0.969$\pm$0.004(5)&0.976$\pm$0.003(2)&\textbf{0.977$\pm$0.003(1)} \\
Yeast-cold &0.898$\pm$0.008(6)&0.9860$\pm$0.001(4)&0.987$\pm$0.001(3)&0.9859$\pm$0.002(5)&0.988$\pm$0.001(2)&\textbf{0.989$\pm$0.001(1)} \\
Yeast-dtt &0.897$\pm$0.008(6)&0.9930$\pm$0.0005(4)&0.9931$\pm$0.0005(3)&0.991$\pm$0.001(5)&0.9933$\pm$0.0004(2)&\textbf{0.994$\pm$0.0004(1)} \\
Yeast-spo5 &0.893$\pm$0.008(6)&0.9732$\pm$0.003(3)&0.970$\pm$0.003(5)&0.9728$\pm$0.003(4)&0.9733$\pm$0.003(2)&\textbf{0.974$\pm$0.003(1)} \\
Yeast-spoem &0.914$\pm$0.011(6)&0.976$\pm$0.003(4)&0.975$\pm$0.002(5)&0.9777$\pm$0.002(3)&0.9780$\pm$0.002(2)&\textbf{0.979$\pm$0.002(1)} \\
Human Gene &0.456$\pm$0.089(6)&0.832$\pm$0.011(3)&0.766$\pm$0.020(4)&0.726$\pm$0.026(5)&0.833$\pm$0.011(2)&\textbf{0.834$\pm$0.011(1)} \\
Natural Scene &0.559$\pm$0.014(5)&0.490$\pm$0.082(6)&0.624$\pm$0.016(4)&0.697$\pm$0.011(3)&0.698$\pm$0.008(2)&\textbf{0.710$\pm$0.017(1)} \\
s-JAFFE &0.930$\pm$0.013(4)&0.920$\pm$0.014(5)&0.9337$\pm$0.018(3)&0.908$\pm$0.019(6)&0.9340$\pm$0.012(2)&\textbf{0.940$\pm$0.015(1)} \\
s-BU\_3DFE &0.924$\pm$0.004(5)&0.914$\pm$0.006(6)&0.938$\pm$0.002(2)&0.926$\pm$0.006(4)&0.935$\pm$0.004(3)&\textbf{0.954$\pm$0.002(1)} \\
Movie &0.850$\pm$0.008(5)&0.806$\pm$0.061(6)&0.880$\pm$0.006(4)&0.895$\pm$0.014(3)&0.905$\pm$0.008(2)&\textbf{0.912$\pm$0.010(1)} \\
\hline\noalign{\smallskip}
Avg. Rank &5.67&4.33&3.33&4.53&2.13&1.00 \\
\hline \hline
\end{tabular}
\end{center}
\vspace{-5mm}
\end{table*}

\begin{table*}[tb]
\scriptsize
\begin{center}
\caption{Experimental Results (mean$\pm$std(rank)) on the Real-world Datasets Measured by Intersection Similarity $\uparrow$} \label{table:Intersection}
\begin{tabular}{lcccccc}
\hline \hline \noalign{\smallskip}
Dataset & PT-Bayes & PT-SVM & AA-$k$NN & AA-BP & SA-IIS & SA-BFGS\\\hline\noalign{\smallskip}
Yeast-alpha &0.660$\pm$0.016(6)&0.949$\pm$0.006(4)&0.9581$\pm$0.002(2)&0.877$\pm$0.008(5)&0.9577$\pm$0.002(3)&\textbf{0.962$\pm$0.003(1)} \\
Yeast-cdc &0.681$\pm$0.015(6)&0.948$\pm$0.006(4)&0.9528$\pm$0.003(3)&0.891$\pm$0.007(5)&0.9531$\pm$0.002(2)&\textbf{0.957$\pm$0.003(1)} \\
Yeast-elu &0.695$\pm$0.019(6)&0.951$\pm$0.003(4)&0.9546$\pm$0.001(3)&0.899$\pm$0.006(5)&0.9547$\pm$0.001(2)&\textbf{0.959$\pm$0.001(1)} \\
Yeast-diau &0.764$\pm$0.008(6)&0.926$\pm$0.004(4)&0.937$\pm$0.002(3)&0.922$\pm$0.004(5)&0.938$\pm$0.002(2)&\textbf{0.940$\pm$0.003(1)} \\
Yeast-heat &0.773$\pm$0.018(6)&0.935$\pm$0.003(4)&0.936$\pm$0.002(3)&0.925$\pm$0.005(5)&0.938$\pm$0.001(2)&\textbf{0.940$\pm$0.001(1)} \\
Yeast-spo &0.765$\pm$0.010(6)&0.906$\pm$0.008(4)&0.908$\pm$0.004(3)&0.902$\pm$0.007(5)&0.914$\pm$0.005(2)&\textbf{0.915$\pm$0.006(1)} \\
Yeast-cold &0.802$\pm$0.012(6)&0.9339$\pm$0.004(5)&0.936$\pm$0.003(3)&0.9340$\pm$0.004(4)&0.938$\pm$0.002(2)&\textbf{0.941$\pm$0.002(1)} \\
Yeast-dtt &0.801$\pm$0.011(6)&0.954$\pm$0.002(4)&0.9549$\pm$0.002(3)&0.950$\pm$0.003(5)&0.9552$\pm$0.001(2)&\textbf{0.958$\pm$0.001(1)} \\
Yeast-spo5 &0.789$\pm$0.011(6)&0.9071$\pm$0.006(3)&0.904$\pm$0.005(5)&0.906$\pm$0.006(4)&0.9072$\pm$0.006(2)&\textbf{0.909$\pm$0.005(1)} \\
Yeast-spoem &0.810$\pm$0.017(6)&0.909$\pm$0.005(4)&0.907$\pm$0.004(5)&0.9108$\pm$0.005(3)&0.9109$\pm$0.005(2)&\textbf{0.913$\pm$0.005(1)} \\
Human Gene &0.470$\pm$0.062(6)&0.781$\pm$0.010(3)&0.742$\pm$0.014(4)&0.671$\pm$0.025(5)&0.783$\pm$0.010(2)&\textbf{0.784$\pm$0.010(1)} \\
Natural Scene &0.350$\pm$0.014(6)&0.364$\pm$0.055(5)&0.544$\pm$0.018(2)&0.499$\pm$0.012(3)&0.487$\pm$0.012(4)&\textbf{0.548$\pm$0.017(1)} \\
s-JAFFE &0.846$\pm$0.016(4)&0.839$\pm$0.015(5)&0.855$\pm$0.021(2)&0.824$\pm$0.022(6)&0.851$\pm$0.016(3)&\textbf{0.860$\pm$0.019(1)} \\
s-BU\_3DFE &0.834$\pm$0.003(5)&0.827$\pm$0.009(6)&0.872$\pm$0.002(2)&0.847$\pm$0.008(4)&0.862$\pm$0.004(3)&\textbf{0.884$\pm$0.003(1)} \\
Movie &0.725$\pm$0.011(5)&0.711$\pm$0.052(6)&0.780$\pm$0.007(4)&0.788$\pm$0.015(3)&0.800$\pm$0.010(2)&\textbf{0.809$\pm$0.013(1)} \\
\hline\noalign{\smallskip}
Avg. Rank &5.73&4.33&3.13&4.47&2.33&1.00 \\
\hline \hline
\end{tabular}
\end{center}
\vspace{-5mm}
\end{table*}

As can be seen from Table~\ref{table:Chebyshev} to Table~\ref{table:Intersection}, for each particular dataset, the rankings of the six LDL algorithms are often different on different measures. This corresponds to the design of a diverse set of evaluation measures described in Section~\ref{subsect:criteria}. Thus, when comparing two LDL algorithms on a particular dataset, all the six measures should be simultaneously considered. When looking at the average ranks over all the 15 real-world datasets, the rankings of the algorithms on five measures (Clark, Canberra, Kullback-Leibler, cosine, and intersection) are very consistent, i.e., SA-BFGS $\succ$ SA-IIS $\succ$ AA-$k$NN $\succ$ PT-SVM $\succ$ AA-BP$\succ$ PT-Bayes. The specialized algorithms (SA-BFGS and SA-IIS) generally perform better than those transformed from traditional learning algorithms (PT-Bayes, PT-SVM, AA-$k$NN, and AA-BP) because they directly aim to minimize the distance between the predicted and real label distributions. Moreover, SA-BFGS improves the performance of SA-IIS and achieves the best performance in most cases by using more effective optimization process. AA-BP and PT-Bayes perform worse than their siblings (AA-$k$NN and PT-SVM, respectively) because the many parameters in the BP neural network makes AA-BP vulnerable to overfitting, and the Gaussian distributions assumption for PT-Bayes might be inappropriate for the complex real-world datasets. Finally, AA-$k$NN performs better than PT-SVM because AA-$k$NN keeps the label distribution and thus keeps the overall labelling structure for each instance, while PT-SVM breaks down the original label distributions by weighted resampling. The average ranking on the Chebyshev distance, however, is slightly different in that the average ranks of PT-SVM and AA-BP are the same. Note that the Chebyshev distance only cares about the worst match over the whole label distribution (see Fig.~\ref{fig:measures}). The transformation from an LDL training set into an SLL training set for the `PT' algorithms (PT-SVM and PT-Bayes) breaks down the original label distributions by weighted resampling. The unbalanced and unstable resampling process could possibly make the worst case worse (e.g., the label with small description degree is not sampled at all). This explains why the `PT' algorithms perform slightly worse on the Chebyshev distance.

One apparent difference between the results on the real-world datasets and those on the artificial dataset is that PT-Bayes performs much worse on the real-world datasets than on the artificial dataset. This might be because that the Gaussian assumption for each label may match the relatively simple artificial toy dataset better than those complex real-world datasets. Another noticeable point is that the values of the Clark distance (Table~\ref{table:Clark}) and the Canberra metric (Table~\ref{table:Canberra}) on the Human Gene and Natural Scene datasets are significantly larger than those on other datasets, no matter which LDL algorithm is applied. The reason is of two-fold. On the one hand, the two datasets might be inherently more difficult to learn than other datasets, which can be evidenced by the relatively poor performances on them evaluated by other measures. On the other hand, such gap on the Clark distance and Canberra metric is particularly significant due to the definition of these two measures. As mentioned before, the Clark distance and Canberra metric are sensitive to small changes near zero \cite{Gordon99}. It can be found from Fig.~\ref{fig:measures} that when the description degree of the $j$-th label in the real label distribution, $d_j$, is close to zero, the $j$-th additive term in the formula of `Clark' or `Canberra' will be close to its maximum value 1. If there are many near-zero description degrees in the real label distribution, then the value of `Clark' or `Canberra' tends to be large. For the Human Gene dataset, one gene is usually related to only a few diseases. Similarly, for the Natural Scene dataset, one image is usually relevant to only a few scene classes. Thus, the real label distributions in both of these two datasets contain many near-zero description degrees, which makes the values of `Clark' and `Canberra' relatively large.

\section{Summary and Discussions}
\label{sect:conclusion}
This paper proposes \emph{label distribution learning}, which is a more general learning framework than single-label learning and multi-label learning. It can deal with not only multiple labels of one instance, but also the different importance of these labels. This paper proposes six working LDL algorithms in three ways: problem transformation, algorithm adaptation, and specialized algorithm design. In order to compare these algorithms, six evaluation measures are used and the first batch of label distribution datasets are prepared and made publicly available. Experimental results on one artificial and fifteen real-world datasets show clear advantages of the specialized algorithms. This illustrates that the characteristics of LDL require special design to achieve good performance.

LDL is motivated by the real-world data with natural measures of description degrees (e.g., gene expression level). However, as a general learning framework, LDL might also be used to solve other kinds of problems. Generally speaking, there are at least three scenarios where LDL could be helpful:
\begin{enumerate}
  \item There is a natural measure of description degree that associates the labels with the instances. This is the most direct application of LDL, as described in this paper.
  \item When there are multiple labeling sources (e.g., multiple experts) for one instance, known as \emph{learning with auxiliary information} \cite{NguyenVH11} or \emph{learning from multiple experts} \cite{ValizadeganNH13}, the annotations from different sources might be significantly inconsistent. In such case, it is usually better for the learning algorithm to integrate the labels from all the sources rather than to decide one or more `winning label(s)' via majority voting \cite{RaykarYZ10}. One good way to incorporate all the labeling sources is to generate a label distribution for the instance: the label favored by more sources is given a higher description degree, while that chosen by fewer sources is assigned with a lower description degree. In this way, the multi-labeling-source problem is transformed into a LDL problem.
  \item Some labels are highly correlated with other labels. Utilizing such correlation is one of the most important approaches to improve the learning process \cite{KangJS06,WangHZ08,SunZZ10}. LDL provides a new way toward this purpose. The key step is to transform an SLL or MLL problem into an LDL problem. This can be achieved by generating a label distribution for each instance according to the correlation among the labels.
\end{enumerate}
Each of the three scenarios actually covers a vast area of applications. A lot of interesting work, both at the theoretical level and at the application level, may be conducted in the future. In addition, the relationship between LDL and traditional machine learning paradigms, e.g., SLL and MLL, is another promising research direction. For example, we have applied LDL to solve the MLL problems by generating the label distributions via an iterative label propagation process among the training samples \cite{LiZG15}. Comprehensive experiments show that the LDL-augmented MLL algorithm could significantly improve the state-of-the-art MLL performance.

\appendix[Derivation of Eq.~(\ref{eq:delta_eq})]
The change of $T(\bm{\theta})$ in Eq.~(\ref{eq:target_fun}) between adjacent steps is
\begin{eqnarray}
T(\bm{\theta}+\bm{\Delta})-T(\bm{\theta})=\sum\limits_{i,j}
d_{\bm{x}_i}^{y_j}\sum\limits_k \delta_{y_j,k} g_k(\bm{x}_i)-\nonumber\\
\sum\limits_i \ln\sum\limits_j p(y_j|\bm{x}_i;\bm{\theta}) \exp\left(\sum\limits_k \delta_{y_j,k} g_k(\bm{x}_i)\right),
\end{eqnarray}
where $\delta_{y_j,k}$ is the increment for $\theta_{y_j,k}$. Applying the inequality $-\ln x \geq 1-x$ yields
\begin{eqnarray}\label{eq:T_delta}
T(\bm{\theta}+\bm{\Delta})-T(\bm{\theta}) \geq \sum\limits_{i,j}
d_{\bm{x}_i}^{y_j}\sum\limits_k \delta_{y_j,k} g_k(\bm{x}_i)+n- \nonumber\\
\sum\limits_{i,j} p(y_j|\bm{x}_i;\bm{\theta}) \exp\left(\sum\limits_k \delta_{y_j,k} g_k(\bm{x}_i)\right).
\end{eqnarray}
Differentiating the right side of Eq.~(\ref{eq:T_delta}) w.r.t. $\delta_{y_j,k}$ yields the coupled equations of
$\delta_{y,k}$ which are hard to be solved. To decouple the interaction among $\delta_{y,k}$, Jensen's inequality is
applied here, i.e., for a probability mass function $p(x)$ and another arbitrary function $f(x)$,
\begin{equation}
\exp \left(\sum\limits_x p(x)f(x)\right) \leq \sum\limits_x p(x)\exp \left(f(x)\right).
\end{equation}
The last term of Eq.~(\ref{eq:T_delta}) can be rewritten as
\begin{eqnarray}
\sum\limits_{i,j} p(y_j|\bm{x}_i;\bm{\theta}) \exp\left(\sum\limits_k \delta_{y_j,k} s(g_k(\bm{x}_i))
g^{\#}(\bm{x}_i)\frac{|g_k(\bm{x}_i)|}{g^{\#}(\bm{x}_i)}\right),
\end{eqnarray}
where $g^{\#}(\bm{x}_i)= \sum\nolimits_k |g_k(\bm{x}_i)|$ and $s(g_k(\bm{x}_i))$ is the sign of $g_k(\bm{x}_i)$. Since
$|g_k(\bm{x}_i)|/g^{\#}(\bm{x}_i)$ can be viewed as a probability mass function, Jensen's inequality can be applied to Eq.~(\ref{eq:T_delta}) to yield
\begin{eqnarray}\label{eq:T_delta_lb}
T(\bm{\theta}+\bm{\Delta})-T(\bm{\theta}) \geq \sum\limits_{i,j}
d_{\bm{x}_i}^{y_j}\sum\limits_k \delta_{y_j,k} g_k(\bm{x}_i)+n- \nonumber\\
\sum\limits_{i,j} p(y_j|\bm{x}_i;\bm{\theta}) \sum\limits_k \frac{|g_k(\bm{x}_i)|}{g^{\#}(\bm{x}_i)}\exp(\delta_{y_j,k}
s(g_k(\bm{x}_i))g^{\#}(\bm{x}_i)).
\end{eqnarray}
Denote the right side of Eq.~(\ref{eq:T_delta_lb}) as $\mathcal{A}(\bm{\Delta}|\bm{\theta})$, which is a lower bound
to $T(\bm{\theta}+\bm{\Delta})-T(\bm{\theta})$. Setting the derivative of $\mathcal{A}(\bm{\Delta}|\bm{\theta})$ w.r.t.
$\delta_{y_j,k}$ to zero gives Eq.~(\ref{eq:delta_eq}).

\section*{Acknowledgment}
This research was supported by the National Science Foundation of China (61273300, 61232007), the Jiangsu Natural Science Funds for Distinguished Young Scholar (BK20140022), and the Collaborative Innovation Center of Wireless Communications Technology.



%

\bibliographystyle{IEEEtranS}
\bibliography{dll}
\vspace{-5mm}\begin{IEEEbiography}[{\includegraphics[width=1in,height=1.25in,clip,keepaspectratio]{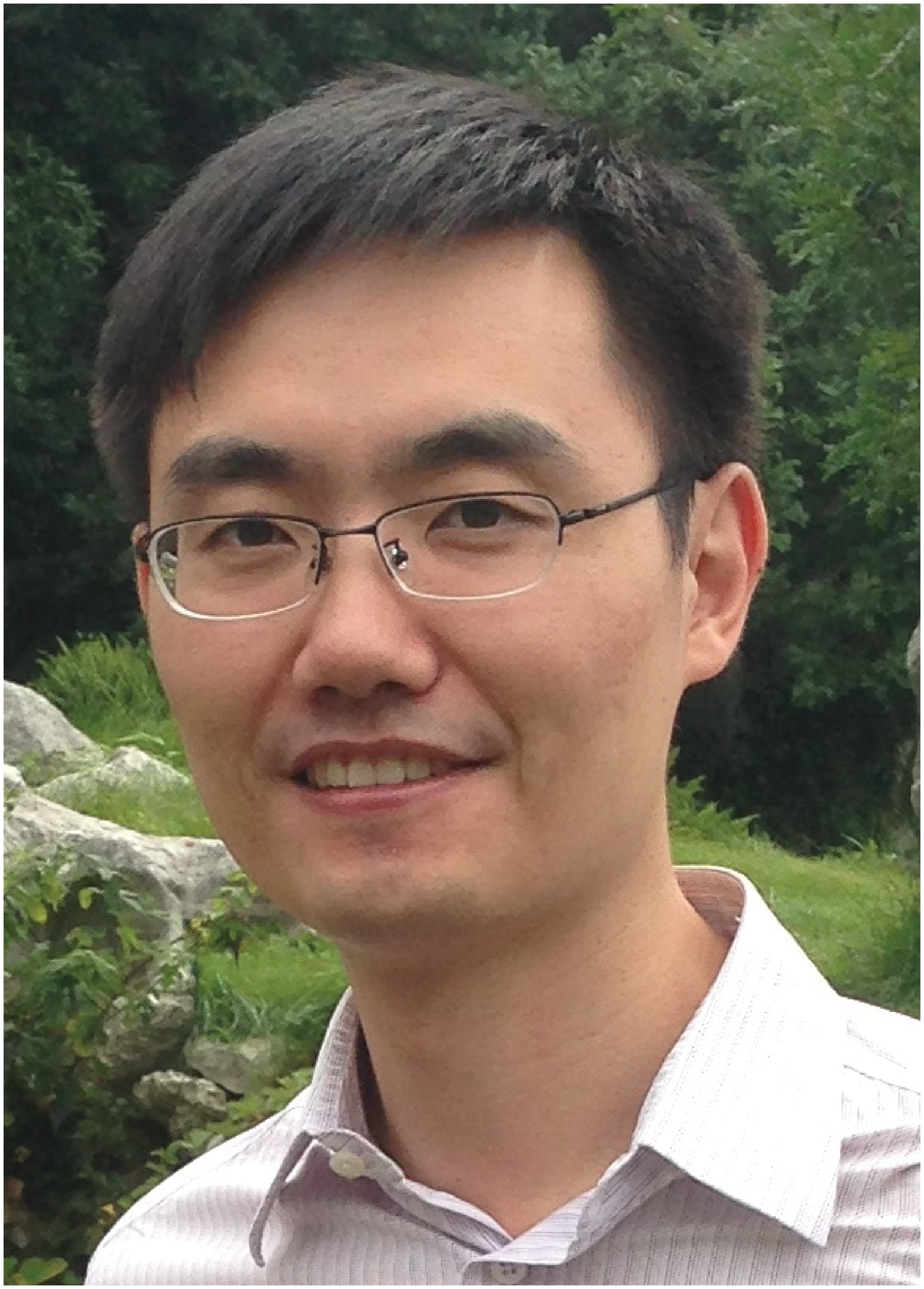}}]
{Xin Geng} (M'13) received the B.Sc. (2001) and M.Sc. (2004) degrees in computer science from Nanjing University, China, and the Ph.D (2008) degree from Deakin University, Australia. He joined the School of Computer Science and Engineering at Southeast University, China, in 2008, and is currently a professor and vice dean of the school. His research interests include pattern recognition, machine learning, and computer vision. He has published over 40 refereed papers and holds 4 patents in these areas.
\end{IEEEbiography}

\end{document}